\documentclass[a4paper, twoside, 11pt]{article}

\usepackage{lmodern} %% Supposedly nicer fonts for main text, barely noticeable
\usepackage[utf8]{inputenc}
\usepackage[english]{babel}
\usepackage{microtype} %Purpose: Improves the typographic quality of the document. What it does: Enables micro-typographic enhancements like character protrusion and font expansion. This leads to better line-breaking and spacing, especially in justified text. Use case: Always recommended for professional-quality typesetting. Improves the overall appearance without manual tweaks.

\usepackage[noblocks]{authblk} %% Affiliations with superscripts (1,2,3)
\usepackage[round]{natbib}
\usepackage{graphicx}
\usepackage{fancyhdr}
\usepackage[tableposition=top, justification=justified, font=small]{caption}
\usepackage{titlesec}
\usepackage{multicol}
\usepackage{afterpage}
\usepackage{lipsum}
\usepackage[dvipsnames, x11names]{xcolor}

\usepackage[inner=2.5cm, outer=2.5cm, top=2.5cm, bottom=3cm, marginparsep=0cm, marginparwidth=0cm]{geometry}
\usepackage{setspace}
\usepackage{array}
\usepackage{makecell}
\usepackage{bm}
\usepackage{tikz}
\usepackage{float}
\usepackage[framemethod]{mdframed}
\usepackage{amsmath}
\usepackage{amssymb}
\usepackage{amsthm}

\usepackage{subcaption}
\usepackage{csquotes}
\usepackage{tabularx}
\usepackage[table]{xcolor}
\usepackage{pifont}
\usepackage{bbm}
\usepackage{booktabs}
\usepackage{diagbox}
\usepackage{smartdiagram}
\usepackage{enumitem}
\usepackage{arydshln} % Dashed lines in tables
\usepackage{framed}
\usepackage{pifont}
\usepackage{algorithm, algorithmic}
\usepackage{multirow} 
\usepackage{threeparttable}

\definecolor{mycustomcolor1}{rgb}{0.08, 0.38, 0.74} %denim
\colorlet{boxcolor}{mycustomcolor1!50}

\usepackage[colorlinks=true, linkcolor=mycustomcolor1, urlcolor=mycustomcolor1, citecolor=mycustomcolor1]{hyperref}

\titleformat{\section}{\Large \bfseries \sffamily}{\thesection.}{0.3em}{}

\definecolor{shadecolor}{RGB}{230,230,230}

\titleformat{\subsection}{\large \bfseries \sffamily }{\thesubsection}{0.3em}{}

\titleformat{\subsubsection}{\normalsize \bfseries \sffamily }{\thesubsubsection.}{0.3em}{}

\titleformat*{\paragraph}{\sffamily\bfseries}

\renewenvironment{abstract}
{\vskip 2.5ex {\noindent \Large \bfseries \sffamily Abstract}\vspace{0.7ex} \\ %
  \bgroup\noindent\ignorespaces}%
{\par\egroup\vskip 2.5ex}

\newenvironment{keywords}
{\bgroup\leftskip 20pt\rightskip 20pt \small\noindent{\bfseries \sffamily Keywords:} }%
{\par\egroup\vskip 10ex}

\fancypagestyle{firstpage}{%
  
  \fancyhead[]{}
  \fancyfoot[C]{}
  \fancyfoot[L]{\footnotesize \sffamily}
}

\fancypagestyle{otherpages}{%
  
   \fancyhead[L]{}
   \fancyhead[C]{}
   \fancyhead[R]{}
  \fancyfoot[L]{\footnotesize \sffamily \textit{}}
  \fancyfoot[C]{\sffamily \thepage}
}

\makeatletter
\renewcommand{\maketitle}{\bgroup\setlength{\parindent}{0pt}

\begin{center}
    \setstretch{2.0} %% allows a nicer formatting of the title by adding more space between lines
    \textbf{\Huge \sffamily \@title}
\end{center}
\vspace{0.15in}

\begin{flushleft}
    \textbf{\bfseries \large \sffamily \@author}
\end{flushleft}\egroup
}

\makeatother
\DeclareCaptionLabelFormat{labelformat}{\bfseries \sffamily #1 #2}
\mdfsetup{skipabove=\topskip, skipbelow=\topskip}

\newcounter{nicebox}

\newfloat{floatbox}{thp}{lop}
\floatname{floatbox}{Box}

\newcommand{\beginsupplement}{%
        \setcounter{table}{0}
        \renewcommand{\thetable}{S\arabic{table}}%
        \setcounter{figure}{0}
        \renewcommand{\thefigure}{S\arabic{figure}}%
        \setcounter{section}{0}
        \renewcommand{\thesection}{S\arabic{section}}
        \renewcommand{\theHsection}{S\arabic{section}}
     }

\DeclareMathAlphabet{\mathsfit}{\encodingdefault}{\sfdefault}{m}{sl}
\SetMathAlphabet{\mathsfit}{bold}{\encodingdefault}{\sfdefault}{bx}{n}

\usepackage{multibib}
\newcites{supp}{Supplementary References}

\definecolor{aliceblue}{rgb}{0.94, 0.97, 1.0}
\definecolor{airforceblue}{rgb}{0.36, 0.54, 0.66}
\definecolor{limegreen}{rgb}{0.2, 0.8, 0.2}

\newcommand{\equalfirst}{$\ast$}  
\newcommand{\equallast}{$\star$}

\makeatletter
\renewcommand*{\@fnsymbol}[1]{%
  \ifcase#1
    {}% 0 - nothing
  \or \textsuperscript{$\ast$}% 1
  \or \textsuperscript{$\star$}% 2
  \or \textsuperscript{\ddag}% 3
  \else \@ctrerr
  \fi
}
\makeatother

\RequirePackage{silence}
\newcommand{\githubrepofig}{https://www.github.com/aramis-lab/CIs_Medical_Imaging_Plots}
\newcommand{\githubrepocode}{http://www.github.com/aramis-lab/CIs_Medical_Imaging}

\newcommand{\E}{\mathbb{E}}

\newcommand{\dd}{\mathrm{d}}

\theoremstyle{plain}
\newtheorem{proposition}{Proposition}
\theoremstyle{definition}

\newcolumntype{C}[1]{>{\centering\arraybackslash}p{#1}}
\newcolumntype{L}[1]{>{\raggedright\arraybackslash}p{#1}}

\begin{document}

\setlength\dashlinedash{0.2pt}
\setlength\dashlinegap{1.5pt}
\setlength\arrayrulewidth{0.3pt}

%%%%%%%%%%%%%%%%%%%%%%%%%%%%%%%%%%%%%%%%%%%%%%%%%%%%%%%%%%%%%%%%%
%           I N F O R M A T I O N   T O   C H A N G E
%%%%%%%%%%%%%%%%%%%%%%%%%%%%%%%%%%%%%%%%%%%%%%%%%%%%%%%%%%%%%%%%%

\newcommand{\papertitle}{Performance uncertainty in medical image analysis: a large-scale investigation of confidence intervals}

% Author name displayed in the running head
%\newcommand{\runningauthor}{André\textsuperscript{$\ast$}, Heitz\textsuperscript{$\ast$} et al} 

% Title displayed in the running head
%\newcommand{\runningheadtitle}{A large-scale investigation of confidence intervals in medical image analysis}

% Chapter number
%\newcommand{\chapternumber}{4}

% E-mail address of the corresponding author
\newcommand{\emailaddress}{pascaline.a0307@gmail.com}
\newcommand{\emailaddressbis}{charles.heitz1308@gmail.com}

% Authors' names and affiliation numbers
% Authors

\author[a]{Pascaline André\textsuperscript{\equalfirst}}
\author[a]{Charles Heitz\textsuperscript{\equalfirst}}
\author[b, e, f]{Evangelia Christodoulou}
\author[b, d]{Annika Reinke}
\author[c, g, h]{Carole H. Sudre}
\author[g, h]{Michela Antonelli}
\author[b, e]{Patrick Godau}
\author[g]{M. Jorge Cardoso}
\author[a]{Antoine Gilson}
\author[a]{Sophie Tezenas du Montcel}
\author[i]{Gaël Varoquaux\textsuperscript{\equallast,}}
\author[b, d, e, j, k]{Lena Maier-Hein\textsuperscript{\equallast}}
\author[a]{Olivier Colliot\textsuperscript{\equallast}}

% Affiliations
\affil[a]{Sorbonne Universit\'{e}, Institut du Cerveau – Paris Brain Institute - ICM, CNRS, Inria, Inserm, AP-HP, H\^{o}pital de la Piti\'{e}-Salp\^{e}tri\`{e}re, F-75013, Paris, France}
\affil[b]{German Cancer Research Center (DKFZ) Heidelberg, Div. Intelligent Medical Systems, Germany}
\affil[c]{Unit for Lifelong Health and Ageing at UCL, Department of Population Science and Experimental Medicine and Hawkes Institute, Centre for Medical Image Computing, Department of Computer Science, University College London, UK}
\affil[d]{DKFZ Heidelberg, Helmholtz Imaging, Germany
}
\affil[e]{National Center for Tumor Diseases (NCT), NCT Heidelberg, a partnership between DKFZ and Heidelberg University Hospital, Germany}

\affil[f]{AI Health Innovation Cluster, Germany }

\affil[g]{School of Biomedical Engineering and Imaging Science, King’s College London, UK}

\affil[h]{Hawkes Institute, Department of Computer Science, University College London, UK}
\affil[i]{SODA project team, INRIA Saclay-Île de France, France}
\affil[j]{Faculty of Mathematics and Computer Science, Heidelberg University, Germany}
\affil[k]{Medical Faculty, Heidelberg University, Germany}

%%%%%%%%%%%%%%%%%%%%%%%%%%%%%%%%%%%%%%%%%%%%%%%%%%%%%%%%%%%%%%%%%

% Alternative option for explaining symbols (instead of footnotes)
% \noindent
% \textsuperscript{$\ast$} These authors contributed equally as first authors. \\
% \textsuperscript{$\star$} These authors contributed equally as senior authors. \\
% \textsuperscript{\ddag} Correspondence to: \href{mailto:\emailaddress}{\emailaddress} and \href{mailto:\emailaddressbis}{\emailaddressbis}.

\begingroup
\renewcommand{\thefootnote}{\fnsymbol{footnote}}
\footnotetext[1]{These authors contributed equally as first authors.}
\footnotetext[2]{These authors contributed equally as senior authors.}
\footnotetext[3]{Correspondence to: \href{mailto:\emailaddress}{\emailaddress} and \href{mailto:\emailaddressbis}{\emailaddressbis}.}
\endgroup

% Title of the paper
\title{\papertitle}

\maketitle

% Restore the geometry and change the page style for the other pages
\afterpage{\aftergroup\restoregeometry}
\pagestyle{otherpages}

\vspace{0.5cm}

% %% Remove for the postprint
% \definecolor{mypurple}{rgb}{.4,.0,.6}
% \newcommand{\preprint}[1]{\textcolor{black}{\textbf{\sffamily #1}}}
% \begin{center}
% \fbox{\parbox{0.70\linewidth}{\preprint{Disclaimer: this is a working
% paper, and is still work in progress. Please do not hesitate to provide comments or point towards missing
% references to \href{mailto:pascaline.a0307@gmail.com,charles.heitz1308@gmail.com}{pascaline.a0307@gmail.com and charles.heitz1308@gmail.com}.}}}
% \end{center}
% \vspace{0.5cm}

% Abstract
\begin{abstract}
Performance uncertainty quantification is essential for reliable validation and eventual clinical translation of medical imaging artificial intelligence (AI). Confidence intervals (CIs) play a central role in this process by indicating how precise a reported performance estimate is. Yet, due to the limited amount of work examining CI behavior in medical imaging, the community remains largely unaware of how many diverse CI methods exist and how they behave in specific settings. The purpose of this study is to close this gap.
To this end, we conducted a large-scale empirical analysis across a total of 24 segmentation and classification tasks, using 19 trained models per task group, a broad spectrum of commonly used performance metrics, multiple aggregation strategies, and several widely adopted CI methods. Reliability (coverage) and precision (width) of each CI method were estimated across all settings to characterize their dependence on study characteristics.
Our analysis revealed five principal findings: 1) the sample size required for reliable CIs varies from a few dozens to several thousands of cases depending on study parameters; 2) CI behavior is strongly affected by the choice of performance metric; 3) aggregation strategy substantially influences the reliability of CIs, e.g. they require more observations for macro than for micro; 4) the machine learning problem (segmentation versus classification) modulates these effects; 5) different CI methods are not equally reliable and precise depending on the use case. Finally, we derived practical implications of this study in the form of a decision tree which shall prove useful to the community. This paves the way for future consensus guidelines on reporting performance uncertainty.

\end{abstract}

% Keywords
% \noindent \sffamily{\textbf{Keywords}}: Image synthesis, Segmentation, Registration, Atlas database

\begin{keywords}
Medical imaging, Validation, Confidence intervals, Segmentation, Classification, Uncertainty quantification, Trustworthy machine learning

\end{keywords}

\section{Introduction}
\label{sec:introduction}
Over the past decade, there has been an exponential increase in the number of papers on machine learning (ML) for medical imaging \citep{xie2025evolution}. Despite this rapid growth, there is still little translation from the literature into routine clinical practice. One hindrance to translation is the lack of uncertainty quantification of performance estimates in research papers \citep{christodoulou2024confidence}. Indeed, performance estimates are computed from test sets of finite size and are therefore inherently subject to sampling variability. As a result, these estimates should be interpreted together with measures of uncertainty. Confidence intervals (CIs) are arguably the most common way to provide a measure of the uncertainty of the reported performance estimate. As a matter of fact, the U.S. Food and Drug Administration (FDA) requires companies to report confidence intervals along with performance estimates when submitting Artificial Intelligence (AI)-enabled medical devices for regulatory approval \citep{fda2024-D-4488}.

\begin{figure}[htbp]
    \centering
    \includegraphics[width=0.8\linewidth]{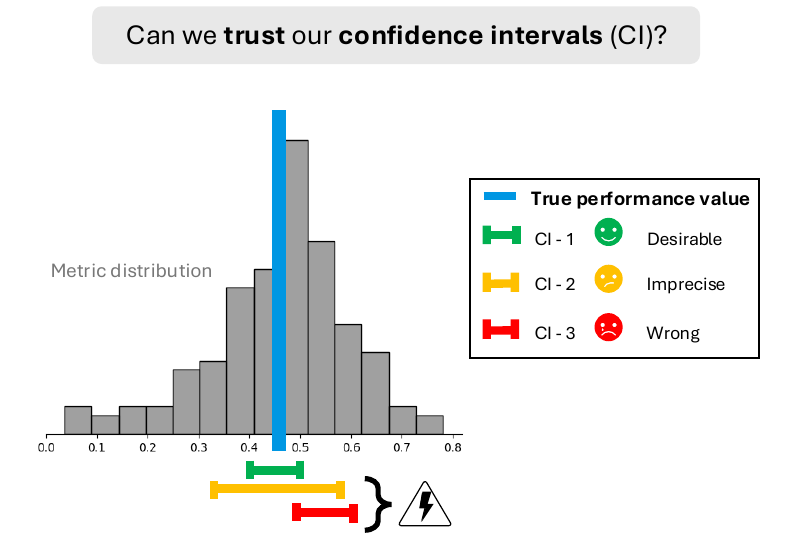}
    \caption{\textbf{\sffamily Not all confidence interval (CI) methods can be universally trusted.} Here, CI 2 is too wide and thus underconfident about the target value. CI 3 is unreliable as it completely misses the target value. CI 1 is both reliable and precise.}
    \label{fig:1}
\end{figure}

However, estimating uncertainty through CIs is not a trivial task. Different CI methods exist and the choice of an inappropriate method can lead to either overconfidence or underconfidence in the estimated model performance (Figure \ref{fig:1}). In particular, methods producing too narrow intervals often lead to overconfidence, as the proposed interval might miss the true performance of the model. Conversely, methods producing too wide intervals are underconfident as they do not manage to locate the true value with enough precision. A good confidence interval method should therefore balance these two aspects by accurately identifying the range of plausible values for the performance estimate while providing an informative assessment of its precision. In other words, a CI method has two key characteristics: reliability and precision. Reliability means that the method most often catches the true value. Precision means that the CI width is as small as possible.

Despite their importance, there are very few papers on CIs in medical imaging AI \citep{jurdi2023confidence,varoquaux2018cross} and their scope remains narrow, typically considering only a small set of models and performance metrics. While related work exists in generalist AI \citep{miller2024adding} and in specific fields, e.g. psychology \citep{rousselet2023introduction} and economics \citep{cherry1996comparison}, they either rely on parametric simulations or study specific statistical regimes that may differ from medical image analysis. 

Medical imaging AI exhibits domain-specific characteristics that can strongly influence the behavior of confidence interval (CI) methods. In particular, a wide range of task- and application-specific performance metrics are required and guidelines have been established to support appropriate metric selection \citep{maier2024metrics}. In addition, medical imaging studies often rely on small to moderate sample sizes compared to those typically used in general ML or computer vision~ \citep{christodoulou2025false}.

\begin{figure}[hbtp]
    \centering
    \includegraphics[width=0.8\linewidth]{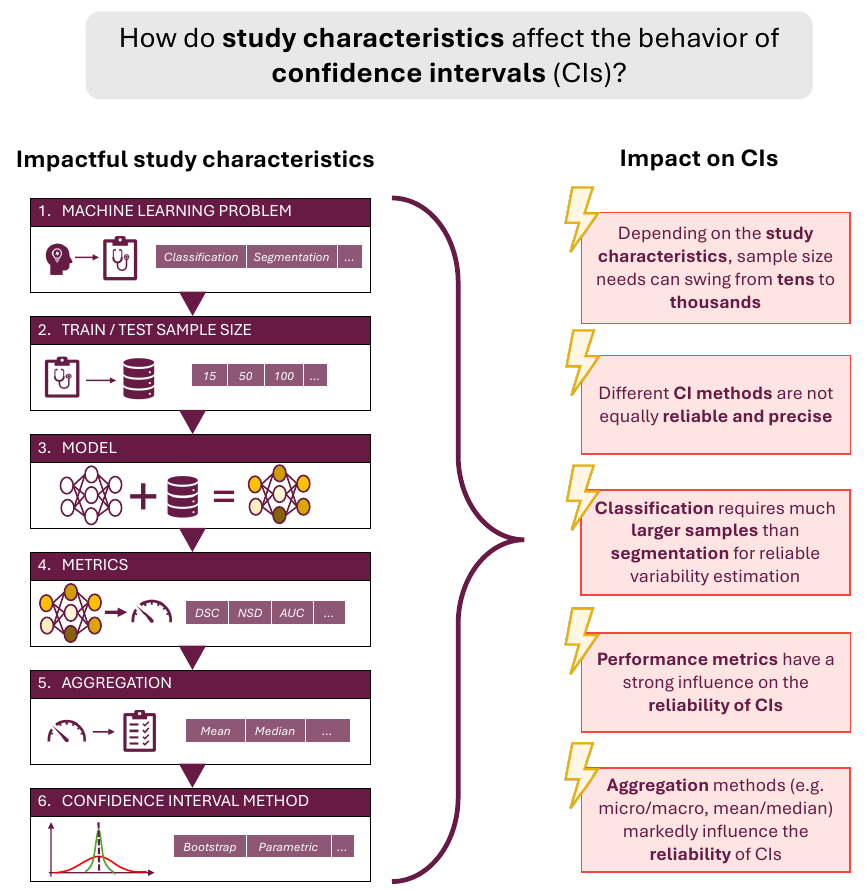}
    \caption{\textbf{\sffamily Link between study characteristics and behavior of confidence intervals (CIs).} Left panel: main characteristics of a medical imaging study. Right panel: key findings of our investigation, revealing how study characteristics impact the reliability and precision of CIs.}
    \label{fig:2}
\end{figure}

In this work, we present a large-scale investigation of the behavior of widely used CI methods in the context of medical imaging. Specifically, our central research question was: how are reliability and precision of CI methods affected by study characteristics including ML problem (classification, segmentation), sample size, performance metrics and aggregation method (Figure \ref{fig:2}, left panel)?

Our main contributions are the following: 
\begin{itemize}
    \item We performed an extensive experimental study of various segmentation and classification tasks (12 for each) and trained models (19 for each learning problem) corresponding to 456 ($2\times12\times19$) benchmark instances covering a variety of settings. 
    \item Across the 456 benchmark instances, we systematically studied the reliability (coverage) and precision (width) of different CI methods (parametric and non-parametric) for different performance metrics, aggregation methods and sample sizes.
    \item We unveiled key relationships between CI behavior and study characteristics (Figure \ref{fig:2}, right panel).
\end{itemize}

A preliminary version of this work appeared in the proceedings of the MICCAI BRIDGE Workshop 2025 \citep{andre2025some}. The present journal version considerably expands upon this by considering: 1) not only segmentation but also classification; 2) 14 performance metrics instead of 2; 3) 5 summary statistics instead of 2; 4) not only the coverage but also the width of the CIs. Such combinations allow us to study CI behavior in 108 different configurations versus 4 in the workshop paper.

\newpage
\section{Background and related work}

This section first provides a brief, non-technical, overview of the concepts used in this paper, including metrics, summary statistics and confidence intervals (Section \ref{subsec:nontechnical}). This part can be skipped by readers who are already familiar with these notions. Readers who prefer a more formal treatment can refer to Supplementary Sections \ref{sec:empirical_evaluation} and \ref{sec:CIs}. Section \ref{sec:related_work} then describes  related work on confidence intervals. 

\subsection{A non-technical overview of metrics, summary statistics and CIs}
\label{subsec:nontechnical}

\subsubsection{Metrics, summary statistics and empirical evaluations}

Medical imaging AI models are typically evaluated empirically through metrics computed on a test set. This test set is assumed to be a set of observations of an independent and identically distributed (i.i.d.) random sample drawn from a more general population.

A wide variety of performance metrics are used in medical imaging AI, and they need to be carefully chosen according to the task at hand~\citep{maier2024metrics}. Among these, an important conceptual distinction is between individual-level metrics and sample-level metrics.

\textbf{Individual-level metrics} are computed at the level of each observation and typically used for tasks where the difference between the prediction and the reference needs a first aggregation at the individual level. A prototypical example is segmentation where a segmentation map is produced for each individual image and compared to a reference using various metrics (e.g. Dice Similarity Coefficient (DSC), Normalized Surface Distance (NSD) etc.). To report a performance value at the sample level, the metric needs to be further aggregated over the test sample through a \textbf{summary statistic} that may for instance reflect the central tendency (e.g. the empirical mean) or the dispersion (e.g. the empirical standard deviation). Indeed, a \textbf{summary statistic} is always computed from a sample. As such, it is only an empirical estimate of its true value, also called population value, to which one has, by definition, never access (e.g. the empirical mean DSC computed from the test set that estimates the true mean DSC of the whole population). 

\textbf{Sample-level metrics} are computed over a set of individuals and typically used in tasks where the individual-level prediction is a simple number or class and one aggregates these predictions over a population. Prototypical examples include regression (e.g. coefficient of determination, Mean Absolute Error…) and classification (e.g. accuracy, Area Under the Curve…). In such a case, the metric is actually a summary statistic. The simplest example is that the accuracy is actually the mean of the indicator function over a set of individuals. 

A critical case for distinguishing between individual-level metrics and sample-level metrics is when one looks at their variance. The variance of the DSC is computed across a sample. It tells you how variable your performance is across observations. This variance is not reduced by increasing sample size, it will only be more precisely estimated. In contrast, the variance of a sample-level metric such as the accuracy reflects the variability across different samples. This variance will shrink with increasing sample size. The same behavior is observed for summary statistics derived from individual-level metrics, such as the variance of the mean DSC, which also decreases with increasing sample size.

A more formal treatment can be found in supplementary~\ref{sec:empirical_evaluation}. 

\subsubsection{Confidence Intervals}

CIs characterize the precision of performance estimates \citep{wasserman2004all}. As explained above, model performance is usually reported as an empirical summary statistic (e.g. mean DSC or accuracy) computed on a test set, which approximates true population performance from finite data. Because the test set is only one finite sample from this population, the reported performance varies depending on which observations are included. In other words, this empirical estimate is noisy or imprecise.
CIs quantify this variability by measuring how much the empirical estimate would change if the model were evaluated on different test sets drawn from the same population. As a result, they provide a range of values that is likely to contain the true population performance (e.g. the true mean DSC).

Formally, a test set is the realization of a random sample. Each realization (i.e. each possible test set) results in a different CI value. Therefore, a CI is a couple of random variables (its bounds). In practice, the term CI is used to refer both to a random interval and its realization. This abuse of terminology is fine, one just needs to remember that coverage is not defined for a single realization but only for a series of realizations.
Note that other methods for reporting variability exist such as Bayesian credible intervals \citep{jeffreys1998theory} but they rely on a Bayesian framework which is different from frequentist CIs. The present paper does not take position on which tool is best but chooses to focus on CIs since they are arguably the most common and are the one mentioned in regulatory guidelines \citep{fda2024-D-4488}.

CIs are characterized by two properties: coverage and width. 
CIs are assigned a specific \textbf{theoretical coverage} value, denoted as $1-\alpha$ (e.g. $95\%$ when $\alpha=0.05$). It is the probability that the random CI contains the true value of the statistic of the underlying true distribution. In practice, one draws a finite number of samples to assess what is referred to as the \textbf{empirical coverage}, defined as the proportion of CIs that contain the true value of the summary statistic (e.g. true mean DSC). As the sample size $n$ increases, the empirical coverage is expected to converge to the theoretical coverage\footnote{In general \cite{wasserman2004all}, to qualify as a CI, the random interval should contain the true value with probability larger or equal to $1-\alpha$, rather than equal to $1-\alpha$. In practice, most the methods that we studied are built to achieve equality but this is not the case of CIs derived from concentration inequalities.}. Of course, data is not infinite and thus what matters is the empirical coverage for a given $n$. Thus for a given sample size $n$, a reliable CI method is expected to produce an empirical coverage that is close to the theoretical coverage. When this is not the case, two distinct failure modes can be identified: undercoverage and overcoverage.

Undercoverage occurs when the empirical coverage is lower than the theoretical coverage mainly because intervals are either: i) individually too narrow, even if they are centered on the true value on average, or ii) on average not centered around the true value (CI 3 on Figure \ref{fig:1}). In contrast, overcoverage occurs when the empirical coverage exceeds the theoretical coverage. This can happen when the CI method produces intervals that are too wide (see CI 2 in Figure \ref{fig:1}). In such a case, the CI is imprecise (a tighter interval could likely have been obtained). Another case of overcoverage is when the statistic is invariant with respect to the sample drawn. Each of the CIs will be reduced to a single point, which will appear as equal to the true statistic, thus producing a $100\%$ empirical coverage. Such point intervals are misleading, as they hide the uncertainty associated with unobserved data realizations.

The \textbf{width} of a CI reflects the precision of the performance estimate: a small width means that our estimation is precise and a large width the opposite.
A small width is desirable but has to be paired with a good coverage, otherwise the interval misleads the user towards overconfidence on their performance estimate. A large width may be linked either to true imprecision of the estimation or to overcoverage of the CI method. For a given CI method, the width is influenced by two factors: it grows with the variability of the performance across observations and it shrinks with the test set size. Typically, the widths of CI are proportional to $1/\sqrt{n}$.

In summary, both the coverage and the width are important but their roles differ: a good CI method should first be reliable (adequate coverage) and then be precise (small width).

\subsubsection{Existing methods for CI computation }
CI methods can be broadly divided into two categories: parametric and non-parametric methods. Parametric methods leverage distributional properties of either the summary statistic or of the individual-level performance metrics, whereas non-parametric methods make fewer to no assumptions on the true distribution of interest.

\paragraph{Parametric CI methods.} They are specific to the summary statistic/sample-level metric. For CIs of the empirical mean, several classical approaches exist, most of which rely on a Gaussian approximation. The {\sl parametric z} method assumes that the metric variance is known (e.g. true DSC variance on the whole population), whereas the {\sl parametric t} method estimates the variance from the data (e.g. empirical DSC variance computed from the test set). For other summary statistics (median, trimmed mean, standard-deviation), such methods exist 
\citep{cramer1999mathematical} but are not the most commonly used in practice. A large body of literature exists for CIs of sample-level metrics. Again methods are specific to the sample-level metric. For metrics that can be expressed as binomial proportions, several well-known approaches have been proposed, including the Agresti-Coull interval \citep{agresti1998approximate}, Wilson interval \citep{wilson1927probable}, Wald (or normal) interval \citep{marquis1820theorie}, and Clopper-Pearson interval \citep{clopper1934use}. Similarly, multiple parametric methods have been developed for the area under the ROC curve (AUC), relying on assumptions about the distribution of the prediction scores \citep{hanley1982meaning,faraggi2000effect}. Please refer to supplementary \ref{subsec:details_parametric_CIs} for more details on parametric CI methods.

\paragraph{Non-parametric CI methods.} They can be divided into two categories: bootstrap and concentration inequalities. \textbf{Bootstrap} is a resampling procedure used to assess the variability of estimators and construct CIs without making distributional assumptions. The general idea is to sample with replacement from the observed data to approximate the sampling distribution of the statistic of interest (e.g.mean, median, standard-deviation, accuracy etc.). Each resampled dataset is used to compute a replicate of the statistic of interest, and the distribution of these replicates serves as an estimate of the underlying distribution of the statistic. Quantiles from this empirical distribution are then computed to produce CIs. Bootstrap methods are not specific to a given summary statistics/sample-level metrics and apply to a wide variety of them. Arguably, the most widely known bootstrap methods are {\sl percentile bootstrap} \citep{efron1992bootstrap}, the reverse percentile (also called {\sl "basic"}) bootstrap \citep{efron1981nonparametric}, the bias-corrected bootstrap or the bias-corrected and accelerated ({\sl BCa}) bootstrap \citep{efron1987better}. Please refer to supplementary \ref{sec:bootstrap} for more details. \textbf{Concentration inequalities} are a type of inequalities that aim at describing how much a random variable concentrates around a typical value (e.g. its mean or median) \citep{boucheron2013concentration,abhishek2019introduction}. Among many other applications, they can be used to build CIs. They make only weak distributional assumptions (typically boundedness or sub-normality) and provide a finite-sample guarantee that the coverage is at least $1-\alpha$. A variety of inequalities can be used to construct CIs. We refer to two of them which are widely known and different in spirit, both applicable to bounded random variables. The arguably most common is based on Hoeffding’s inequality \citep{hoeffding1963probability}. However, it is independent of the variability of the data. An alternative is to use inequalities which make use of the empirical variance such as the Empirical Bernstein inequality \citep{maurer2009empirical}.  Please refer to supplementary \ref{sec:concentration} for more details.

\subsection{Related work}
\label{sec:related_work}
\subsubsection{Studies on CI methods comparison in other fields}

In general statistics, there is a huge body of literature on the behavior of CI methods, e.g. \citep{hesterberg2015teachers,chang2024continuity,schall2012empirical,donner1986comparison,brown2001interval}. However, they all rely on parametric simulations except for \citep{schall2012empirical} which only considers the parametric $t$ CI method. In contrast, our work is based on real-world datasets and uses non-parametric simulations allowing us to more faithfully capture the statistical regimes encountered in medical image analysis. Moreover, we evaluate multiple CI methods that are commonly used in practice.

Domain-specific investigations of CIs also exist in fields such as psychology \citep{rousselet2023introduction} and economics \citep{cherry1996comparison}.
Nevertheless, they also rely on parametric simulations and focus on specific statistical regimes that may differ from medical image analysis. In particular, medical imaging studies are often characterized by smaller sample sizes and by the use of specialized performance metrics (e.g., overlap- or distance-based measures) whose statistical properties differ from those of standard binomial metrics commonly used in other fields, potentially leading to different CI behavior.

\subsubsection{CIs in ML}

A large part of the recent literature on uncertainty in ML focuses on uncertainty of individual predictions (i.e., uncertainty associated with model outputs) rather than with aggregate performance measures \citep{abdar2021review,he2025survey}, e.g. using methods based on conformal prediction \citep{campos2024conformal, angelopoulos2023conformal,dewolf2023valid,kompa2021empirical}. Specific papers on this topic have been devoted to medical imaging, e.g. \citep{lu2022fair, kutiel2023conformal, lambert2024trustworthy, huang2024review}.
However, this topic addresses a fundamentally different notion of uncertainty than the one considered in this paper. While prediction-focused studies seek to characterize uncertainty at the level of individual predictions, our focus is on the uncertainty of performance metrics. In particular, we study CIs for aggregate performance estimates (summary statistics), which quantify uncertainty in the performance of a model  rather than in its individual predictions. 

In general ML, several papers have been devoted to variability of learning procedures, e.g.  \citep{schulz2024constructing,agarwal2021deep,bouthillier2021accounting,bayle2020cross,eve2026crossing}. Here, the studied uncertainty is not about the performance of a trained model but about the whole learning procedure, including also variability coming from different training sets, random seed or hyperparameters rather than the uncertainty associated with the performance of a model that has already been trained. Both are important but are distinct problems. In the present paper, we focus on trained models, which correspond to the case of putting a model into ``production", for instance  translation to care or application to clinical trials.

There are also general ML studies focusing on evaluating the performance of trained models. For example, \citep{miller2024adding} is devoted to large language models (LLMs), addressing challenges specific to generative and language-based tasks. However, the regimes of generalist LLMs are very distinct from those of medical imaging AI. Some works deal with both learning procedures and trained models, for example \citet{dietterich1998approximate,raschka2018model}, who study statistical tests for comparing multiple trained models, a related but distinct problem from the performance uncertainty of one model.  \citet{varoquaux2023evaluating} report typical binomial CI values for different values of the accuracy. On  the other hand, we look at the consequences of the specificities of medical imaging studies.

To the best of our knowledge, only a limited number of studies have examined CIs in ML for medical imaging. \citet{christodoulou2024confidence} showed that performance results are often reported without any assessment of variability, potentially leading to misleading claims \citep{christodoulou2025false}. \citet{varoquaux2018cross} investigated performance variability in neuroimaging classification tasks and reported alarmingly large error bars in the sample size regimes typical of the field. Still in neuroimaging, \citet{jurdi2023confidence} evaluated  typical CI widths for segmentation tasks. However, their analysis was limited to two performance metrics, two tasks and one model. Moreover, they did not assess the coverage, which is essential for evaluating the reliability of a CI method. In contrast, we provide a large-scale study of the behavior of CI methods in medical image analysis by using a large dataset spanning multiple tasks, models and metrics, and by studying both the width and the coverage. 

\section{Methods}
\begin{figure}[ht]
    \centering
     \includegraphics[width=\linewidth]{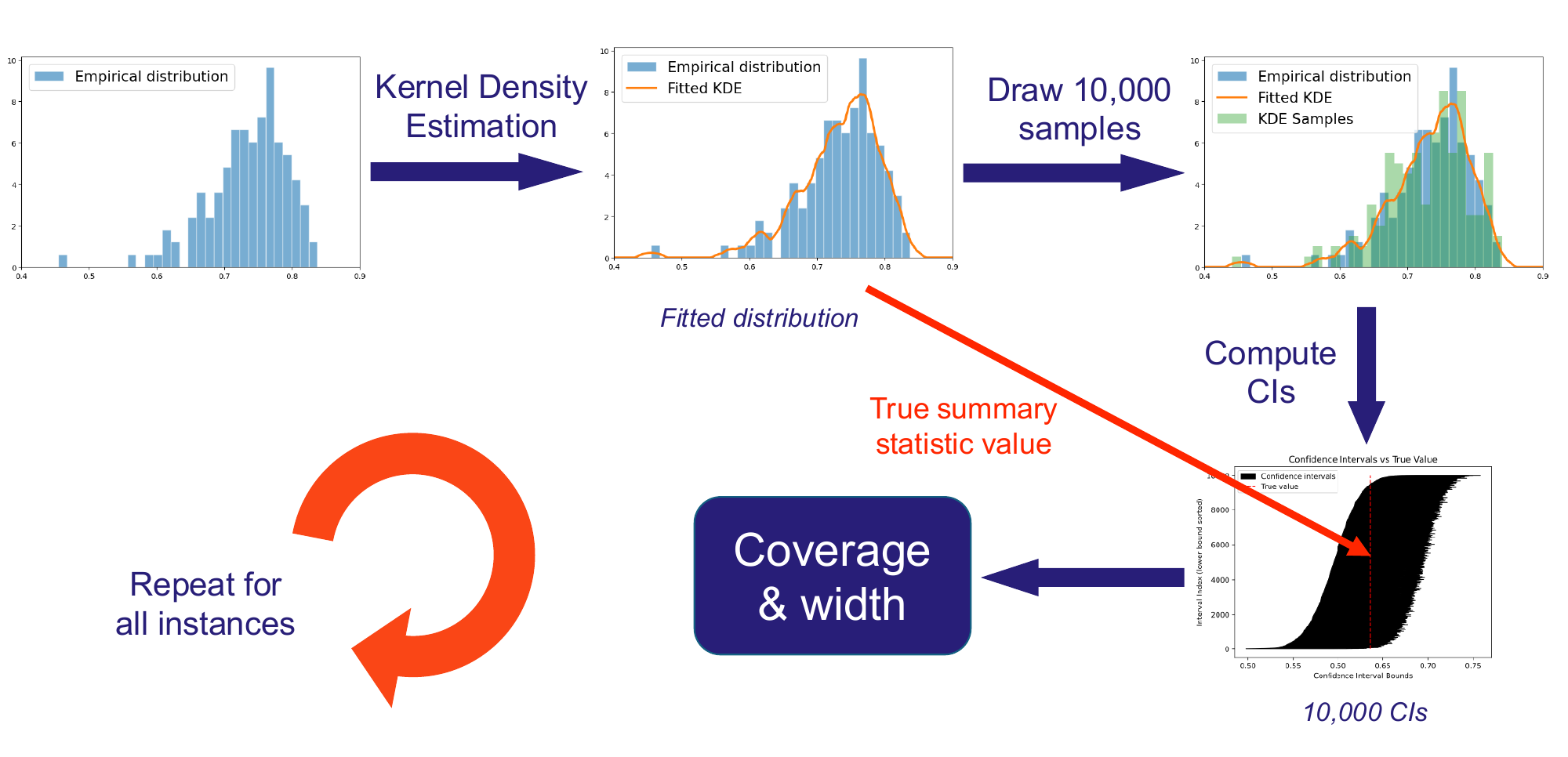}
    \caption{\textbf{Experimental design.} One starts with a real benchmark instance: a given model applied to a given task. A kernel density estimation (KDE) is performed on the empirical distribution (metric for segmentation and logits for classification). Then, multiple (in our case 10,000) samples are drawn from this KDE. A CI is computed for each sample, providing 10,000 CIs. One then gets an empirical coverage value and average interval width. The process is repeated for all benchmark instances, sample sizes and CI methods. This provides the coverage (resp. width) behavior for a given metric and summary statistic. The process is further repeated to perform the full investigation.}
    \label{fig:experiment}
\end{figure}

In this section, we present the methods used to investigate CI behavior. For both segmentation and classification, the general principle is as follows (Figure \ref{fig:experiment}). We started with real benchmark data, comprising 19 ML models applied to 12 different tasks. We call {\sl benchmark instance},  a given model applied to a given task. We thus considered $19 \text{ models} \times 12 \text{ tasks} = 228$ benchmark instances for segmentation and the same number for classification. We then fitted a kernel density estimate (KDE) to the empirical distribution of every benchmark instance. We treated this fitted KDE as the true underlying performance distribution for that instance. From this true distribution, we drew 10,000 synthetic samples at several sample sizes (Supplementary Section~\ref{sec:n_samples} shows that $10,000$ allows to estimate coverage with a margin of error $\varepsilon < 0.01$ at a 95\% confidence level). For each sample, we computed a CI, resulting in 10,000 CIs. These CIs were then used to compute both the coverage (proportion of CIs that contains the true performance value) and the average interval width. The process was repeated for each CI method, each benchmark instance, each performance metric, each summary statistic, each sample size, and each ML problem (classification and segmentation). 

\subsection{Segmentation}

\subsubsection{Experimental data}

We used the Medical Segmentation Decathlon (MSD) \citep{antonelli2022medical} dataset, which includes 16 medical image segmentation tasks. Only tasks with more than 50 test observations (12 tasks) were kept for our study:~\footnote{Below that number, estimating the real distribution was unreliable.} 3 brain tumor magnetic resonance imaging (MRI) tasks, 2 hippocampus MRI tasks, 2 liver computed tomography (CT) tasks, 2 pancreas CT tasks, 1 colon CT task, and 2 hepatic vessel CT tasks. Performance results were available for 19 different ML models, as described in \citet{antonelli2022medical}. Overall, this resulted in 228 segmentation benchmark instances (19 models across 12 tasks).

We evaluated all instances using eight common segmentation metrics: Dice Similarity Coefficient (DSC), Average Symmetric Surface Distance (ASSD), Normalized Surface Distance (NSD), Mean Average Surface Distance (MASD), Hausdorff Distance (HD),  95th percentile Hausdorff Distance (HD95), Intersection over Union (IoU) and Boundary Intersection over Union (Boundary IoU). These metrics cover different aspects of segmentation quality: DSC and IoU measure region overlap, NSD and boundary IoU assess contour overlap, HD and HD95 quantify worst-case contour deviation, and MASD and ASSD capture average contour displacement. In addition, we included clDice, a widely used metric for tree-structured segmentations such as blood vessels or lung airways, which was applied to the two hepatic vessel tasks. Among these metrics, NSD and Boundary IoU require a hyperparameter of tolerance around the border of the predicted and reference masks. We used the same value as in \citet{antonelli2022medical} where these values have been specifically adapted to each organ (Supplementary Section \ref{sec:hyperparameter}). Overall, we covered a wide range of metrics used for different purposes. 

The performance of each benchmark instance was assessed using each metric, yielding a total of $228 \text{ benchmark instance}\times8 \text{ metrics}+ 19 \text{ models}\times2\text{ tasks}\times1 \text{ ClDice} = 1862$ empirical metric distributions.

\subsubsection{Data characterization}

We characterized the diversity of our empirical metric distributions, by studying descriptive statistics: mean, standard-deviation, skewness (asymmetry) and kurtosis (heavy tails) (see Supplementary Section \ref{sec:diversity_segm} for details). Results are shown on Supplementary Figure~\ref{fig:skewness_kurt_segm}. One can appreciate that the values are extremely diverse and extend well beyond the normal range for skewness and kurtosis.

Empirical coverage estimation requires sampling from a distribution for which the true value of the summary statistic is known. In our setting, this distribution corresponds to the segmentation metric for a given benchmarking instance, which requires fitting a separate distribution for each instance before running simulations. To obtain reliable coverage estimates, the simulated data must accurately reflect the empirical behavior of the observed metrics. While prior work, outside medical imaging, typically relied on parametric assumptions for such simulations, e.g.\citep{hesterberg2015teachers,chang2024continuity,donner1986comparison,brown2001interval}, it is unclear whether segmentation metrics actually follow any standard parametric distribution.

To investigate this, we conducted Kolmogorov-Smirnov tests. Each benchmarking instance was tested against a wide range of commonly used parametric families: Normal, Student’s $t$, Skew-Normal, Log-Normal, Logistic, Exponential, Chi-square, Beta (for bounded metrics), Gamma, Pareto, and Weibull distributions. Statistical significance level was set at $p\leq 0.05$ with Bonferroni correction applied to account for multiple testing. Results showed that none of these parametric distributions was an adequate fit (Supplementary Table \ref{tab:distribution-results})~\footnote{To the exception of ASSD and MASD for which Log-Normal may be a reasonable fit. However, to keep consistency across metrics and because a non-parametric fit is able to accommodate a parametric distribution, we also resorted to non-parametric simulations for these metric.}. Therefore, parametric simulations would not reflect the reality and non-parametric simulation approaches are required instead.

\subsubsection{Non-parametric simulations}

In this section, we describe the non-parametric simulation procedure used in our study. 

First, we distinguish between “continuous” and “discrete” metrics. Medical images are discrete by nature, so all segmentation metrics are theoretically discrete. However, most metrics can take such a large range of possible values because images contain a large number of pixels or voxels, that they can, in practice, be seen as continuous. Nevertheless, there are metrics (in our case HD and HD95) which remain discrete because they can only take values from a limited set (1, $\sqrt{2}, \sqrt{3}$, etc.).

For metrics treated as “continuous,” we used kernel density estimation (KDE) to estimate their underlying distributions from finite samples.
KDE builds a continuous distribution by placing a kernel (a probability function) at each data point, with a bandwidth that controls the kernel’s width. While standard KDE uses the same bandwidth for all points, we adopted an adaptive approach in which each data point has its own bandwidth. The final estimated distribution is obtained by averaging the kernels across all data points. Following \cite{wilcox2012introduction}, we used an Epanechnikov (parabolic) kernel with adaptive bandwidths. This choice provides smoothness near zero while remaining bounded, and the adaptive bandwidth helps prevent probability mass from extending outside the valid range of the metric. Details of the algorithm are provided in Supplementary Section \ref{sec:kde}.

For discrete metrics (HD and HD95), we directly sampled from their empirical probability mass functions (PMF). One can notice that PMF sampling is equivalent to performing KDE with a Dirac $\delta$ kernel.

We assessed whether the KDE faithfully represents the original samples (Supplementary Section~\ref{sec:kde_segm_faith}). Results are shown on Supplementary Figure~\ref{fig:kde_faithfulness}. Overall, they indicate that KDE distributions are good approximations of the original samples.

We drew samples of size $n = 10,25,50,75,100,125,150,200,250$, which are typical regimes found in medical image analysis. For instance, a meta-analysis of MICCAI 2023 segmentation papers found that the median sample size was 62 (Q1: 25, Q3: 223) \citep{christodoulou2025false}.

\subsubsection{Considered summary statistics}
Because segmentation metrics are computed at the individual-case level, a summary statistic (e.g. the mean) is necessary to report performance at the sample level. We assessed 5 commonly used summary statistics.
For central tendency, we considered the mean, as it is the most widely reported statistic; the median, which is more robust to outliers and a standard robust choice; and the trimmed mean, recommended by \cite{agarwal2021deep}, which offers a flexible compromise between the mean and the median by reducing the influence of extreme values. In addition, we included two measures of dispersion: the standard deviation (STD) and the interquartile range (IQR), where the IQR serves as a robust, order-statistic-based alternative to STD that is less sensitive to outliers.

\subsubsection{Considered CI methods}
We studied five CI methods, which we perceive to be particularly frequently used and are implemented in SciPy \citep{virtanen2020scipy}, considering that our community is mostly Python based \citep{eisenmann2022biomedical}. We studied two parametric methods for the mean: parametric $t$ and $z$ (see formulas in Supplementary Section~\ref{subsec:details_parametric_CIs}). We studied three bootstrap methods for each considered summary statistics: percentile, basic and BCa (see Supplementary Section~\ref{sec:bootstrap}). For bootstrap, we used 9,999 resamplings which is the default in SciPy~\footnote{The use of an odd number of resamplings comes from the fact that most bootstrap-based methods rely on quantiles of the bootstrap distribution. Performing an odd number of resamplings enables classical quantiles to correspond to actual resamplings, rather than being the average of two values.}. We used the SciPy default number of resamplings not only because it leads to good estimates but also because most users won't change this value\footnote{The ideal number will in fact depend on the problem at hand and there is no universal recommendation. \citep{hesterberg2015teachers} recommends $10,000$ for routine use. \cite{rousselet2023introduction} state that $1,000$ is often enough but that, given the low cost of the computation, there is usually no good reason not to go up to $10,000$. Further, using formula (9.5) in \cite{efron1987better}, one can show that 9,999 resamplings allows to estimate the distribution with a $1.28\%$ error.  Overall, $9,999$ appears as a good default choice. One could consider a lower number if dealing with extremely large datasets and very costly metrics/summary statistics.}.

We also assessed  concentration inequalities because they are well-known in the ML community and offer finite-sample guarantees. They are straightforward to implement. Specifically, we computed Hoeffding’s-based and Empirical Bernstein CIs (see Supplementary Section~\ref{subsec:concentration_extended} for details). 

% See Supplementary Section \ref{sec:CIs} for details about these methods.

\subsection{Classification}

\subsubsection{Experimental data}
Our benchmark dataset is based on \cite{godau2024beyond} and includes 12 classification tasks across diverse imaging modalities: fundus photography \citep{cataract_dataset, APTOS2019BlindnessDetection, liu2022deepdrid}; confocal laser endomicroscopy \citep{ghatwary2019vivo}; computed tomography \citep{yang2020covid}; dermoscopy \citep{kawahara2018seven, rotemberg2021patient}; gastroscopy and colonoscopy \citep{borgli2020hyperkvasir}; sonography \citep{al2020dataset}; and X-ray \citep{rajpurkar2017mura, irvin2019chexpert}. Some of the datasets are imbalanced and multiclass\footnote{Multiclass classification is also sometimes called exclusive label classification. The non-exclusive labels case is not considered as it consists of multiple binary classification tasks performed simultaneously and tested separately, binary data being already included in our dataset.
}. Table~\ref{tab:classif} provides the total sample size, the number of classes (ranging from 2 to 6) and the sample size per class for each task. Similarly to segmentation, we evaluated 19 trained ML models.

We considered the following commonly used classification metrics: Accuracy, Balanced accuracy (BA), F1 score, Average Precision (AP), Area Under Receiver Operating Characteristic curve (AUC) and Matthews Correlation Coefficient (MCC). 

Classification metrics are sample-level metrics, similarly to summary statistics in segmentation. Thus, there is no notion of test set empirical distribution, only a sampling distribution. Confusion-matrix-based metrics (accuracy, F1 score, balanced accuracy etc.) only require binary labels and their sampling distribution could have been simulated using binomial or multinomial true distributions. However, threshold-based metrics such as AUC and AP require the actual post-softmax scores computed from the model outputs (logits). 
For consistency, we computed all metrics from the raw logits. For the confusion-matrix-based-metrics, we used softmax normalization, followed by an argmax to obtain the final binary labels. Then the main difference with segmentation is that we will fit the logit distributions, while we fitted the individual metric distributions for segmentation. We will then sample and compute classification metrics for each sample (similarly to summary statistics for segmentation).

\subsubsection{Data characterization}
Again, we studied the diversity of the empirical distributions of the logits by analyzing the multivariate skewness and kurtosis \citep{mori1994multivariate}. Supplementary Figure \ref{fig:skewness_kurt_classif} shows a wide variety of performance values, skewness and kurtosis and that, unsurprisingly, skewness and kurtosis are dependent on the number of classes present in the data.

In line with our approach for segmentation, we tested whether the empirical logits distributions followed some common parametric distributions using Kolmogorov-Smirnov tests. Since the test is defined only for one-dimensional distributions, we evaluated multivariate distributions through their marginals: for example, if data follow a multivariate normal distribution, all marginals must be Gaussian; conversely, if any marginal is non-Gaussian, the joint distribution cannot be multivariate Gaussian. This reasoning extends to other distribution families as well. We tested the logits against various parametric families, namely Normal, Student, Skewnorm, Logistic and Pareto families. Significance level was Bonferroni corrected. The results showed that none of these parametric distributions provided an adequate fit (Supplementary Table \ref{tab:distribution-results-classif-bonferonni}), indicating that non-parametric simulations of the logits is required.

\subsubsection{Non-parametric simulations}
We performed multivariate KDE on the logits following the same algorithm as for segmentation but with a multivariate kernel (see Supplementary Section \ref{sec:kde}). We assessed whether the KDE preserves the properties of AUC and AP across different number of classes (Supplementary Section~\ref{sec:kde_class_assessment}). Results are shown on Supplementary Figure~\ref{fig:kde_curve_preservation}. These results show that the KDE preserves the properties of AUC and AP, independently of the number of classes.

We drew samples of sizes $n = 10,25,50,75,100,125,150,200,250$ to be consistent with segmentation. Furthermore, we also studied larger samples ($n=500,1000,5000$), as these are more common in classification settings \citep{christodoulou2025false}. However, due to prohibitive computational cost, we could only do this for some of the performance metrics. Specifically, for Macro BA and Macro F1, we also sampled $n=500$, $1000$ and $5000$. For Macro AUC, we also sampled $n=500$, and $1000$. To further reduce the computational costs, we used $3,500$ samples
instead of $10,000$ and $3,499$ bootstrap resamplings instead of $9,999$.

\subsubsection{Considered multiclass aggregation methods}
\label{sec:aggregation}
In the multiclass setting, we considered two common metric aggregation strategies. Micro-averaging treats a task with $d$ classes as $d$ binary classification tasks; labels are one-hot encoded and the metric is computed over the resulting $n\times d$ binary sample. In macro-averaging, one computes the metric separately for each of the different classes, then takes the average of the metric across all classes.

Note that accuracy and balanced accuracy can be interpreted as recall under different aggregation schemes: accuracy corresponds to micro-averaged recall, while balanced accuracy corresponds to macro-averaged recall. As a result, these metrics are intrinsically micro- and macro-aggregated, respectively. The MCC does not require an aggregation method as it has a multi-class generalization.

\subsubsection{Considered CI methods}

Again, we selected commonly used methods, namely those which are available through standard Python implementations: Scipy \citep{virtanen2020scipy} or statsmodels \citep{seabold2010statsmodels}). For parametric CIs of proportions, we evaluated the four methods implemented in statsmodels: Wald, Wilson, Agresti-Coull, and Clopper-Pearson. Note these parametric methods are valid only for binomial proportion metrics (thus only for the accuracy in our study). In addition, we considered bootstrap-based intervals using the BCa, basic, and percentile methods provided by SciPy. These are applicable to all classification metrics.

\subsection{Coverage convergence pace (CCP)}

Empirical coverage for two-sided intervals converges to theoretical coverage at typical rate $1/n$ \citep{hall1992bootstrap}.
We call the first-order coverage constant associated with this rate in the coverage expansion “coverage convergence pace” (CCP), as a higher constant usually implies slower convergence. See Supplementary Section \ref{sec:ccp_def} for more details.

\subsection{Statistical analysis}
 We tested the claims in the results section using permutation tests. More specifically, we tested our claims on either the coverage or the width across all sample sizes. When not explicitly stated otherwise, we performed the tests for CIs of the mean using percentile method, as they are the most common use cases. 
 We corrected for multiple testing using the False Discovery Rate (FDR) \citep{benjamini1995controlling} across all our claims and sample sizes, with a total of 3078 tests.
 Details of each test can be found in Supplementary Section \ref{sec:tests}.
\subsection{Implementation and computational details}

The code is available on Github: \url{\githubrepocode}.

Segmentation metrics were computed using the Metrics Reloaded~\citep{maier2024metrics} software package\footnote{\url{https://github.com/Project-MONAI/MetricsReloaded}} from MONAI \citep{cardoso2022monai}. 
Some classification metrics were computed using scikit-learn \citep{pedregosa2011scikit}.   We had to re-implement some of the metrics like AUC or AP, as their scikit-learn implementation supports many edge-cases tests which were not necessary in our setting but added a lot to the computation time. Re-implementations were unit tested against the original functions to ensure validity.

 For flexibility reasons, we reimplemented the bootstrap methods from Scipy \citep{virtanen2020scipy}, following the exact same algorithms and performing unit tests against the original version to assess validity.

All computations were performed on the French supercomputer Jean Zay. The cluster contains a variety of GPU and CPU nodes, but we only used CPU nodes. There are 720 nodes for CPU usage, each composed of 2 Intel Xeon Gold 6248 20C 2.5GHz, along with 192Go of RAM. This means each node contains 40 cores, with 3.9Go of RAM per core. One hour of computation time represents 1 core used for 1h. 

\section{Results}
\label{sec:results}
In sections below, we highlight the main results regarding the behavior of CI methods and their relationships with characteristics of medical imaging study.

The full results, across all tasks, metrics and summary statistics, span over 108 graphs: 9 metrics $\times$ 5 summary statistics $\times$ 2 (coverage and width) for segmentation, 9 couples (metrics, aggregation method if needed) $\times$ 2 (coverage and width) for classification. All these are displayed in Supplementary Section \ref{sec:full_results} (Supplementary Figures \ref{fig:cov_segm_all} to \ref{fig:width_classif_all})~\footnote{To ease zooming, all figures are also available as separate PDF files at \url{\githubrepofig}.}. 

All our claims have been statistically tested and the results can be found in Supplementary Section \ref{sec:tests}.

\subsection{Behavior of different CI methods}

\begin{figure}[ht]
    \centering
    \includegraphics[width=\linewidth]{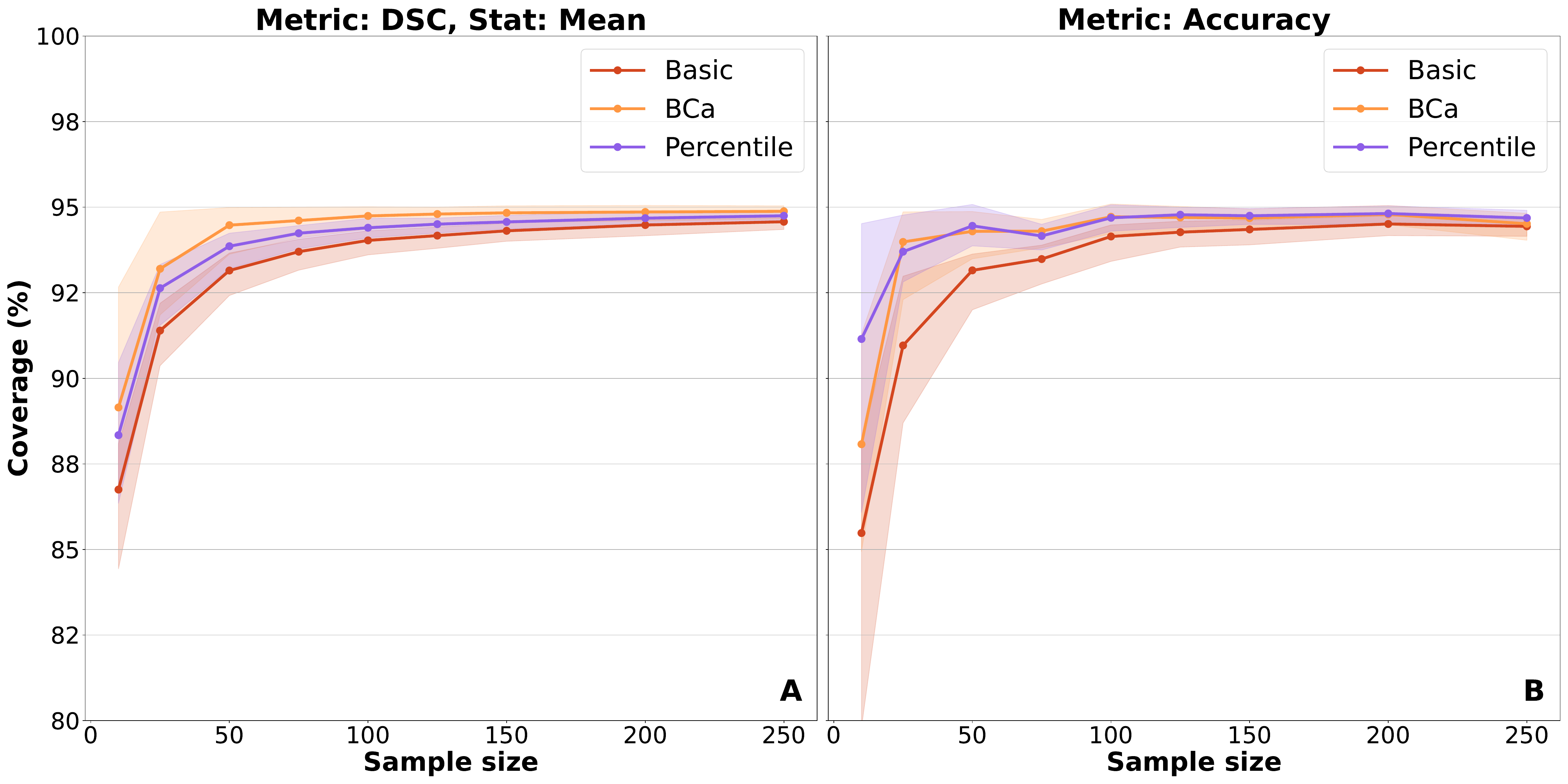}
    \caption{\textbf{Comparison of different bootstrap methods.} Coverage of CIs of the mean DSC (Panel A) and of the accuracy (Panel B) for different bootstrap CI methods. 
    Other settings (summary statistics and metrics) are available in Supplementary Section \ref{sec:full_results}.}
 \vspace{0.2ex}
   {\scriptsize \sl \linespread{0.2}\selectfont  Solid lines show the median across benchmark instances, with Q1–Q3 shown as a shaded region, this applies to other figures unless stated otherwise.
   \par}
   {\scriptsize \sl \linespread{0.2}\selectfont  The target coverage is 95\%, this applies to other figures unless stated otherwise.
   \par}
    \label{fig:fail_basic}
\end{figure}

\paragraph{Basic underperforms compared to the other bootstrap methods in a vast majority of cases.} Basic is most often the worst in terms of coverage. This can be seen on Figure \ref{fig:fail_basic} for CIs of the mean DSC and accuracy. This also holds for other settings, as one can see in Supplementary Section \ref{sec:full_results}. One exception is macro-aggregated AP where basic performs better than other methods (see Section~\ref{sec:consequences}).

\paragraph{BCa has a good behavior in many cases, but fails dramatically for the median.} The median represents the most striking failure case. This behavior is illustrated for the DSC in Figure \ref{fig:fail_bca} but the problem holds across all metrics as shown in Supplementary Figure \ref{fig:bca_all}. For instance, this is even more catastrophic for HD, HD95, ASSD and MASD. As one can see,  the coverage of the BCa interval decreases for large values of $n$.
We can also see this behavior holds for a few other cases, such as the IQR of the HD, as shown in Supplementary Figure \ref{fig:cov_segm_all}.
On the other hand, BCa has excellent coverage for the mean across all metrics (Supplementary Figure \ref{fig:cov_segm_all}, first row). Its coverage is better than percentile's in very small samples but they quickly become comparable.

\begin{figure}[!ht]
    \centering
    \includegraphics[width=0.6\linewidth]{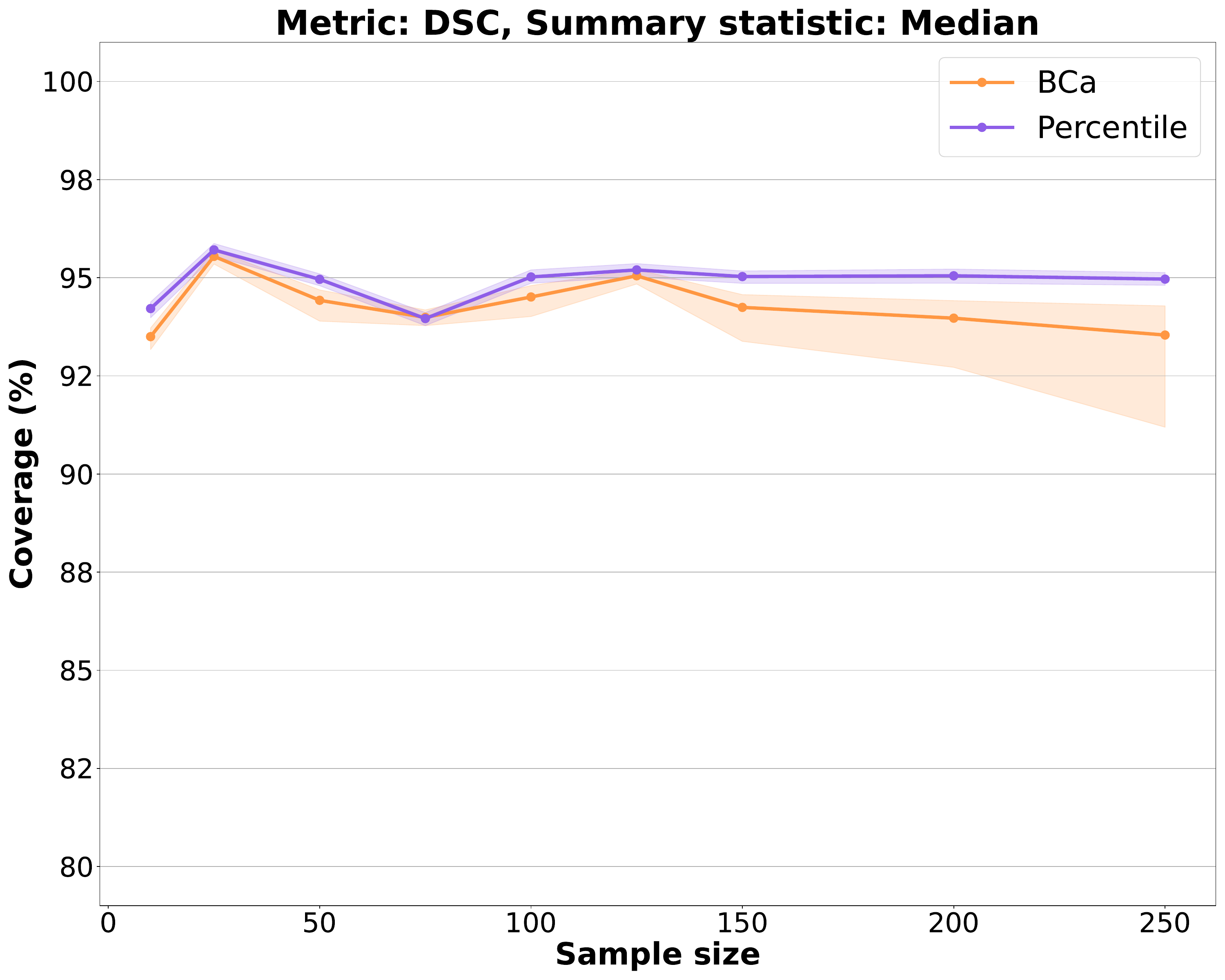}
    \caption{\textbf{BCa fails for CIs of the median.} BCa’s coverage (orange) decreases with increasing sample size while that of percentile (purple) is adequate. Graphs for other segmentation metrics are shown in Supplementary Figure \ref{fig:bca_all}.}
    \label{fig:fail_bca}
\end{figure}

\begin{figure}[!htp]
    \centering
    \includegraphics[width=0.8\linewidth]{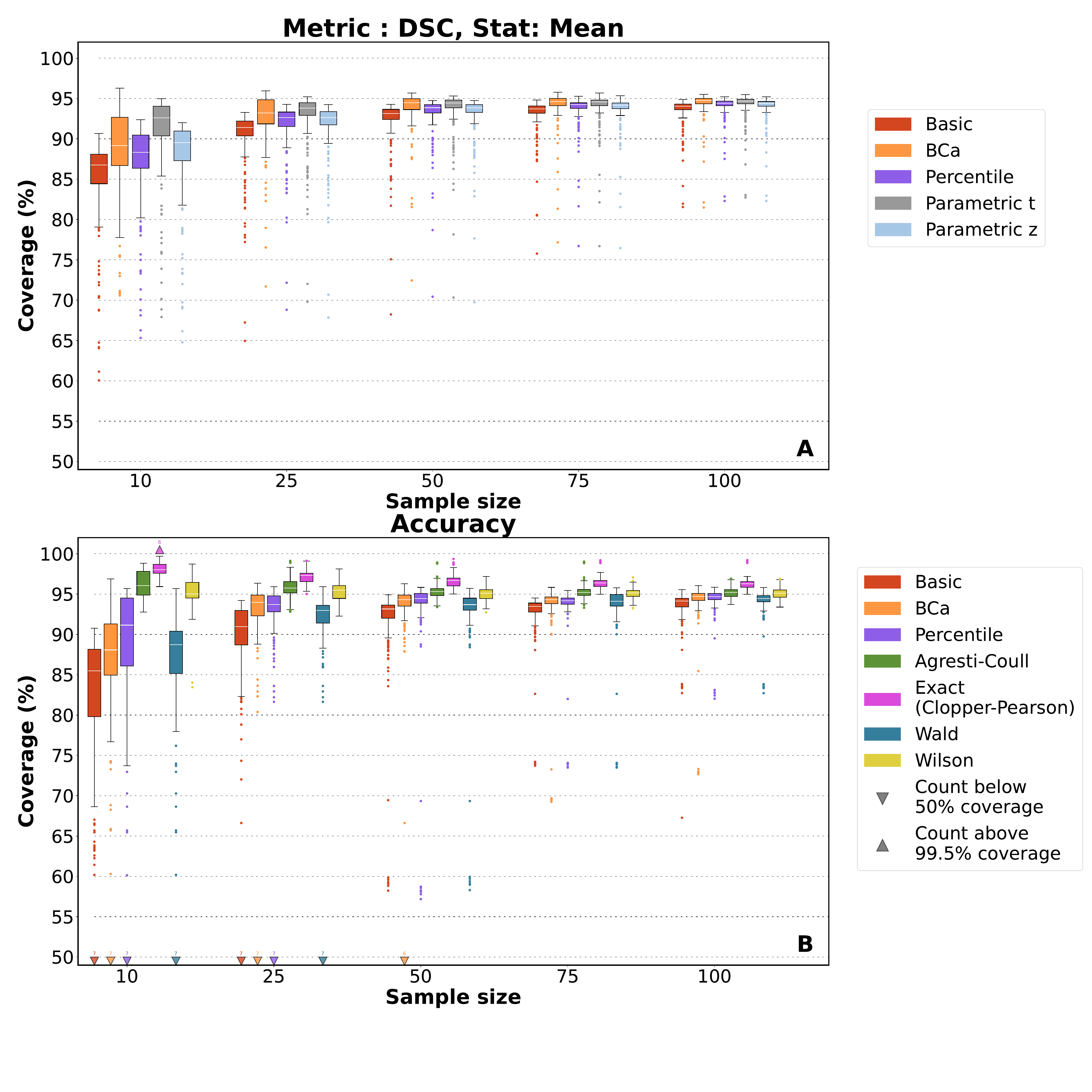}
    \caption{\textbf{Some parametric methods outperform bootstrap on very small samples but this advantage quickly disappears.} Coverage values for CIs of the mean DSC and of the accuracy for multiple sample sizes. 
    Panel A: parametric $t$ and $z$, BCa, basic and percentile bootstraps are shown for the mean of the DSC. Parametric $t$ is the best method and outperforms bootstrap when $n=10$. 
    Panel B: Agresti-Coull, exact, Wald, Wilson, BCa, basic and percentile bootstraps are shown for accuracy. Wilson is the best method and outperforms bootstrap when $n\leq25$.
    Results for other segmentation metrics are available in Supplementary \ref{sec:full_results}.}
      \vspace{0.2ex}
   {\scriptsize \sl \linespread{0.2}\selectfont  Across the paper, boxplots display the median and interquartile (IQR) range (Q1-Q3) across benchmark instances, with whiskers extending to the most extreme values within 1.5 times the IQR range. Points represent individual benchmark instances.
   For coverage plots, if there are instances below $50\%$ or above $99.5\%$ coverage, their number is indicated above a triangle. 
   \par}

    \label{fig:param_small_samples}
\end{figure}

\paragraph{Percentile is robust across a very wide range of settings.}
Its asymptotic behavior is always adequate, which is not always the case for BCa (see the case of the median above). Compared to basic, its coverage is far better in a majority of cases. In other cases, their coverage are comparable, except in one case:  macro-aggregated AP. Nevertheless, it is important to note that, while percentile is more robust, BCa has most often a better coverage for CIs of the mean, in particular when working with small samples and skewed metrics (see Section~\ref{sec:consequences}). BCa also has a better coverage for some classification metrics when sample size is small (AUC, F1).

\begin{figure}[!ht]
    \centering
    \includegraphics[width=0.8\linewidth]{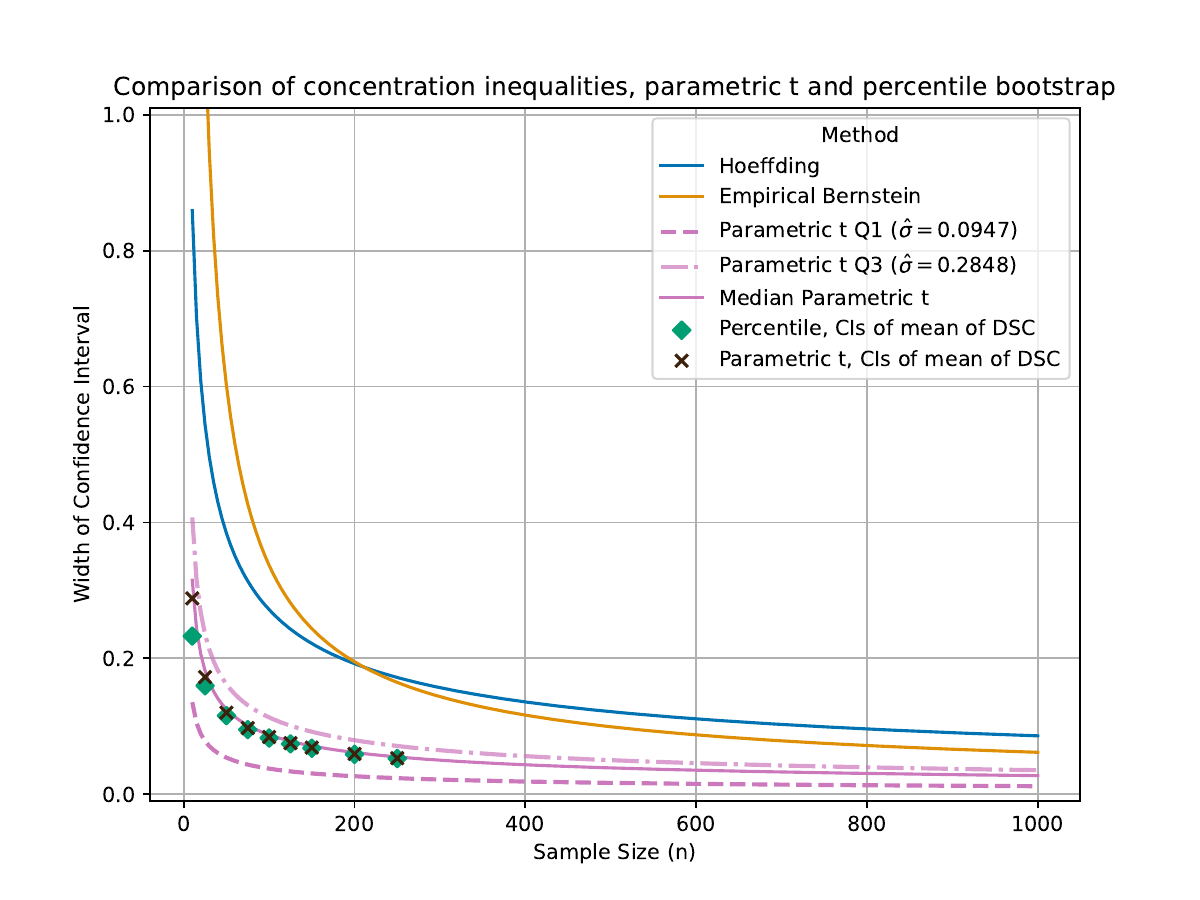}
    \caption{\textbf{Concentration inequalities produce very conservative intervals.} Comparison of the widths for various CI methods: Hoeffding's, parametric $t$ and percentile bootstrap.  “Typical parametric $t$” corresponds to the width obtained with the median standard deviation (STD)  across all 228 benchmark instances for the DSC. Widths for Q1 and Q3 STD are shown in dotted lines. Real data points (percentile and parametric $t$) were added to the plot, to compare them with theoretical values.}
    \label{fig:hoeffding_paramt}
\end{figure}

\paragraph{Some parametric approaches outperform bootstrap for very small samples but this advantage vanishes rapidly.} This behavior is illustrated in Figure \ref{fig:param_small_samples}. For the mean, the parametric $t$ interval exhibits substantially better coverage than bootstrap methods for very small sample sizes. For instance, for the mean DSC with $n=10$ the coverage is $92.5\% $ for parametric $t$ compared to $88\%\text{-}89\%$ for bootstrap methods (Figure \ref{fig:param_small_samples} Panel A).
This pattern is consistent across all segmentation metrics (Supplementary Figure \ref{fig:cov_segm_all}, upper row). As shown in Supplementary Figure \ref{fig:width_segm_all} (upper row), this advantage arises because the parametric $t$ intervals are wider, due to Student’s $t$ correction.
However, this benefit rapidly vanishes as the sample size increases: BCa is as reliable as parametric $t$ for $n\geq50$ (even sometimes for $n=25$) and percentile is not far behind. For accuracy, the Wilson interval is advantageous for very small samples (Figure \ref{fig:param_small_samples} Panel B). For $n=10$, its median coverage is close to 95\% while percentile is at $91\%$ and BCa at $88\%$. According to statistical tests (Supplementary Figure~\ref{fig:test_param_boot_classif}), Wilson is significantly better than bootstrap methods for all sample sizes. But as can be seen on Figure \ref{fig:param_small_samples}, the difference is meaningful only for sample sizes smaller than 25. In this case, the improved coverage is not driven by increased interval width (Supplementary Figure \ref{fig:width_classif_all}). Note nevertheless, that for $n=10,$ the median width is huge (about 0.5) for classification and still very large (about 0.2) for segmentation. As expected, the Wald interval performs poorly \citep{brown2001interval}. The exact (Clopper–Pearson) interval yields systematic overcoverage.

\paragraph{Concentration inequalities produce very imprecise CIs.} CIs from  concentration inequalities are way wider than parametric or bootstrap ones (Figure \ref{fig:hoeffding_paramt}). Hoeffding’s width is typically over 3 times larger than that of parametric $t$ or percentile bootstrap (median scenario across benchmark instances). When taking $Q3$ of benchmark instances, the width is still twice as large. When SD of the metric decreases, things get even worse (e.g. 7 times larger for $Q1$). As for Empirical Bernstein, CIs are even wider than Hoeffding’s at moderate $n$. On the other hand, it is much better than Hoeffding's when $n$ is large. This behavior starts at $n\approx200$ for the median scenario. However, it is still over twice as wide as parametric $t$ at $n=2,000$ and $1.51$ times wider when $n\rightarrow \infty$ (Supplementary Figures~\ref{fig:concentration_10_5} and~\ref{fig:concentration_10_9}, Supplementary Table~\ref{tab:concentration_summary}). More details are provided in Supplementary Section~\ref{subsec:concentration_extended}.

\subsection{Classification requires much more observations than segmentation}

\begin{figure}[!ht]
    \centering
    \includegraphics[width=0.8\linewidth]{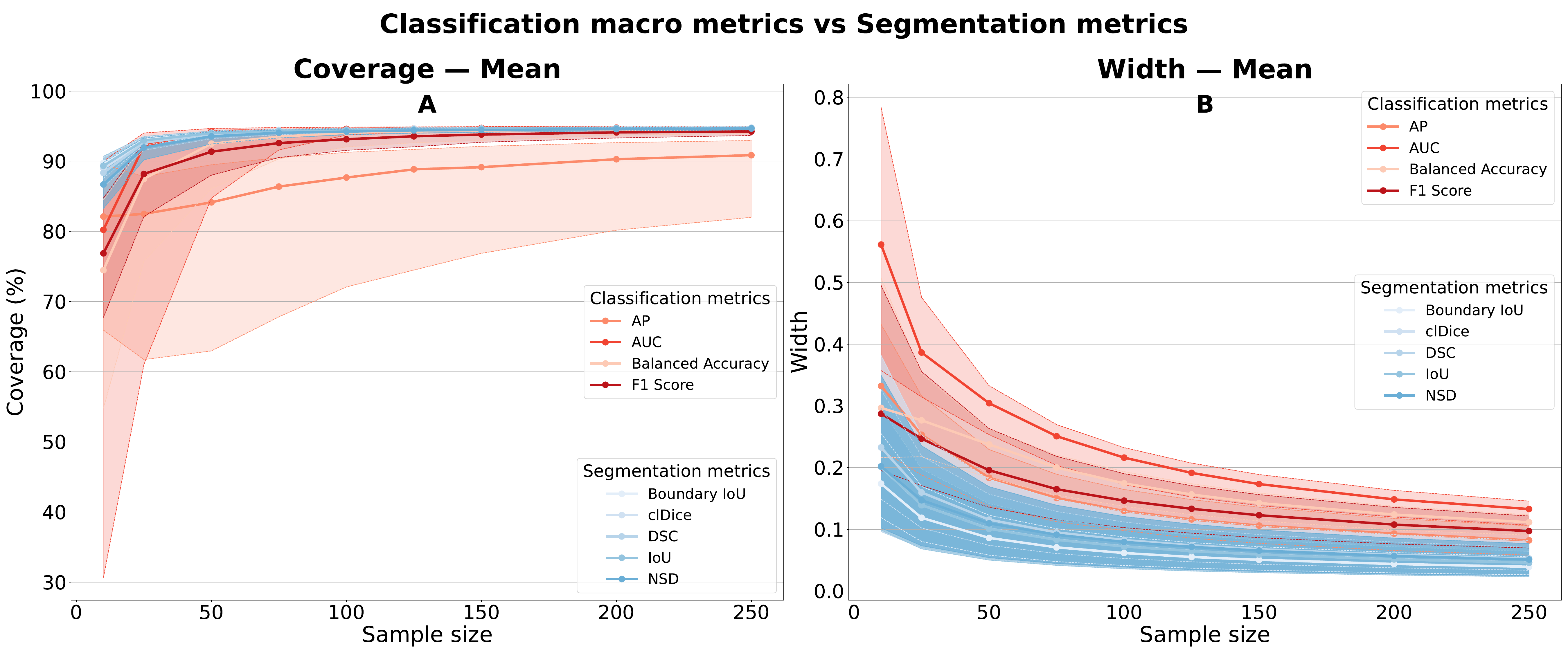}
     \includegraphics[width=0.8\linewidth]{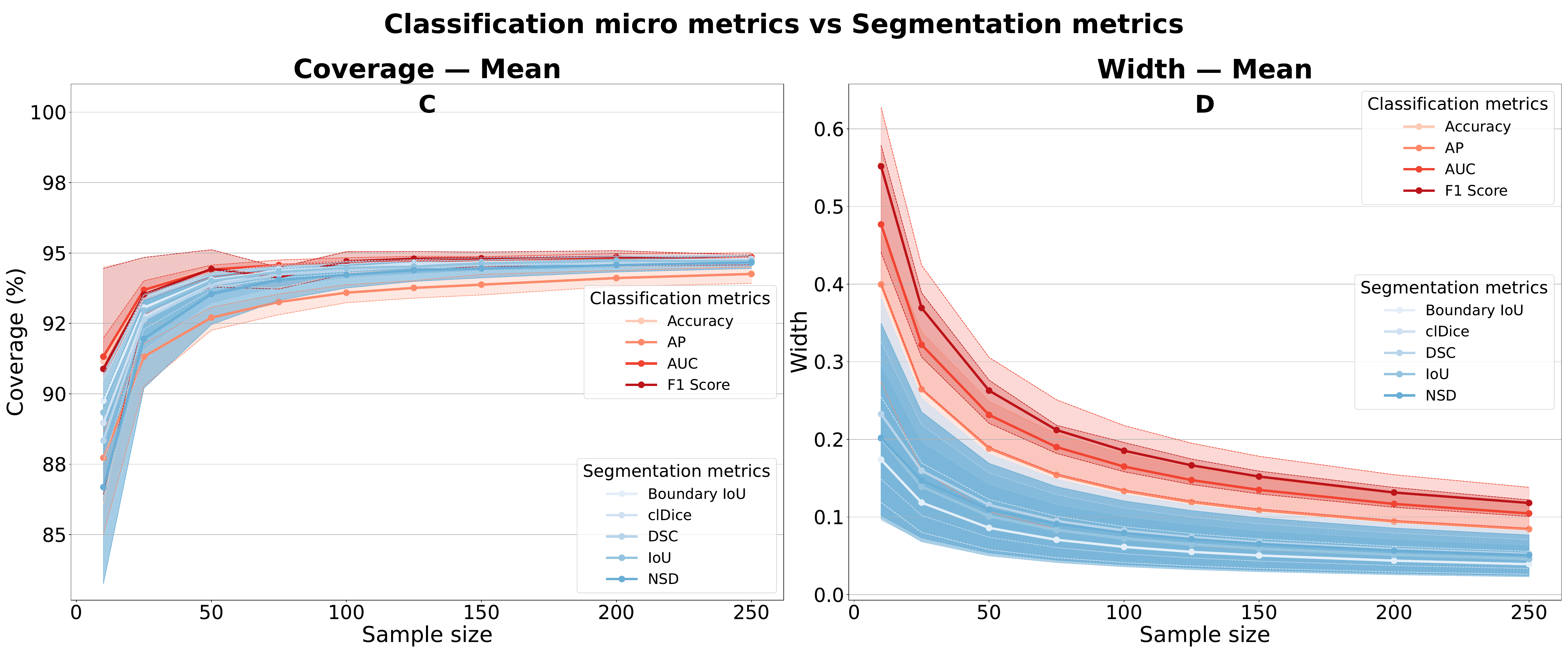}
    \caption{\textbf{Classification requires larger samples than segmentation to get reliable and precise CIs.} CIs for classification metrics (red) are compared to CIs for the mean of bounded segmentation metrics (blue) distributions, in terms of coverage (panels A for macro and C for micro) and width (panel B for macro and D for micro). CIs are considerably wider for classification (both macro and micro) than for segmentation. Coverage converges more slowly for classification macro than for segmentation but no such difference is observed for classification micro. Other summary statistics of segmentation metrics are shown on Supplementary Figure~\ref{fig:segm_classif} and~\ref{fig:segm_classif_micro}.}
    \label{fig:classif_segm}
\end{figure}

Coverage converges more slowly for classification macro than for segmentation (Figure \ref{fig:classif_segm}, Panel A). The difference was statistically significant for all metrics (Supplementary Figure \ref{fig:test_segm_classif}). On the other hand, there was no such trend for coverage for segmentation vs classification micro (Figure \ref{fig:classif_segm}, Panel C). Importantly, CIs are wider for classification than for segmentation (Figure \ref{fig:classif_segm}, Panels B and D), and this holds for macro and micro. In other words, much larger samples are needed to achieve a given width for classification than for segmentation. For instance,  to achieve a width of 0.1, one typically needs $\sim 50$ observations for segmentation and $\sim~250$ for classification. 

\begin{figure}[!ht]
    \centering
    \includegraphics[width=0.9\linewidth]{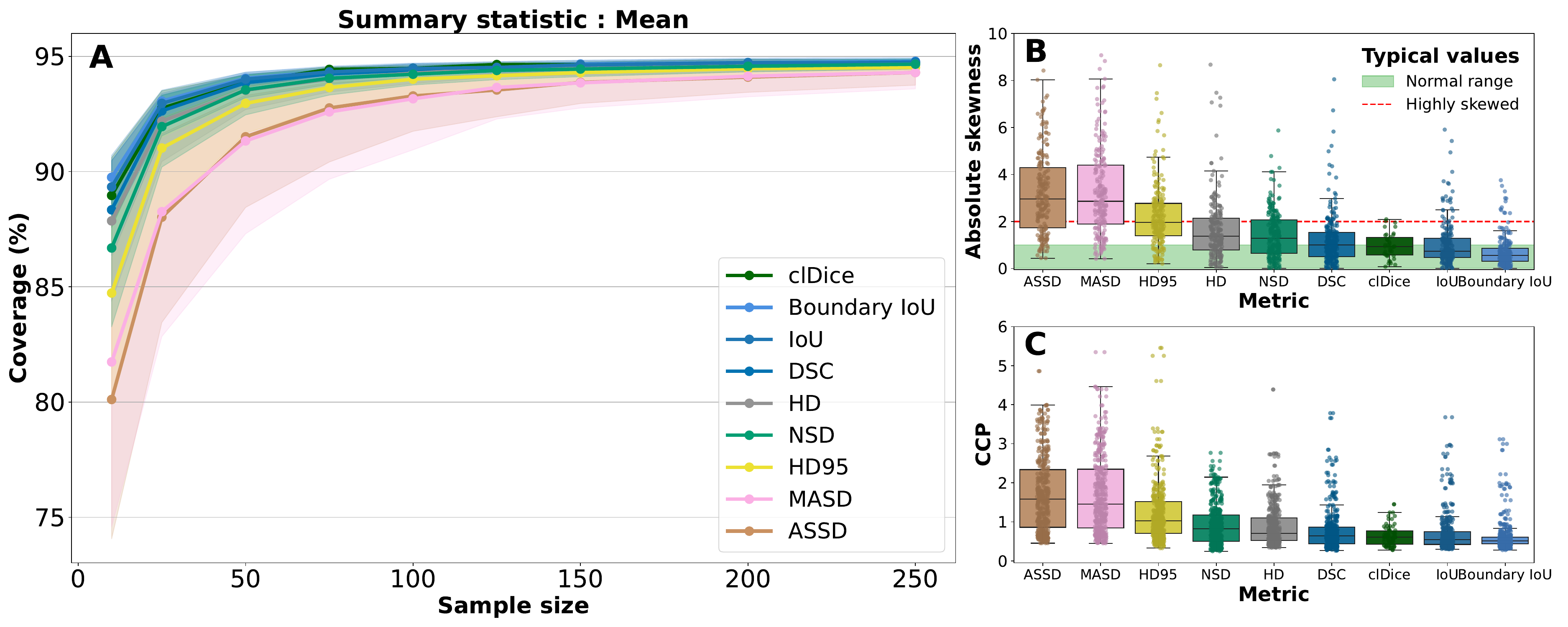}
    \caption{\textbf{Overlap-based metrics converge faster than distance-based metrics.} Panel A: Coverage for CIs of the mean of different segmentation metrics using percentile method. The coverage of ASSD and MASD converge slower to 0.95 than that of other metrics. Panel B: Skewness values by metric, one point is the skewness of one benchmark instance. Panel C: Coverage convergence pace (CCP) values by metric for CIs of the mean using percentile bootstrap, one point is the CCP of one benchmark instance. 
    One observes very similar patterns for both the skewness and the convergence pace.
    The graph for other CI methods and summary statistics is in Supplementary Figure~\ref{fig:cov_metrics_segm}. }
    \label{fig:metrics}
\end{figure}

\subsection{The consequences of metric choice}
\label{sec:consequences}
\paragraph{The choice of metrics markedly impacts  the difficulty to quantify their estimation noise.} Not all metrics behave equally: some require substantially larger sample sizes to achieve reliable coverage (Figure \ref{fig:metrics}, Panel A). Visually, ASSD and MASD exhibit the slowest convergence of empirical coverage for CIs of the mean toward 95\%. These differences are statistically significant as shown in~Supplementary Figure~\ref{fig:test_segm}. Interestingly, the metrics with the slowest convergence are also the most skewed and correspond to the distance-based metrics (Figure \ref{fig:metrics}, Panel B). To quantify the relationship between skewness and convergence pace across our non-independent benchmark instances, we fitted a linear mixed-effects model with convergence pace as the response variable and skewness as the explanatory variable, where the fixed effects are the skewness and the random effects the groups (task, metric). Following the methodology described in \citet{nakagawa2013general}, the model yielded an 
$R^2$ value of $0.48$, indicating that skewness explains approximately $48\%$ of the variance in convergence pace. The corresponding R value is $0.69$ (CI : $[0.58, 0.78]$) and can be interpreted as a correlation between the skewness and the CCP. This result is consistent with the rankings shown on Figure \ref{fig:metrics}, Panels B/C. 

\begin{figure}[H]
    \centering
   \includegraphics[width=\linewidth]{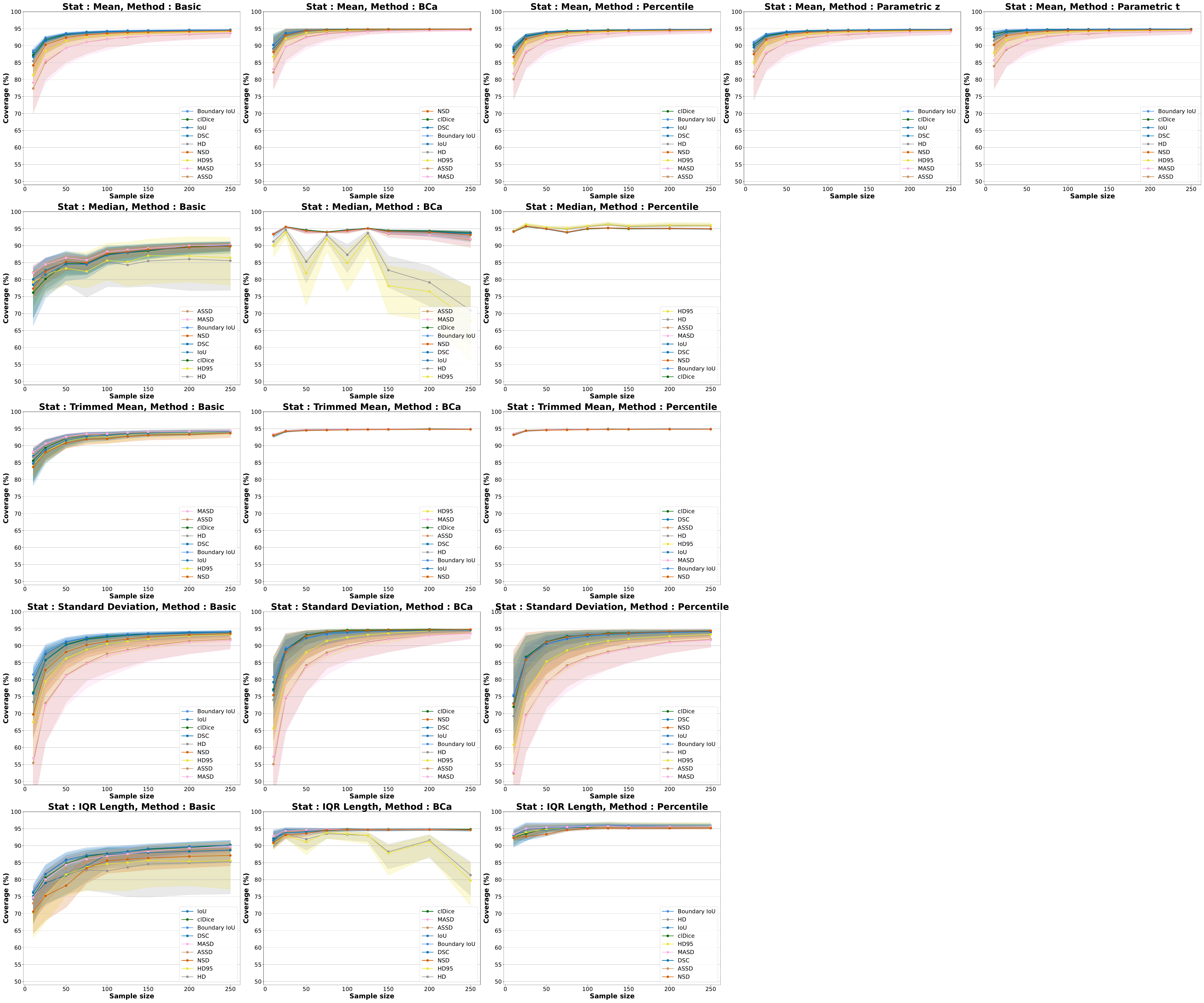}
    \caption{\textbf{Coverages for all CI methods and all summary statistics.} Differences in convergence pace hold across all CI methods for the mean and for STD. On the other hand, no consistent differences are observed for robust statistics (median, IQR, trimmed-mean). For median and IQR, one observes the failures of BCa and basic which were previously reported.}
    \label{fig:cov_metrics_segm}
\end{figure}

Supplementary Figure~\ref{fig:cov_metrics_segm} shows that the differences in CCP hold across all CI methods for the mean. They also hold when the summary statistic is STD but not for robust statistics (median, IQR, trimmed mean).

\paragraph{For macro-aggregated AP, we observe unexpected results.}
Percentile and BCa converge particularly slowly while this is not the case for basic bootstrap (Supplementary Figure~\ref{fig:cov_classif_all}). This better behavior of the basic bootstrap is an exception as it was not observed across all other 53 coverage graphs.

\subsection{The metric aggregation method impacts the coverage of CIs}
\begin{figure}[!ht]
    \centering
    \includegraphics[width=0.8\linewidth]{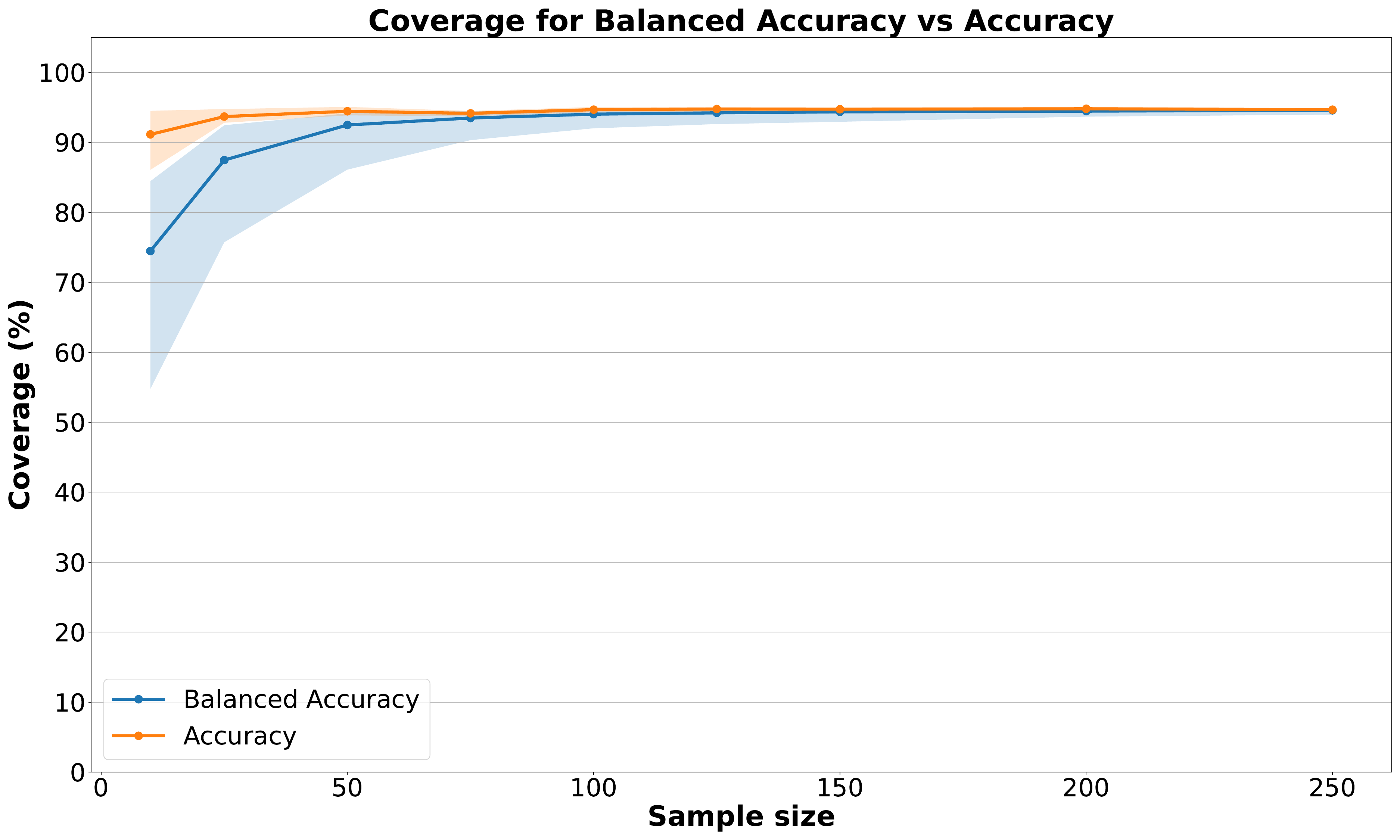}
    \caption{\textbf{The aggregation method for classification metrics impacts the coverage.} Accuracy (micro averaging) converges at a faster pace than balanced accuracy (macro averaging), for percentile method. 
The same observation holds for other comparisons between micro- and macro-aggregated metrics as shown in Supplementary Figure \ref{fig:micro_macro_all}.}
    \label{fig:micro_macro}
\end{figure}

\begin{figure}[H]
    \centering
    \includegraphics[width=0.75\linewidth]{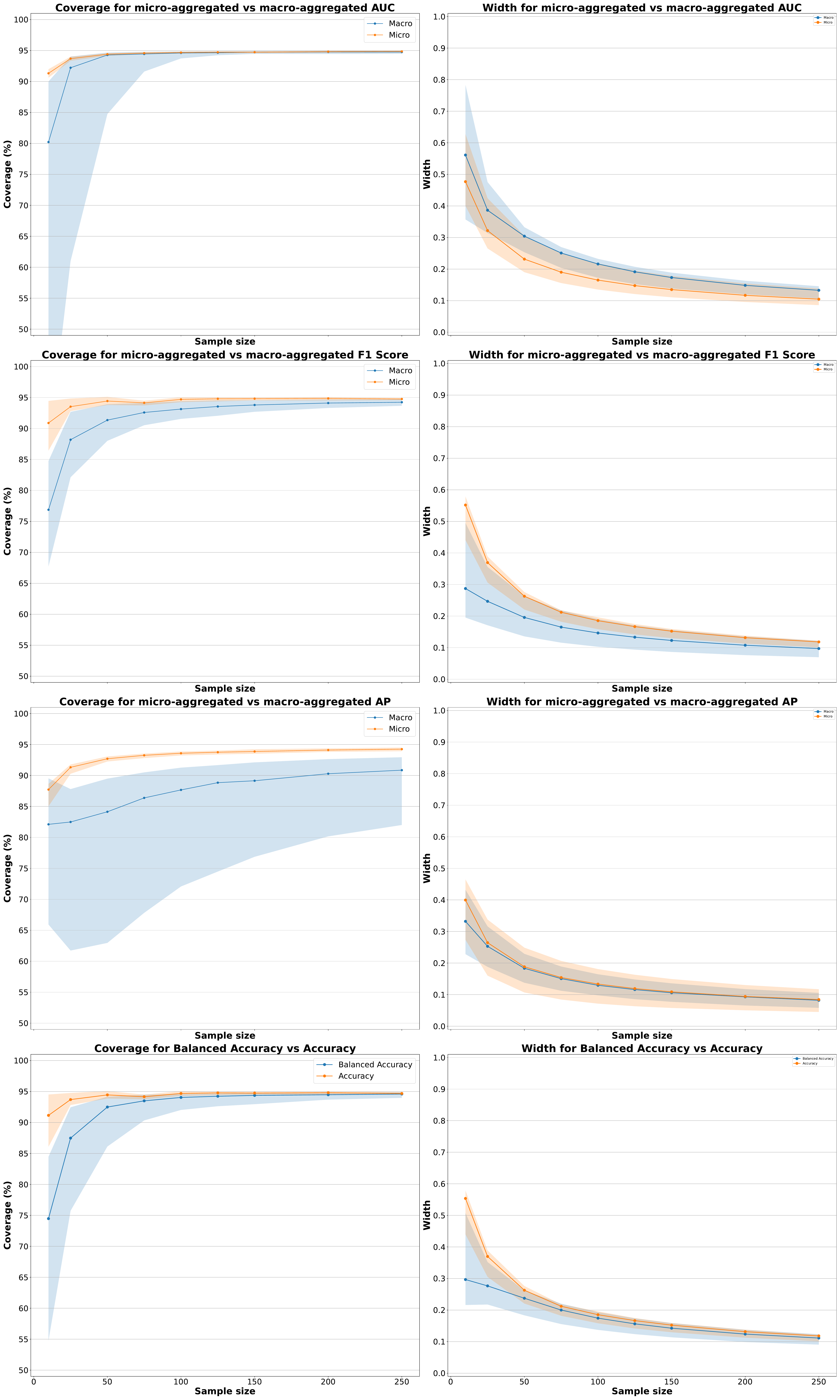}
    \caption{\textbf{Macro averaging requires more samples than micro averaging to achieve a given coverage.} One can observe that a given macro-averaged metric always converge more slowly than its micro counter-part. On the other hand, there are no consistent differences across widths. All intervals were obtained using percentile method. The colors represents the different aggregation method (micro and macro) for different classification metrics (AP, AUC, F1, Balanced Accuracy, Accuracy).}
    \label{fig:micro_macro_all}
\end{figure}

\paragraph{For classification, the aggregation method impacts the reliability of CIs.} Coverage convergence is slower for macro- than for micro-aggregation across all classification metrics. For instance, to achieve a coverage of $\sim94\%$, accuracy (micro-recall) requires only $\sim50$ observations while balanced accuracy (macro-recall) needs 3 times more, as shown on Figure \ref{fig:micro_macro}. This behavior also holds for other metrics, as shown on Supplementary Figure \ref{fig:micro_macro_all}. On the other hand, we did not observe a consistent difference for CI width across metrics (Supplementary Figure \ref{fig:micro_macro_all}).

\begin{figure}[H]
    \centering
    \includegraphics[width=\linewidth]{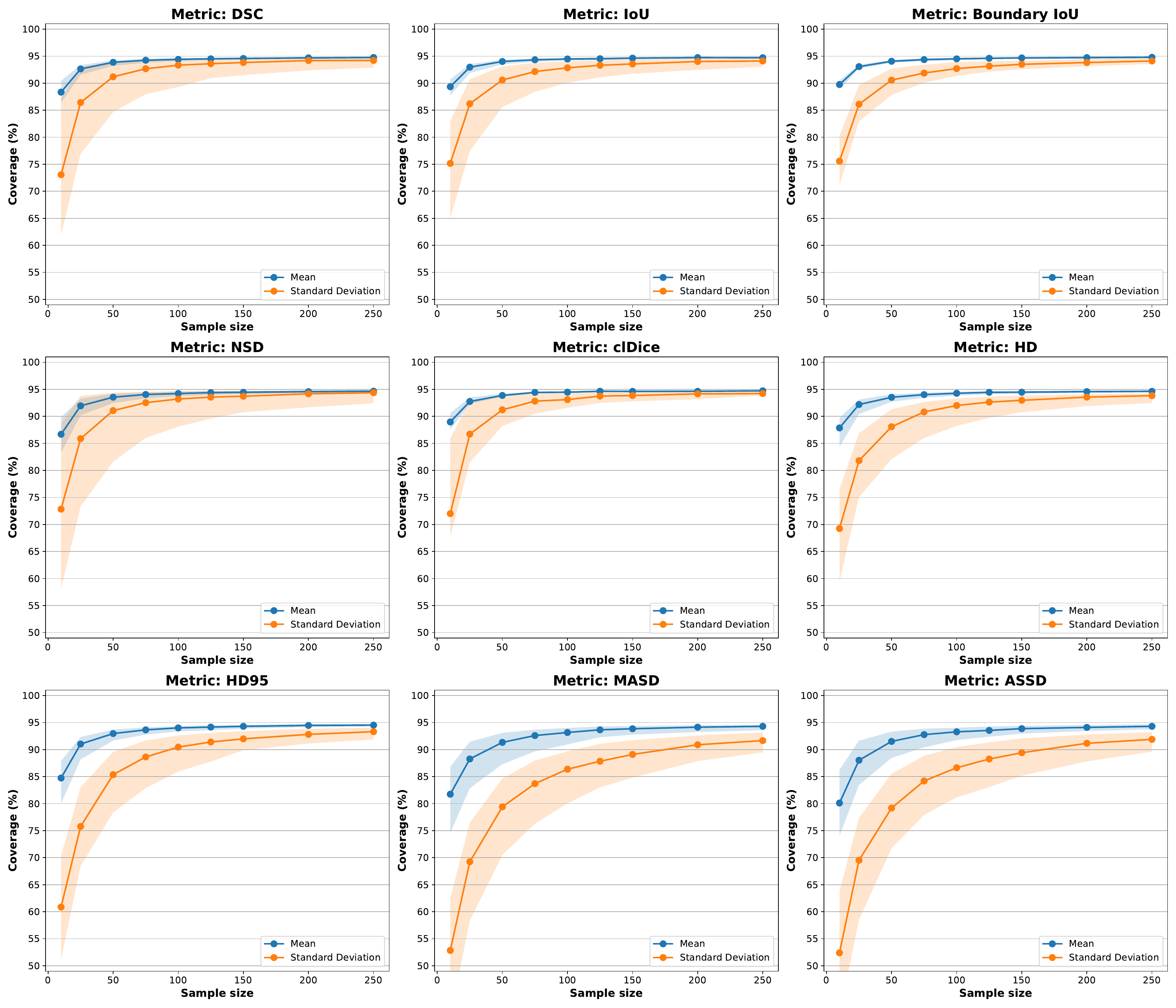}
    \caption{\textbf{Coverage converges at a slower pace for spread measures than for central tendency.} This holds across all segmentation metrics (IoU, Boundary IoU, ASSD, clDice, HD, DSC, NSD, MASD, HD95).}
    \label{fig:spread_vs_central}
\end{figure}

\paragraph{Empirical coverage converges more rapidly for central tendency  than for spread measures.} One needs fewer observations to reliably quantify the mean than the standard-deviation. Supplementary Figure \ref{fig:spread_vs_central} shows that this observation holds for all metrics. For median vs IQR, the difference holds for very small samples ($n=10$ and $n=25$) but is not systematic for larger samples.

\subsection{Sample size needs vary depending on  study characteristics and the CI method}

\begin{figure}
    \centering
    \includegraphics[width=\linewidth]{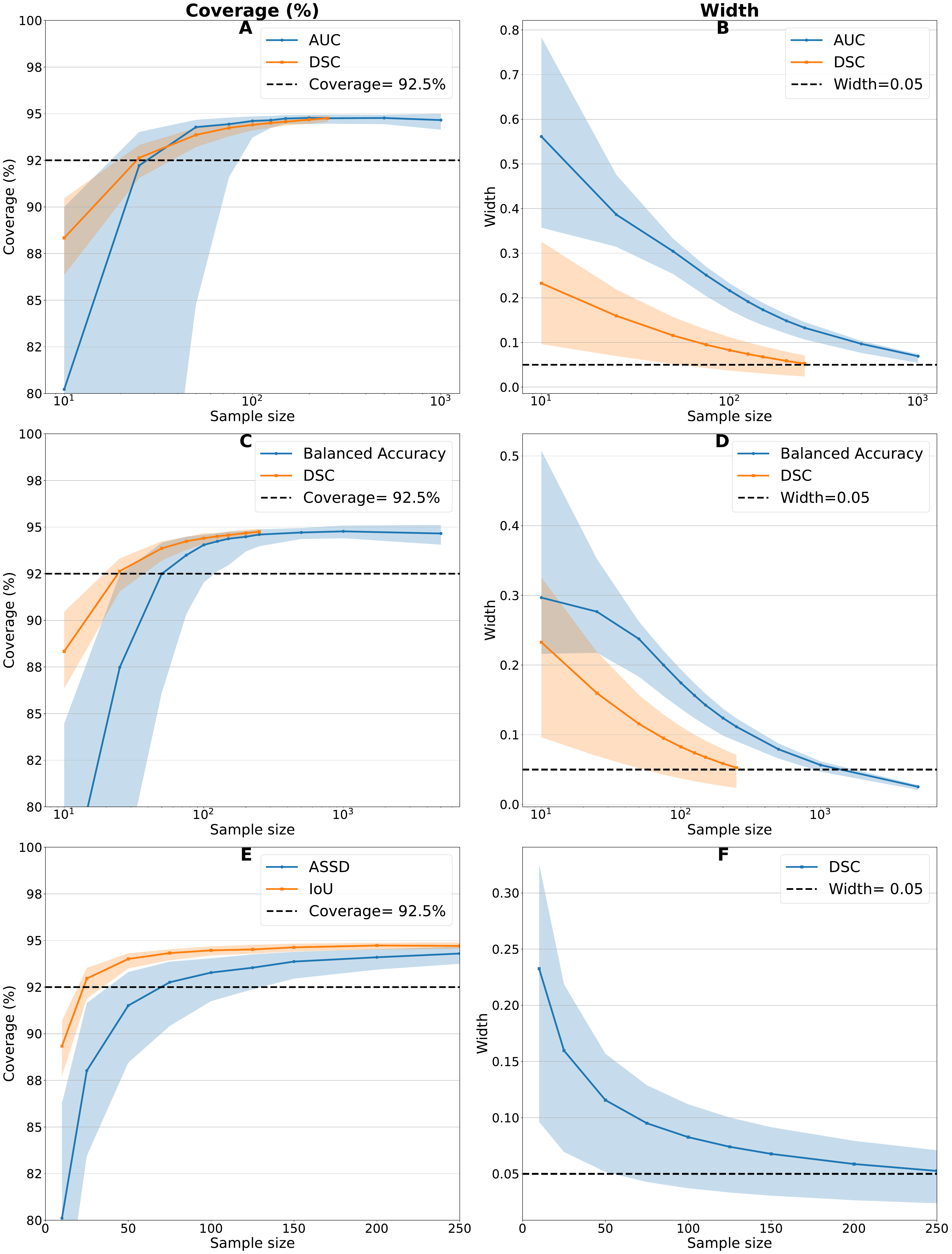}
    \caption{\textbf{Sample size needs can swing drastically from one study to another.} Each panel (A–F) illustrates a typical scenario:
    Panel A-B: Median, Q1 and Q3 coverage and width for CIs of the mean DSC vs AUC using percentile method.
    Panel C-D: Median, Q1 and Q3 coverage for CIs of the mean DSC vs Balanced Accuracy using percentile method.
    Panel E: Median, Q1 and Q3 coverage and width for CIs of the mean  IoU vs ASSD using percentile method.
    Panel F: Median, Q1 and Q3 width for CIs of the mean DSC using percentile method.}
    \label{fig:sample_needs}
\end{figure}

Depending on study characteristics and the CI method, required sample sizes to get reliable and precise CIs can vary by orders of magnitude, ranging from tens to thousands of observations, as illustrated by the 108 coverage and width plots (Supplementary Section \ref{sec:full_results}). Here, we highlight several representative cases. Regarding width, one can for instance see that, under a median scenario, one needs $n\approx250$ for mean DSC vs $n\approx1000$ for balanced accuracy to achieve a width of $0.05$ (Figure \ref{fig:sample_needs}, Panel D). For AUC, needs are even larger than for balanced accuracy since $0.05$ is not reached at $n=1000$ (Figure \ref{fig:sample_needs}, Panel B). One can find even wilder swings when looking across instances, e.g. $25\%$ of segmentation instances actually need less than 50 observations to reach a width of $0.05$ (Figure \ref{fig:sample_needs}, Panel F). This illustrates that, to achieve a given precision on performance, some studies will need less than 50 observations while others will need thousands. Differences can also be found in terms of coverage. For instance, to achieve at least $92.5\%$ coverage under a median scenario, one needs $n\approx25$ for mean DSC and AUC (Figure \ref{fig:sample_needs}, Panel A), $n\approx50$ for balanced accuracy and $n\approx75$ for ASSD (Figure \ref{fig:sample_needs}, Panel A, C and E). When looking across instances, differences get larger, and one needs $n\approx125$ for 25\% of ASSD instances.

\subsection{Fail cases}

A small number of benchmark instances exhibit severely degraded coverage. We analyze those instances below, defining a coverage failure case as an instance whose coverage deviates beyond $Q3 - 1.5 \times IQR$. 
\paragraph{Segmentation:  failures are easy to identify.} Failures of CIs of the mean in segmentation tasks are often associated with the presence of outliers in the empirical distributions of the performance metrics (see all failure cases for the percentile CI of the mean DSC in Supplementary Figure \ref{fig:fail_segm}). However, one can see that these fixed-sample failures seem to possess adequate asymptotic behavior, meaning that larger samples could solve the issue. Furthermore,  one should note that CIs of the median and of the trimmed mean have a good coverage in those cases, when using the percentile bootstrap (Supplementary Figure~\ref{fig:cov_segm_all}). 

\paragraph{Classification: failures are mostly linked to imbalance.} In classification tasks, coverage failures are predominantly associated with the use of imbalanced test datasets. For example, when using the percentile bootstrap with a test set size of 50, all coverage failures for accuracy occur in instances with imbalanced datasets, as shown in Supplementary Table \ref{tab:fail_accuracy}. Similarly, for balanced accuracy, $78\% (18/23)$ of the coverage failures correspond to instances with imbalanced datasets (Supplementary Table \ref{tab:fail_ba}).

\subsection{Robustness with respect to KDE settings}

To assess whether the key findings of the paper are robust under different KDE settings, we performed an ablation study in which we changed the different parameters of our KDE. Specifically, we studied different configurations specified in Table \ref{tab:kde_configs}. As explained in Supplementary Section \ref{sec:abl}, using the Gaussian kernel also requires changing the boundary correction method. This explains why both parameters change in Ablation 2. For Ablation 3, we chose to perform it as it uses usual choices for all 3 KDE components. 

\begin{table}[ht]
    \centering
    \begin{tabular}{l c c c}
        \toprule
        Configuration & Kernel & Bandwidth & Boundary correction \\
        \midrule
        Base & Epanechnikov & Adaptive & Trimmed \\[4pt]
        Ablation 1 & Epanechnikov & \textbf{Scott} & Trimmed \\
        Ablation 2 & \textbf{Gaussian} & Adaptive & \textbf{Trunc./Renorm.} \\
        Ablation 3 & \textbf{Gaussian} & \textbf{Scott} & \textbf{Trunc./Renorm.} \\
        \bottomrule
    \end{tabular}
    \caption{KDE configurations used for the ablation study.
           Bold entries indicate parameters that differ from the
           base configuration.
           The boundary correction applies to the segmentation
           step only; the classification step shares the same
           kernel and bandwidth without boundary correction.}
    \label{tab:kde_configs}
\end{table}

\begin{figure}[H]
    \centering
    \includegraphics[width=0.9\linewidth]{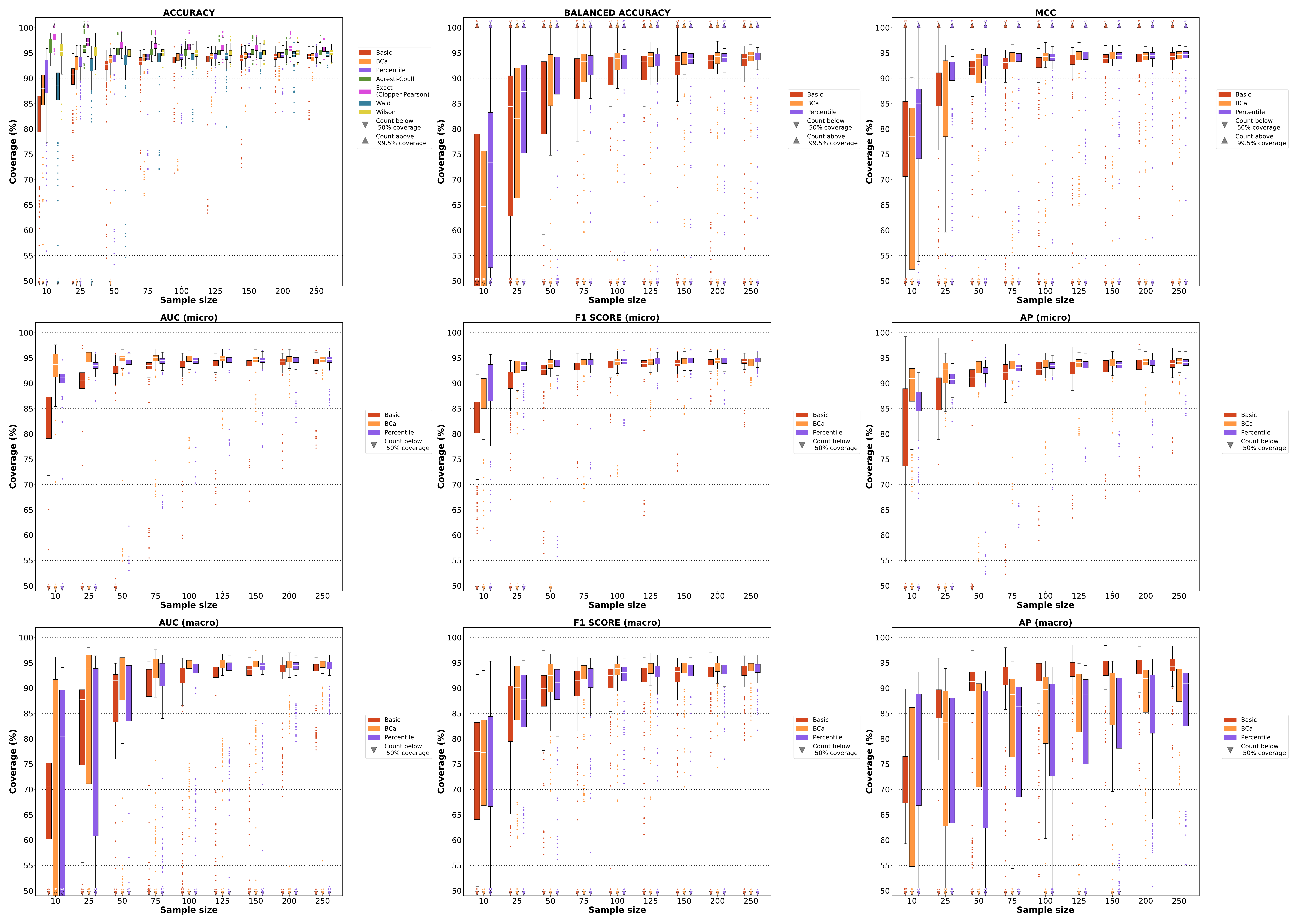}
    \caption{Coverages for all metrics and all aggregation methods for classification instances for Ablation 1. To ease zooming, this figure is also available as a separate PDF files at \url{\githubrepofig/blob/main/supplementary/epanechnikov_scott_trimmed/coverages_classif.pdf}.}
    \label{fig:cov_classif_all_ep_scott_trimmed}
\end{figure}

\begin{figure}[H]
    \centering
    \includegraphics[width=0.9\linewidth]{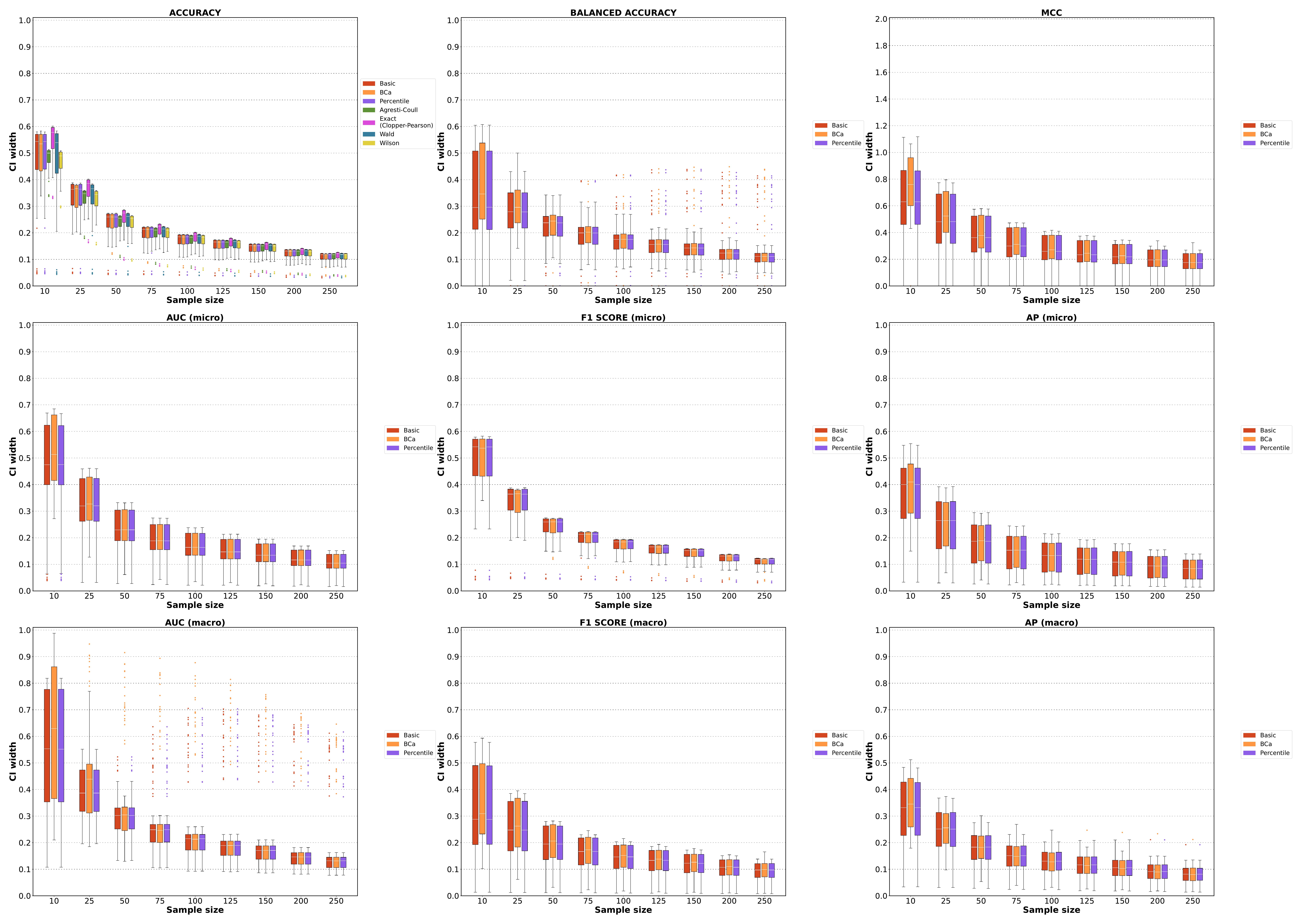}
    \caption{CI widths for all metrics and all aggregation methods for classification instances for Ablation 1. To ease zooming, this figure is also available as a separate PDF files at \url{\githubrepofig/blob/main/supplementary/epanechnikov_scott_trimmed/width_classif.pdf}.}
    \label{fig:width_classif_all_ep_scott_trimmed}
\end{figure}

\begin{figure}[H]
    \centering
    \includegraphics[width=0.9\linewidth]{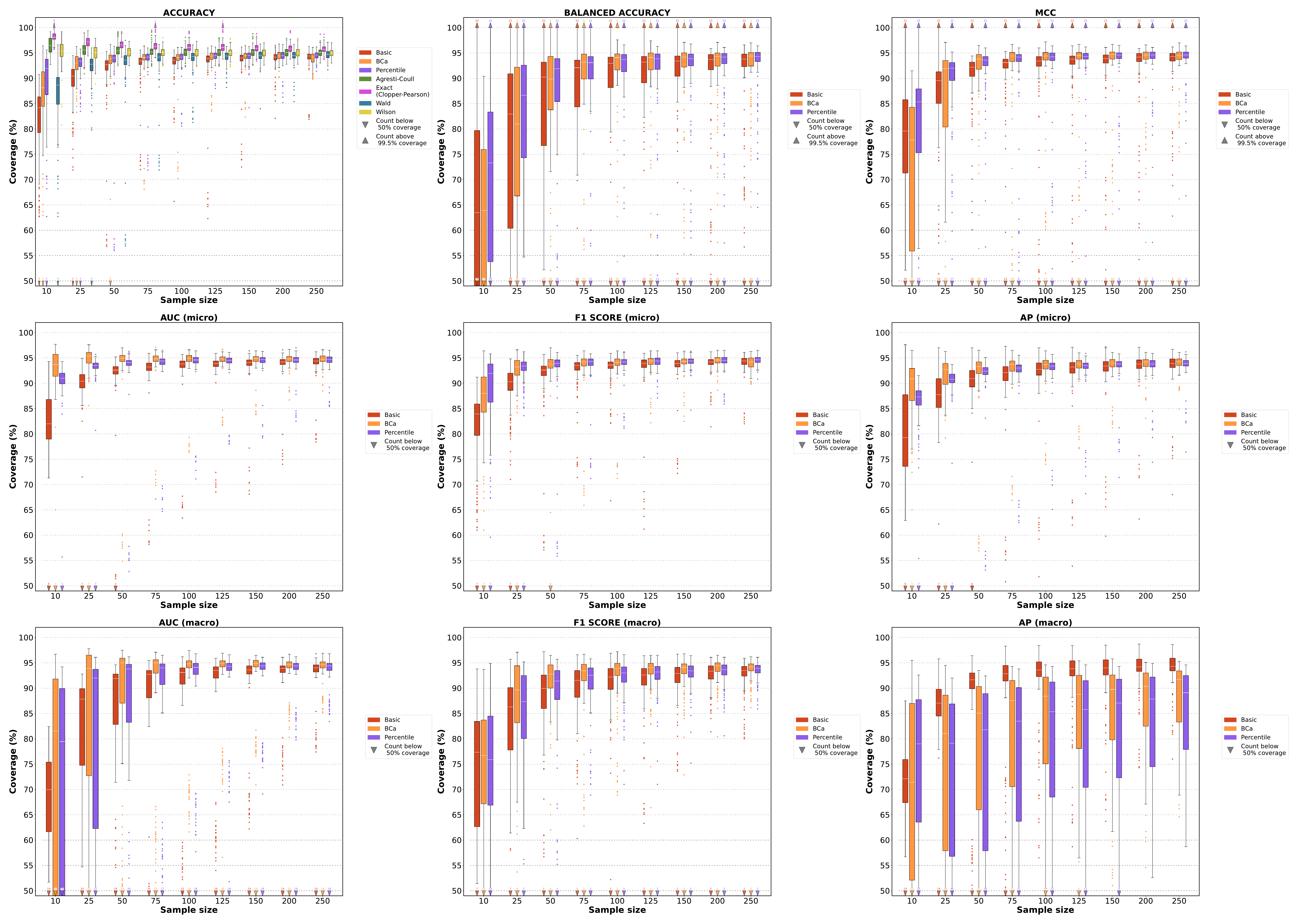}
    \caption{Coverages for all metrics and all aggregation methods for classification instances for Ablation 2. To ease zooming, this figure is also available as a separate PDF files at \url{\githubrepofig/blob/main/supplementary/gaussian_adaptive/coverages_classif.pdf}.}
    \label{fig:cov_classif_all_gauss_adaptive}
\end{figure}

\begin{figure}[H]
    \centering
    \includegraphics[width=0.9\linewidth]{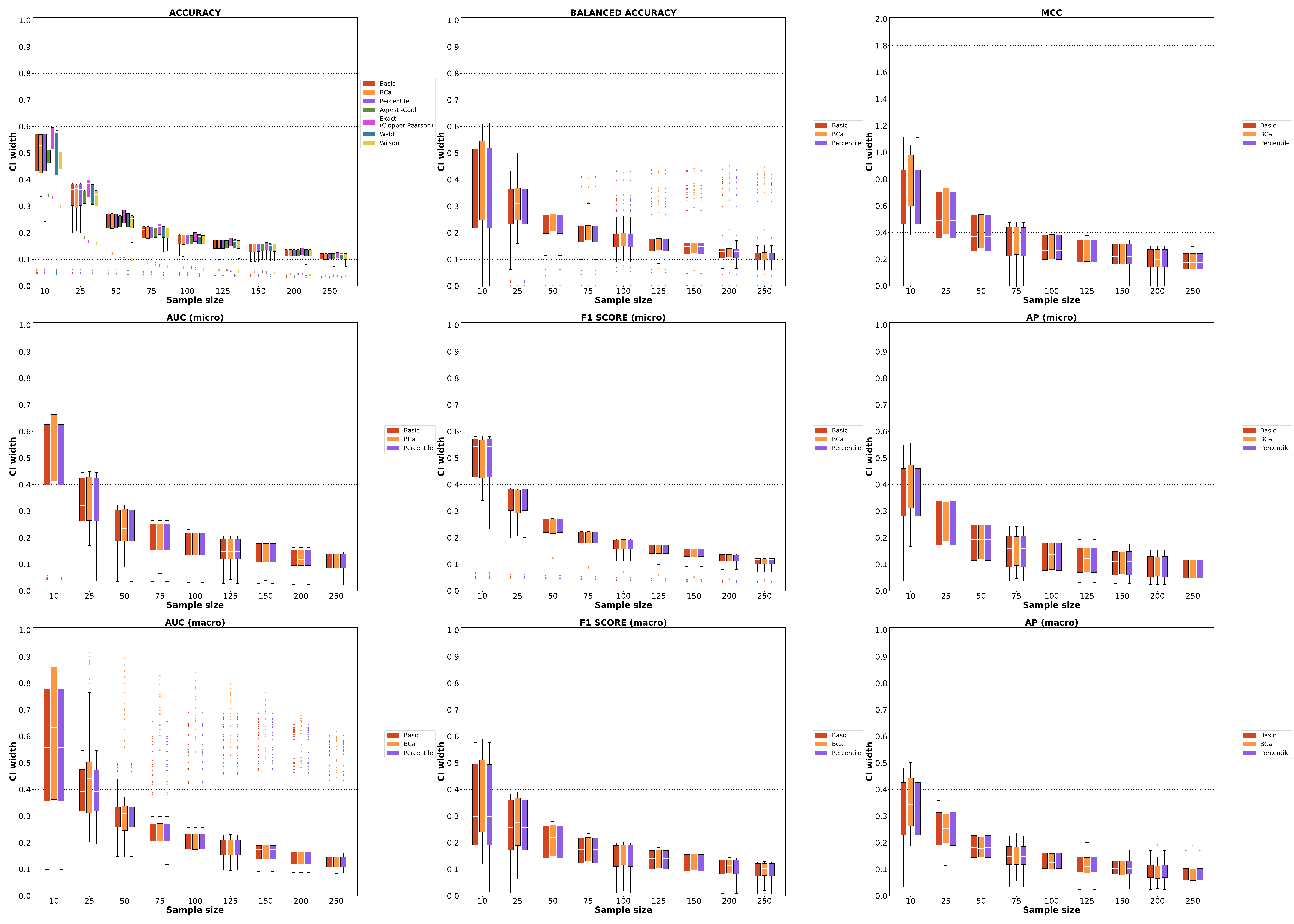}
    \caption{CI widths for all metrics and all aggregation methods for classification instances for Ablation 2. To ease zooming, this figure is also available as a separate PDF files at \url{\githubrepofig/blob/main/supplementary/gaussian_adaptive/width_classif.pdf}.}
    \label{fig:width_classif_all_gauss_adaptive}
\end{figure}

\begin{figure}[H]
    \centering
    \includegraphics[width=0.9\linewidth]{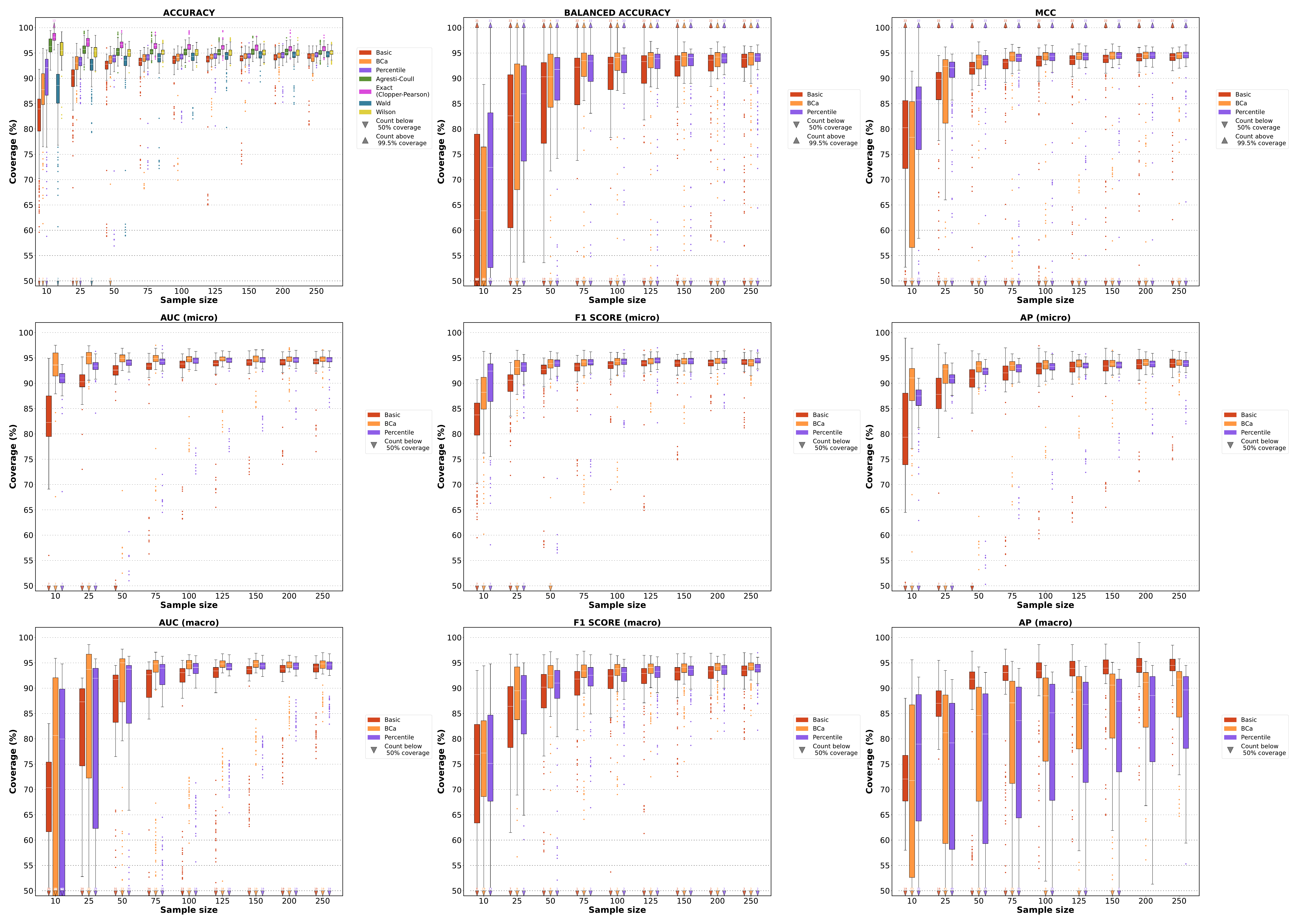}
    \caption{Coverages for all metrics and all aggregation methods for classification instances for Ablation 3. To ease zooming, this figure is also available as a separate PDF files at \url{\githubrepofig/blob/main/supplementary/gaussian_scott/coverages_classif.pdf}.}
    \label{fig:cov_classif_all_gauss_scott}
\end{figure}

\begin{figure}[H]
    \centering
    \includegraphics[width=0.9\linewidth]{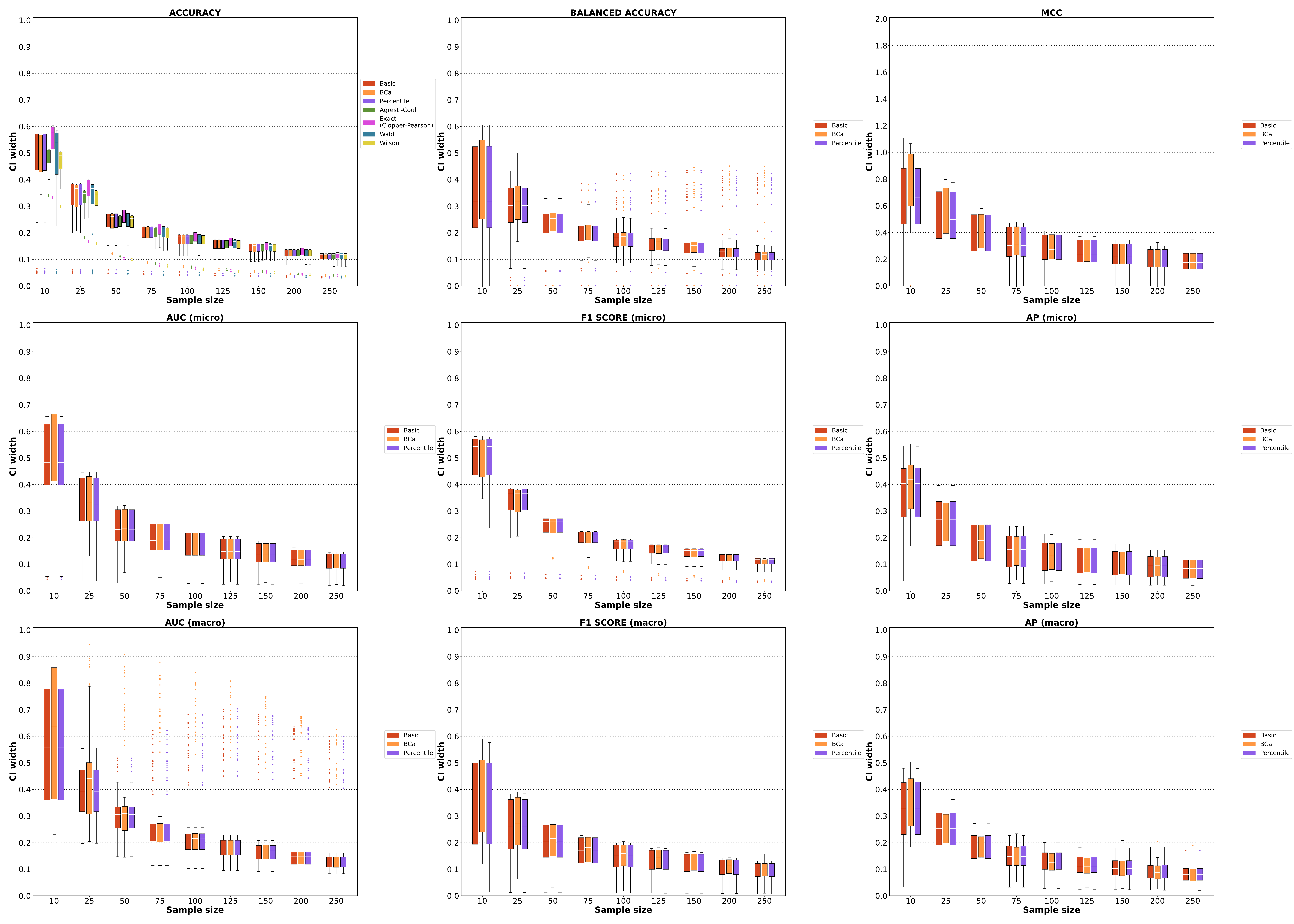}
    \caption{CI widths for all metrics and all aggregation methods for classification instances for Ablation 3. To ease zooming, this figure is also available as a separate PDF files at \url{\githubrepofig/blob/main/supplementary/gaussian_scott/width_classif.pdf}.}
    \label{fig:width_classif_all_gauss_scott}
\end{figure}

We then computed coverage and width values for each of our configurations, and obtained the raw results shown in Supplementary Figures \ref{fig:cov_segm_all_ep_scott_trimmed} to \ref{fig:width_classif_all_gauss_scott}. We started by doing a careful systematic visual inspection of all coverage and width graphs produced by different KDE settings, and found that the results are particularly stable and thus robust to variations in the KDE elements. To synthetically assess the robustness of our claims, we performed the same statistical tests as for the main results, following the same multiple comparison correction method. Results are shown in Supplementary Table \ref{tab:kde_sensitivity_claims}. These results are consistent with those obtained using our proposed KDE procedure. This supports that our claims are robust to different KDE settings.

\subsection{Computation times}

In total, our investigation took about $250,000$ hours of computation time. We aimed to dissect the computational time occurred by each aspect of the study, constructing an upper bound on the total time. Details are presented in Table \ref{tab:combined_times_vertical_groups}. The predicted upper bound was $265,106$ hours, not far from the real use time. In this table, the time given for “1 bootstrap” is actually across all sample sizes from n=10 to 250, thus close to a total sample size of 1,000. As one can see, one bootstrap is most often of the order of a second and at worst 20 seconds for the most costly classification metric. Then, one coverage computation means $10,000$ samples from the true distribution each assessed by   bootstrapping with 9,999 repetitions, thus $10^8$ resamplings for one benchmark instance, one metric, one summary statistic and one given $n$. Coverage computation remains very reasonable, typically a few hours, up to 50 hours in the worst case, keeping in mind that the actual time would be much lower with parallelization. On the other hand, our whole study was costly because this had to be done over 228 instances, 3 bootstrap methods, different metrics, and different summary statistics.

\begin{table}[!ht]
\centering
\caption{{\bf \sffamily Computational cost of the study}. The table presents the predicted total computational times based on the dissection of the different components as well as the actual computational time.
Columns correspond to different components: a single bootstrap, coverage computation, 3 bootstrap methods (percentile, basic and BCa), scaling to 228 benchmarking instances, and the resulting total computational cost.
For classification metrics, the Total column includes additional sampling when explicitly indicated by footnotes. Importantly, note that while our investigation was computationally heavy due its systematic nature, quantifying performance uncertainty for users is actually very cheap: a few seconds to get a bootstrap CI, at worst a few hours when one would like to verify their coverage.
}
\footnotesize
\setlength{\tabcolsep}{6pt}
\begin{threeparttable}
\begin{tabular}{c lccccc}
\hline
 &  & \textbf{1 bootstrap} & \textbf{Coverage} & \textbf{3 methods} & \textbf{228 instances} & \textbf{Total} \\
\hline
\multirow{5}{*}{\rotatebox{90}{\textbf{Segm.}}}
 & Mean   & 0.33 s   & 55 min   & 2.8 h   & 684 h   & 6,156 h\tnote{a} \\
 & Median & 1.18 s   & 3.2 h    & 10 h    & 2,280 h & 20,520 h\tnote{a} \\
 & IQR    & 2.52 s   & 7 h      & 21 h    & 4,788 h & 43,092 h\tnote{a} \\
 & STD    & 0.3 s    & 51.6 min & 2.6 h   & 592.8 h & 5,336 h\tnote{a} \\
 & Trimmed mean   & 0.72 s   & 2 h      & 6 h     & 1,368 h & 12,312 h\tnote{a} \\
\hline 
\multirow{4}{*}{\rotatebox{90}{\textbf{Micro}}}
 & Accuracy & 0.336 s & 56 min  & 168 min & 638 h   & 638 h \\
 & AUC      & 13.44 s & 37 h    & 112 h   & 25,536 h & 25,536 h \\
 & AP       & 15.12 s & 42 h    & 126 h   & 28,880 h & 28,880 h \\
 & F1       & 2.68 s  & 7.4 h   & 22.3 h  & 5,092 h & 5,092 h \\
\hline
\multirow{4}{*}{\rotatebox{90}{\textbf{Macro}}}
 & BA  & 6.72 s & 18.6 h & 56 h     & 12,768 h & 23,089 h\tnote{b} \\
 & AUC & 17.87 s & 49.6 h & 148.9 h  & 33,953 h & 40,298 h\tnote{c} \\
 & AP  & 20.1 s  & 55.8 h & 167.5 h  & 38,190 h & 38,190 h \\
 & F1  & 3.57 s  & 9.9 h  & 29.75 h  & 6,783 h  & 12,283 h \\
\hline
 & MCC & 3.36 s & 9.3 h & 28 h & 3,684 h & 3,684 h \\
\hline
\multicolumn{2}{l}{\textbf{Predicted total}} &  &  &  &  & \textbf{265,106 h} \\
\hline
\multicolumn{2}{l}{\textbf{Actual total}} &  &  &  &  & \textbf{243,614 h} \\
\hline

\end{tabular}

\begin{tablenotes}[flushleft]
\footnotesize
\item[a] For each summary statistic, there are 9 segmentation metrics.
\item[b] For Macro BA and Macro F1, we also sampled $n=500$, $1000$ and $n=5000$ but with only $3,500$ samples instead of $10,000$ and $3,499$ bootstrap resamplings instead of $9,999$.
\item[c] For Macro AUC, we also sampled $n=500$ and $n=1000$ but with only $3,500$ samples instead of $10,000$ and $3,499$ bootstrap resamplings instead of $9,999$.
\end{tablenotes}
\end{threeparttable}
\\
\vspace{0.2cm}
    {\footnotesize \sl \linespread{0.9}
Segm.: Segmentation summary statistics;
Micro: micro-averaged classification metrics;
Macro: macro-averaged classification metrics;
IQR: interquartile range;
STD: standard deviation;
BA: balanced accuracy;
AUC: area under the ROC curve;
AP: average precision;
F1: F1 score;
MCC: Matthews correlation coefficient.
Times are expressed in seconds (s), minutes (min), or hours (h).        \par
    }

\label{tab:combined_times_vertical_groups}
\end{table}

\newpage
\section{Discussion}

To our knowledge, this is the first large-scale investigation of CI methods in medical imaging. We unveiled key findings  about the behavior of CIs and their relationship to characteristics of medical imaging studies (our main findings are summarized in Figure \ref{fig:2}).

\paragraph{Sample sizes.} The sample size required to have both reliable and precise CIs can vary wildly: from tens to thousands. Our results show that this is influenced by all characteristics of a medical imaging study: ML problem, task, performance metrics, aggregation method. Furthermore, there are important variations across benchmark instances. Finally, this is also influenced by the choice of the CI method.

\begin{figure}[H]
    \centering
    \includegraphics[width=\linewidth]{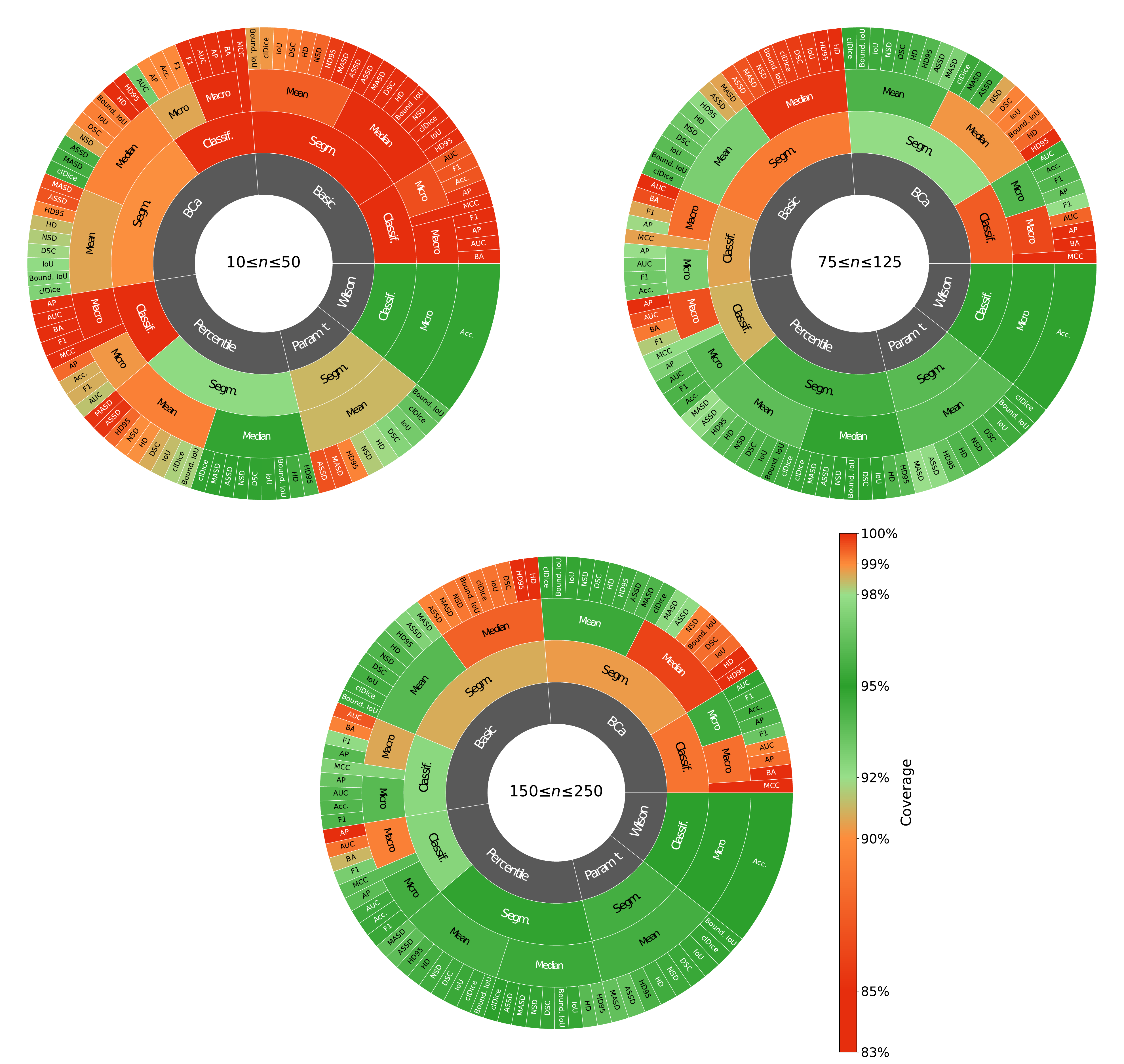}
    \caption{\textbf{Summary of the typical behavior of each CI method for different settings (sample size regime, task, aggregation method).} The sunbursts were colored according to the median coverage across instances. For leaves (metrics), the median is averaged across samples sizes in the considered regime. Going from leaves to the root, the coverage of a parent node is the average of its children’s coverages.}
    \label{fig:sunburst}
\end{figure}

\begin{figure}[H]
    \centering
    \includegraphics[width=\linewidth]{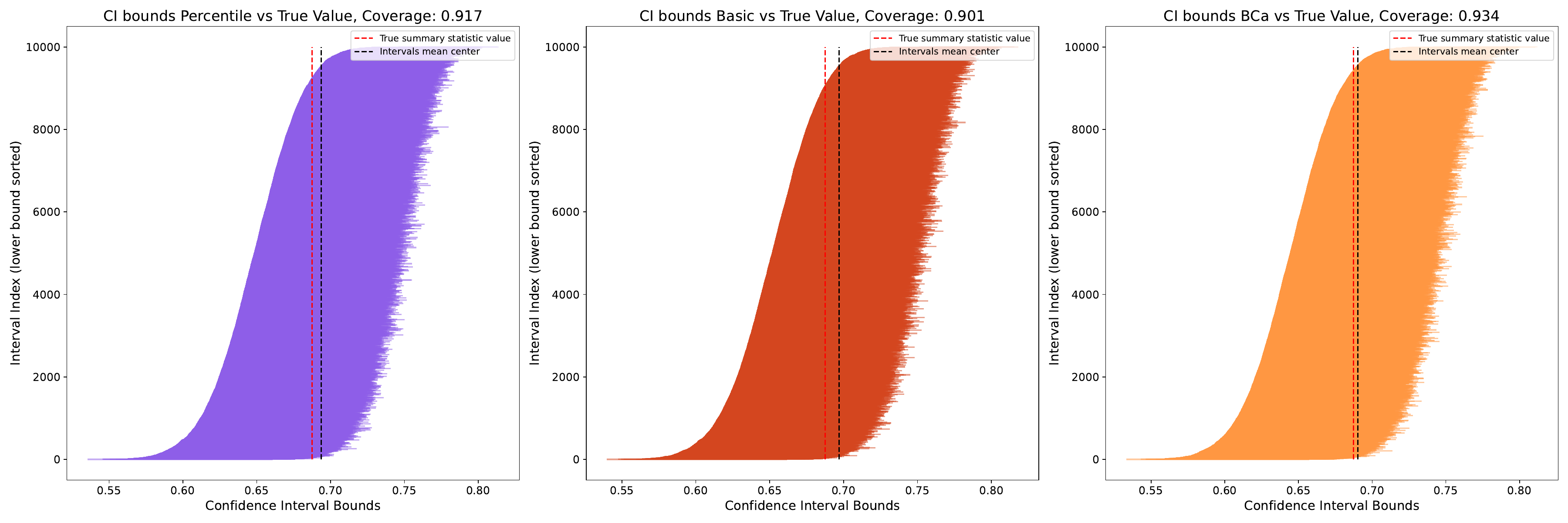}
    \caption{\textbf{The basic bootstrap produces intervals that are shifted with respect to the true value, when compared to other bootstrap methods.} In particular, when we compare the position of basic and percentile intervals, we can notice that the basic intervals are shifted by 0.003 on average, which leads to a drop of 2 coverage points.}
    \label{fig:explication_basic}
\end{figure}

\paragraph{CI methods.}An important takeaway regarding CI methods is that the percentile bootstrap is the most robust across settings. In contrast, the basic bootstrap is almost always the least reliable method. It exhibits poor coverage because the resulting intervals are biased on average with respect to the true value (Supplementary Figure \ref{fig:explication_basic}). Basic is a misnomer, as its principle is less intuitive than that of percentile and it is not a good default.
The behavior of the BCa bootstrap is more nuanced. In some settings, it provides the best coverage due to its second-order accuracy, but it can also fail catastrophically, most notably when the median is used as the summary statistic. More generally, such failures can occur for statistics that remain unchanged after the removal of a single data point. This problem that arises with the median had been noted  before by \citet{abdi2010jackknife} but is not widely known. We detail it in Supplementary Section \ref{sec:fail_bca}. In short, there is first an intrinsic issue. The acceleration factor can be undefined in some cases. Even when defined, it is at best estimated from three values and thus poorly estimated. Finally, it can also lead to numerical instability. Some parametric methods can be advantageous for very small sample sizes but this advantage quickly vanishes. As expected, concentration inequalities produce very wide intervals, reflecting the cost of finite-sample, distribution-free guarantees. The typical behavior of each CI method for different settings (sample size regime, task, aggregation method) is summarized in Figure~\ref{fig:sunburst}. Readers are encouraged to refer to Supplementary Figures~\ref{fig:cov_segm_all} to ~\ref{fig:width_classif_all} to visualize the behavior across instances, beyond typical scenarii. Finally, contrary to common belief, bootstrap methods are computationally inexpensive. Namely the cost is at worst a few seconds in our regimes. For millions of observations and costly sample-level metrics, it could at worst reach a few hours (Supplementary Section~\ref{sec:suppl_bootstrap_time} and Supplementary Table~\ref{tab:bootstrap_times_max}).

\paragraph{Classification vs segmentation.} Classification requires substantially larger sample sizes than segmentation to achieve comparable CI properties, both in terms of coverage and width. This observation is consistent with the findings of \cite{jurdi2023confidence} for interval width. In this current work, we make this claim more robust by backing it up with hundreds of benchmark instances versus 4 in their paper and many more metrics. We further show that it is also true for coverage. The greater difficulty of uncertainty quantification in classification is likely due to the fact that classification metrics convey only one bit of information per individual in the test set, whereas segmentation metrics aggregate many bits per individual (approximately one per pixel or voxel). As a result, classification metrics exhibit higher estimation noise, requiring larger sample sizes to achieve the same level of confidence in performance estimates. Furthermore, things can get even harder in imbalanced settings, as shown by the failed cases.

\paragraph{Metrics.}
Segmentation metrics exhibit markedly different behaviors. In particular, Mean Average Surface Distance (MASD) and Average Symmetric Surface Distance (ASSD) require larger sample sizes to achieve satisfactory coverage than other metrics. We show that this effect can be explained by the skewness of their distributions: the higher the skewness, the larger the sample size required. The relationship between convergence pace and skewness is well-documented in statistics \citep{hesterberg2015teachers, rousselet2023introduction, cumming2001primer}. In particular, Edgeworth expansions of coverage probabilities reveal an explicit dependence of coverage error on  skewness~\citep{hall1992bootstrap}. Specifically, \citep{hall1992bootstrap} showed that the coverage of an $\alpha$-level CI method can be expanded for any fixed sample size as a polynomial function as follows: $C = 1 - 2\alpha + n^{-1}B(z_{1-\alpha})\phi(z_{1-\alpha})+O(n^{-2})$, where $B$ is a polynomial whose coefficients depend on the skewness and the kurtosis of the instance distribution, $\phi$ is the PDF of the standard Gaussian distribution, and $z_{1-\alpha}$ is the $1-\alpha$ quantile of the standard Gaussian distribution.
As shown in Figure~\ref{fig:metrics}, more skewed distributions correspond to slower convergence. Further, we observe that distance-based metrics tend to produce more skewed distributions than overlap metrics. A major explanation is that  they are not bounded on the right. Indeed, the boundedness of overlap metrics constrains their skewness given their variance and their mean: for a variable supported on $[0, 1], |\gamma| \le \max(\mu, 1 - \mu)/\sigma$, where $\mu$ is the mean, $\sigma^2$ is the variance and $\gamma$ is the skewness. Thus, the uncertainty related to distance-based metrics which are unbounded on the right is intrinsically harder to estimate and this effect is maximal for the most skewed ones, namely ASSD and MASD. 

The skewness also likely explain the pattern across summary statistics. We observe (Supplementary Figure \ref{fig:cov_segm_all}) that the mean corresponds to a slower coverage convergence compared to the median or the trimmed mean. Indeed, the mean inherits the full skewness of the instance distribution, whereas the median and the trimmed mean are governed by the local skewness around the center of the distribution and are largely insensitive to the right tail. The same reason explains why BCa is the most robust bootstrap method in these regimes, since its acceleration constant is a first-order correction for precisely this skewness term. As a consequence, when dealing with skewed metrics, BCa requires less observations to achieve a given coverage.

Metric choice should of course be driven by clinical relevance rather than statistical convenience \citep{maier2024metrics}. Nonetheless, once metrics are selected, it is important to be aware of their implications for sample size requirements and uncertainty quantification.

For macro-aggregated AP, we observe an unexpected critical failure of both BCa and percentile bootstrap methods, while basic bootstrap outperforms them. A plausible explanation is the known positive bias of the standard AP estimator \citep{sahiner2016semi}, a setting in which the basic bootstrap can be advantageous. However, this does not explain why the same behavior is not observed for micro-aggregated AP.

\paragraph{Aggregation methods.} Micro-averaging needs smaller samples than macro-averaging to produce reliable CIs. This difference is likely due to how the metrics are computed. In macro-aggregation, the metric is first computed separately for each class using only the samples belonging to that class (with $n_d$ observations for class $d$), and the final score is obtained by averaging these class-level metrics across all classes. As a result, the macro metric is computed on a sample of size $\sum_{k=1}^dn_k=n$. In contrast, micro aggregation computes the metric by pooling all observations from all classes, effectively using the full set of $n\times d$ class-observation pairs. This pooling increases the effective sample size, which explains why micro-aggregated metrics require fewer observations than macro-aggregated ones to achieve similar coverage.

There are many applications in which macro averaging is preferable to micro averaging, as micro-averaged metrics can mask performance changes in small or rare classes. By giving equal weight to each class, macro aggregation provides a more balanced and clinically meaningful assessment in such settings.

We also observe that measures of dispersion converge more slowly than measures of central tendency. This is expected, as estimating variability inherently requires more observations than estimating a mean or median, leading to larger sample size requirements for reliable CIs.

\paragraph{Failure cases.}
For segmentation, failure cases correspond to empirical metric distributions that contain outliers. Outliers in the empirical metric distribution indicate the presence of outliers in the population. With small sample sizes, these outliers may be missing or overrepresented in the test data. Missing outliers can shrink CIs, creating overconfidence in performance estimates, while overrepresented outliers increase variability and can lead to inaccurate estimates. Both cases risk misleading conclusions, such as retaining poorly performing models or discarding clinically relevant ones.

In classification, coverage failures are associated with the use of imbalanced datasets. When working with small or imbalanced samples, performance estimates for minority classes are inherently unreliable, leading to considerable uncertainty about how the model will perform on the general population. This especially arises when using macro-aggregated metrics which compute performance separately for each class. Since minority classes often contain only a few observations, accurately estimating their performance is difficult. This is clinically important: even in large datasets, rare classes, which may correspond to  uncommon but medically relevant disease types, can make precise performance estimation very challenging.

\paragraph{Computation times.}
As one can see, the total computation time of our investigation is consequent. This is due to the breadth of our study and not to the weight of individual computations. Not only bootstrap is very cheap, even a coverage computation is actually very reasonable (a dozen CPU hours on a single core).

\paragraph{Limitations.}
KDE choices could potentially impact the results. However, we paid particular attention to the design of our KDE and showed that our claims are robust to various KDE settings. Nevertheless, we are aware that our studies estimate the coverage against a model-based surrogate rather than the unknown real population.

We deliberately focused on a set of  CI methods which we believe are the most likely to be used by our community. However, many other methods exist such as for instance parametric methods for other summary statistics such as the median or for other classification metrics such as AUC \citep{qin2008comparison}. It could be interesting to study their added value.

The computation of CIs rely on an independence assumption between the observations in the test set, which is not always the case when dealing with medical images. For instance, one may have multiple images from the same patient, leading to a hierarchical structure inside the test set. In such cases, the CI method needs to take into account the correlation between clustered observations. Failure to do so will in general lead to underestimating the interval width. However, the factors influencing the magnitude of this effect are varied and complex. Treating this issue is beyond the scope of the present paper and requires a separate study.

To keep the scope manageable, we studied only the two most common tasks in medical image analysis: classification and segmentation. Expanding this research to other tasks, such as for instance detection, registration or synthesis would be valuable. We hypothesize that many of our findings for segmentation would generalize to other tasks that rely on individual-level metrics, such as image reconstruction or synthesis, although this would require explicit experimental verification.

\paragraph{Implications.}
We believe our work has important and actionable implications for researchers of the community. Awareness about performance uncertainty is rising but researchers often do not know which method to choose for which setting. 

The comparison of CI methods carries important implications. Many users can be tempted to blindly rely on “BCa” or “basic” because they are often recommended over “percentile” and because “BCa” is the default in some bootstrap implementations. In the AI community, which is largely Python-based, the SciPy bootstrap function~\citep{virtanen2020scipy} is particularly valuable, as it makes these methods widely accessible. This function implements the three most common bootstrap variants with “BCa” as the default. Our results show that no single method works best in all settings. We therefore encourage the medical imaging community to carefully evaluate the options rather than blindly relying on the default.

Furthermore, our work brings important information about the different settings that may impact the robustness of uncertainty reporting. We identified the main characteristics that impact the reliability and precision of CIs, so that researchers know which parameters to be aware of when reporting performance.

We have summarized the implications for practitioners into decision trees, shown in Figure \ref{fig:recommendation_trees}. The procedure used to create these trees can be found in Supplementary Section \ref{sec:decision_tree}. The tree allows to decide upon which CI method to choose depending on study characteristics. This provides useful and actionable advice to the medical imaging AI community. However, this does not constitute comprehensive guidelines on performance uncertainty reporting. Indeed, further work would be required, for instance addressing other metrics or non-independent data, as well as building a consensus from a panel of experts.

\paragraph{Conclusion.} In summary, this paper is the first large-scale investigation of confidence intervals for performance uncertainty in medical imaging. As such, it bridges an important gap in the literature. Our results have direct and immediate implications for researchers and can help prevent the community from falling into some major pitfalls. They constitute important building blocks towards future guidelines for performance uncertainty reporting in medical imaging AI. 

\begin{figure}[p]
    \centering
    \includegraphics[width=\linewidth]{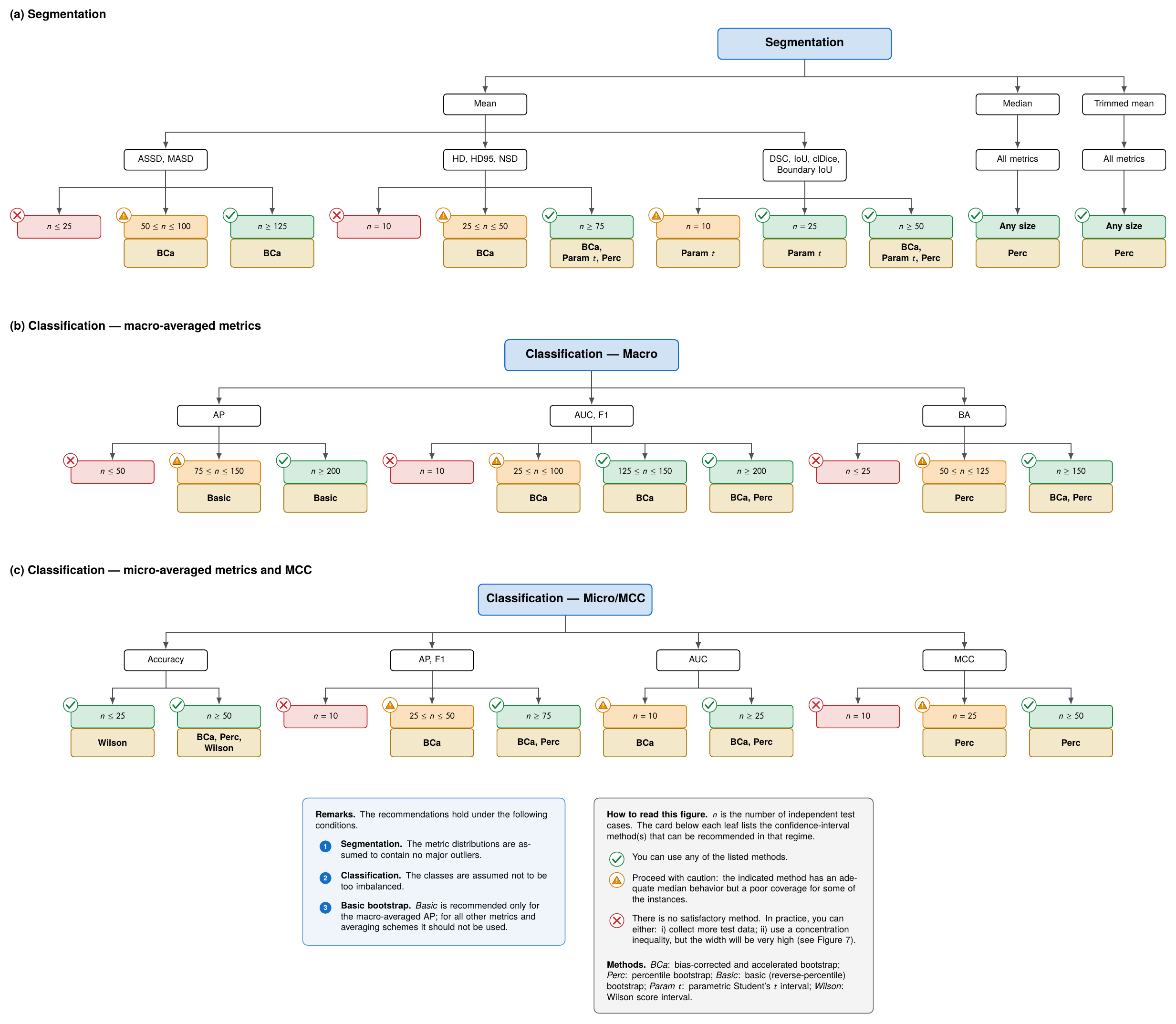}
    \caption{\textbf{Implications for practitioners in the form of a decision tree.} The trees were built automatically the procedure detailed in Supplementary Section~\ref{sec:decision_tree}. Remember what no method can fix: outliers in the per-sample distribution (segmentation) and class imbalance (classification) drive nearly all coverage failures (Figure \ref{fig:fail_segm}, Tables \ref{tab:fail_accuracy}-\ref{tab:fail_ba}).}
    \label{fig:recommendation_trees}
\end{figure}

% Acknowledgments section

\newpage

\section*{Acknowledgments}
The authors are grateful to Reuben Dorent and Charlotte Godard for stimulating discussions. This work was granted access to the HPC resources of IDRIS under the allocation AD011016489 made by GENCI (Grand Equipement National de Calcul Intensif).

\section*{Funding sources}
The research leading to these results has received funding from the French government under management of Agence Nationale de la Recherche as part of the “France 2030” program (reference ANR-23-IACL-0008, project PRAIRIE-PSAI), and as part of the "Investissements d'avenir" program (reference ANR-10-IAIHU-06, project Agence Nationale de la Recherche-10-IA Institut Hospitalo-Universitaire-6) and from the European Union’s Horizon Europe Framework Programme (grant number 101136607, project CLARA). This publication received funding from the European Research Council (ERC) under the European Union’s
Horizon 2020 research and innovation program (grant agreement no. 101002198, NEURAL SPICING). Part of
this work was also funded by Helmholtz Imaging (HI), a platform of the Helmholtz Incubator on Information
and Data Science. Moreover, this project has received funding from the National Center for Tumor Diseases
(NCT) Heidelberg’s Surgical Oncology Program. This publication was also supported through state funds
approved by the State Parliament of Baden-W\"urttemberg for the Innovation Campus Health Life Science
Alliance Heidelberg Mannheim.
%% Add more if needed 

% Acknowledgments section
\section*{Disclosure of interests}

% %% Use for preprint
% Disclosure of interests related to the present article. The authors declare that they have no known competing financial interests or personal relationships that could have appeared to influence the work reported in this paper.

% Disclosure of interests unrelated to the present article. OC reports having received consulting fees from Therapanacea (2022-2024). OC reports that other principal investigators affiliated to the team which he co-leads have received grants (paid to the institution) from Sanofi (2020-2022) and Biogen (2022-2023). OC reports that his spouse was an employee of myBrainTechnologies (2015-2023) and was an employee of DiamPark (2023-2025). OC holds the following (unpaid) editorial responsibilities: Senior Area Editor for IEEE Transactions on Medical Imaging (IEEE), Associate Editor of Medical Image Analysis (Elsevier) Associate Editor of the Journal of Medical Imaging (SPIE).

% For MedIA
Disclosure of interests related to the present article.
Olivier Colliot and Annika Reinke serve as (unpaid) associate editors for Medical Image Analysis (Elsevier). They had no involvement in the peer-review of this article and have no access to information regarding its peer-review.  Full responsibility for the editorial process for this article was delegated to another journal editor.
If there are other authors, they declare that they have no known competing financial interests or personal relationships that could have appeared to influence the work reported in this paper.

Disclosure of interests unrelated to the present article. OC reports having received consulting fees from Therapanacea (2022-2024). OC reports that other principal investigators affiliated to the team which he co-leads have received grants (paid to the institution) from Sanofi (2020-2022) and Biogen (2022-2023). OC reports that his spouse was an employee of myBrainTechnologies (2015-2023) and was an employee of DiamPark (2023-2025). OC holds the following (unpaid) editorial responsibilities: Senior Area Editor for IEEE Transactions on Medical Imaging (IEEE), Associate Editor of the Journal of Medical Imaging (SPIE).

\section*{Declaration of Generative AI and AI-assisted technologies in the writing process}
During the preparation of this work the author(s) used ChatGPT in order to improve readability and language. The author(s) further used Claude to assist in preparing a draft version of some portions of the code as well as some portions of the proofs. After using these service, the authors have reviewed and edited the content (text, proofs, code) as needed and take full responsibility for the content of the published article and its associated code.

% References section
\bibliographystyle{apalike}
\bibliography{references}

\newpage
\begin{center}
%\begin{flushleft}
    \setstretch{2.0} %% allows a nicer formatting of the title by adding more space between lines
    \textbf{\huge \sffamily Supplementary Material}
\end{center}

\vspace{0.7cm}

\beginsupplement

\section{Metrics and summary statistics}
\label{sec:empirical_evaluation}
This section provides a more formal presentation of metrics, summary statistics and empirical evaluations. 

Let $\mathcal{Z}$ denote a space of \textbf{inputs} and $\mathcal{Y}$ a space of \textbf{targets}. Inputs and targets can take various forms depending on the use case (e.g. image and image for segmentation, image and classes for classification, etc.). A trained machine learning model is a function $f: \mathcal{Z} \longrightarrow\mathcal{Y}$.

Let $\mathcal{D}$ denote a distribution on $\mathcal{Z}\times\mathcal{Y}$. When evaluating a machine learning model, one uses a collection of pairs $(Z_i,Y_i)_{i=1}^n\overset{i.i.d.}{\sim}\mathcal{D}$, called \textbf{test sample}~\footnote{With a slight abuse of terminology, we will use test set and test sample interchangeably; Formally, the test set is the realization of a test sample.}.  When evaluating the model's performance, one will compute the \textbf{predicted target} that the model associates to each input. We will denote these predictions as $Y'_i = f(Z_i)$. 

Now, let us define what a metric and a summary statistic are. As mentioned in Section~\ref{subsec:nontechnical}, the term metric is often used both for individual-level (e.g. Dice Similarity Coefficient (DSC)) and sample-level metrics (e.g. accuracy). Here we will use a stricter terminology where metric is used only for individual-level metrics and we say summary statistic for sample-level metrics. A \textbf{metric} is defined as a function $m:\mathcal{Y}^\mathbb{R}\times\mathcal{Y}^\mathbb{R} \longrightarrow \mathcal{X}$. $\mathcal{X}$ can be for instance $[0,1]$ (DSC, Normalized Surface Distance (NSD)\ldots), $\mathbb{R^+}$ (Average Symmetric Surface Distance (ASSD), Mean Average Surface Distance (MASD)\ldots), or $\{0,1\}$. Usually, this function is evaluated in each pair $(Y'_i,Y_i)$, to provide metric values $X_i := m(Y'_i,Y_i)$, but it can happen that the scores require other targets or predicted targets, such as the metric “precision at the threshold $Y'_i$". The collection $(X_i)_{i=1}^n$ forms the \textbf{empirical metric distribution}. In the following, $(X_i)_{1\leq i\leq n}$ are independent and identically distributed (i.i.d.).

Formally, $X_i$ are random variables through their dependence upon the couples $(Z_i,Y_i)$, meaning they follow some distribution $\mathcal{F}$, defined on the codomain of $m$.  \textbf{Summary statistics} are derived from the distribution $\mathcal{F}$\footnote{Or equivalently from the test set, if one looks at the realization.}. Formally, a summary statistic is a function $t : \mathcal{P}(\mathcal{M})\longrightarrow \mathbb{R}$, where $\mathcal{P}(\mathcal{M})$ denotes the set of all probability distributions supported on any set $\mathcal{M}$. Examples of such statistics are the mean, the median, the standard-deviation, the accuracy etc. The mean is the summary statistic $t : \mathcal{F} \mapsto \int_\mathbb{R} xd\mathcal{F}(x)$. Accuracy is the mean when the individual-level metric is the indicator function $m(Y',Y) = \mathbbm{1}\{Y'=Y\}$. 

When performing model validation, one aims at assessing the \textbf{true value} $t(\mathcal{F})$, where $\mathcal{F}$ is the underlying distribution of $X$. In practice, we only observe a finite sample $(X_i)_{1\leq i\leq n}$  of $X$, which provides an estimation of the distribution, denoted $\Hat{\mathcal{F}}$. This is the quantity one will report when presenting validation results of a ML model. Since $t(\Hat{\mathcal{F}})$ is an estimate from a finite number of observations $n$, it is noisy, in other words it is imprecise. It is crucial to characterize how precise it is. This is what confidence intervals (CIs) do.

\newpage

\section{Confidence interval methods}
\label{sec:CIs}
This section describes the CI methods used in the paper.

\subsection{Parametric confidence intervals}
\label{subsec:details_parametric_CIs}
\subsubsection[Parametric z interval]{Parametric $\boldsymbol{z}$  interval (for the mean)}

Let $\mu = \mathbb{E}[X]$, the true mean of the metric distribution and $\sigma^2=\mathbb{V}[X]$ the true variance of the metric distribution. This CI method assumes that the true variance is known~\footnote{In practice, we consider that the empirical variance is close enough to the true variance for large $n$.}. The Central Limit Theorem states that: 
$$
\frac{\sqrt{n}(\Bar{X_n} - \mu)}{\sigma} \underset{n\rightarrow\infty}{\overset{d}{\longrightarrow}} \mathcal{N}(0,1)
$$
where $\Bar{X_n} = \frac{1}{n}\sum_i X_i$ is the empirical mean of the sample $X_1,...,X_n$. \\

Therefore, if we denote $z_\alpha$ the $\alpha$ quantile of the standard Gaussian distribution, we know that asymptotically : 

\begin{align*}
    &\mathbb{P}\left(z_{\alpha/2} \le \frac{\sqrt{n}(\bar{X_n} - \mu)}{\sigma} \le z_{1-\alpha/2}\right) \underset{n\rightarrow\infty}{{\longrightarrow}} 1-\alpha \\
    \Leftrightarrow\; &\mathbb{P}\left(\frac{\sigma}{\sqrt{n}}z_{\alpha/2} \le \Bar{X_n} - \mu \le \frac{\sigma}{\sqrt{n}}z_{1-\alpha/2}\right) \underset{n\rightarrow\infty}{{\longrightarrow}} 1-\alpha\\
    \Leftrightarrow\; & \mathbb{P}\left(\bar{X_n} - \frac{\sigma}{\sqrt{n}}z_{\alpha/2} \le \mu \le \bar{X_n} - \frac{\sigma}{\sqrt{n}}z_{1-\alpha/2}\right) \underset{n\rightarrow\infty}{{\longrightarrow}} 1-\alpha\\
    \Leftrightarrow\; & \mathbb{P}\left(\bar{X_n} + \frac{\sigma}{\sqrt{n}}z_{1-\alpha/2} \le \mu \le \bar{X_n} - \frac{\sigma}{\sqrt{n}}z_{1-\alpha/2}\right) \underset{n\rightarrow\infty}{{\longrightarrow}} 1-\alpha
\end{align*}

The last equivalence comes from the symmetry of the Gaussian distribution around 0, so $-z_\alpha = z_{1-\alpha}$. \\

Therefore, we have constructed an asymptotical interval with a theoretical coverage of $1-\alpha$, namely the \textbf{parametric} $\boldsymbol{z}$ interval:
$$CI_z = \left[\bar{X_n}\pm \frac{\sigma}{\sqrt{n}}z_{1-\alpha/2}\right]$$

\subsubsection[Parametric t interval]{Parametric $\boldsymbol{t}$ interval (for the mean)}

We place ourselves in the same setting as before, but we remove the hypothesis that the variance is known. Instead, we estimate it by the usual variance estimator $\Hat{\sigma}^2 = \frac{1}{n-1}\sum_i (X_i - \bar{X_n})^2$. If $X_1, \ldots, X_n$ are i.i.d. and follow a Gaussian distribution $\mathcal{N}\left(\mu, \sigma^2\right)$, then using the fact that $(n-1)\times\hat{\sigma}^2$ follows a $\chi^2$ distribution with $n-1$ degrees of freedom, we have access to the exact distribution of $\frac{\sqrt{n}(\bar{X_n} - \mu)}{\Hat{\sigma}} $:

$$
\frac{\sqrt{n}(\bar{X_n} - \mu)}{\Hat{\sigma}} \hookrightarrow \mathcal{T}(n-1)
$$
where $\mathcal{T}(n-1)$ is the Student's $t$ distribution with $n-1$ degrees of freedom. Again, by using the quantiles of order $1-\alpha/2$ of the $t$-distribution $t_{n-1,1-\alpha/2}$, we can construct an interval with coverage $1-\alpha$. With the a similar derivation, one obtains the \textbf{parametric} $\boldsymbol{t}$ interval: 
$$CI_t = \left[\bar{X_n}\pm \frac{\Hat{\sigma}}{\sqrt{n}}t_{n-1,1-\alpha/2}\right]$$

\subsubsection{Wald parametric interval (for proportions)}

We are in the setup where we observe $X\sim \text{Binomial}(n,p)$. We define the sample proportion as $\Hat{p} = X/n$, which estimates $\mathbb{E}\left[X\right]$ and the sample variance as $ \Hat{p}(1-\Hat{p})$, which estimates $\mathbb{V}[X]$.  The Wald interval is simply the parametric $z$ interval applied to a binomial proportion, thereby approximating a binomial distribution with a Gaussian distribution using the Moivre-Laplace theorem. The interval is
$$CI_W = \left[\Hat{p}\pm \sqrt{\frac{\Hat{p}(1-\Hat{p})}{n}}z_{1-\alpha/2}\right]$$

The Wald interval is unreliable when $\Hat{p}$ gets close to 0 or 1. For example, if one observes $\Hat{p}=1$, the interval is reduced to a single point, which does not make sense, since one only observed a finite sample.

\subsubsection{Agresti-Coull parametric interval (for proportions)}

The Agresti-Coull method addresses the above shortcoming of the Wald interval. Still with $X\sim \text{Binomial}(n,p)$, one  considers the following
\begin{align*}
    \Tilde{n} &= n + z_{\alpha/2}\\
    \Tilde{X} &= X + \frac{z_{\alpha/2}}{2}\\
    \Tilde{p} &= \Tilde{X}/\Tilde{n}
\end{align*}

In plain English, this is equivalent to observing a few more observations, split evenly between success and failure. The \textbf{Agresti-Coull} interval is the same as the Wald interval, but using $\Tilde{p}$ and $\Tilde{n}$ instead of $\Hat{p}$ and $n$: 
$$CI_{AC} = \left[\Tilde{p} \pm z_{\alpha/2}\sqrt{\frac{\Tilde{p}(1-\Tilde{p})}{\Tilde{n}}}\right]$$

\subsubsection{Wilson parametric interval (for proportions)}

The Wilson interval is obtained by inverting the score test for binomial proportions. We still observe $X\sim \text{Binomial}(n,p)$. The log-likelihood up to an additive constant is 
$$
l(p) = X\ln(p) + (n-X)\ln(1-p)
$$
The score function being defined as the first derivative of the log-likelihood, we obtain: 
$$
s(p) = \frac{X}{p} - \frac{n-X}{1-p} = \frac{X-np}{p(1-p)}
$$
The Fisher information is defined as: 
$$
I(p) = -\mathbb{E}\left[\frac{d^2l}{dp^2}\right] = \mathbb{E}\left[\frac{X}{p^2} + \frac{n-X}{(1-p)^2}\right] = \frac{np}{p^2} + \frac{n(1-p)}{(1-p)^2} = \frac{n}{p} + \frac{n}{1-p} = \frac{n}{p(1-p)}
$$
The standardized score test statistic is defined as: 
$$
Z(p) = \frac{s(p)}{\sqrt{I(p)}} = \frac{X-np}{p(1-p)} \frac{\sqrt{p(1-p)}}{\sqrt{n}} = \frac{X-np}{\sqrt{np(1-p)}}
$$
For a two-sided CI, we want to know the values of $p$ such that $|Z(p)| \le z_{\alpha/2}$. Denoting $z=z_{\alpha/2}$ for simplicity, we obtain: 

\begin{align*}
    |Z(p)| &\le z\\
    \Leftrightarrow\; \left\lvert \frac{X-np}{\sqrt{np(1-p)}} \right\rvert &\le z \\
    \Leftrightarrow\; \frac{(X-np)^2}{np(1-p)} &\le z^2
\end{align*}
This can be expressed as an inequality on a second-order polynomial in $p$:
$$
(n+z^2)p^2 - (2X+z^2)p + \frac{X^2}{N} \le 0
$$
The two roots of this polynomial equation are: 
$$
p = \frac{2X + z^2 \pm z\sqrt{\frac{nz^2 + 4X(n-X)}{n}}}{2(n+z^2)}
$$

We can rewrite this in terms of $\Hat{p}=X/n$ : 
$$
p = \frac{\Hat{p}+z^2/(2n)}{1+z^2/n} \pm \frac{z}{1+z^2/n} \sqrt{\frac{\Hat{p}(1-\Hat{p})}{n} + \frac{z^2}{4n^2}} 
$$

The \textbf{Wilson} interval is then defined as the values of $p$ that stand between the roots, that is
$$CI_{Wilson} = \left[\frac{\Hat{p}+z_{\alpha/2}^2/(2n)}{1+z_{\alpha/2}^2/n} \pm \frac{z_{\alpha/2}}{1+z_{\alpha/2}^2/n} \sqrt{\frac{\Hat{p}(1-\Hat{p})}{n} + \frac{z_{\alpha/2}^2}{4n^2}}\right]$$

\subsubsection{Clopper-Pearson parametric interval (for proportions)}

This CI method is also sometimes called \textbf{exact} CI for a binomial proportion, as it directly inverts the cumulative distribution function (CDF) of the Binomial distribution, rather than relying on Gaussian approximations like the methods presented before. Let us define: 

$$
p_L = \underset{p}{\text{argmin}} \left\{\mathbb{P}_p(X\geq x) \geq \frac{\alpha}{2}\right\}
$$
$$
p_U = \underset{p}{\text{argmax}} \left\{\mathbb{P}_p(X\leq x) \geq \frac{\alpha}{2}\right\}
$$
where 
$\mathbb{P}_p(X \ge x)
=
\sum_{k=x}^{n}
\binom{n}{k}
p^k (1-p)^{\,n-k}$.

In practice, if we denote $\text{Beta}^{-1}(q;a,b)$ the $q$-th quantile of a Beta$(a,b)$ distribution\footnote{The Beta distribution appears because the binomial CDF integrates to a Beta function.}, we compute these quantities as: 
\begin{align*}
    p_L &= \text{Beta}^{-1}(\frac{\alpha}{2};x,n-x+1)\\
    p_U &= \text{Beta}^{-1}(1-\frac{\alpha}{2};x+1,n-x)
\end{align*}

The \textbf{Clopper-Pearson} CI is then defined as:
$$CI_{CP} = [p_L,p_U]$$

\subsection{Non-parametric confidence intervals using bootstrap}

This section describes bootstrap methods \citepsupp{davison1997bootstrapsuppl}. We are now looking to estimate any summary statistic $\theta$ from our metric distribution $\mathcal{F}$.~\footnote{Some summary statistics cannot be bootstrapped but this does not apply to all those considered in this paper.} The key idea of the bootstrap is to approximate the sampling distribution of the summary statistic.

\label{sec:bootstrap}
\subsubsection{Percentile bootstrap interval}

 Let us denote $B$ the number of bootstrap samples to draw from our original sample. To create a bootstrap sample, we draw with replacement a sample of size $n$ from our original sample of size $n$ as well. For each bootstrap sample, indexed by $b = 1,...,B$, we compute the value of the summary statistic of interest on the sample, yielding a new estimate of the true value of the population statistic, which we denote $\theta^*_b$. The estimates $(\theta^*_b)_{b=1}^B$ form the \textbf{bootstrap distribution}. 

Once we have this distribution, since we want to produce an $\alpha$-level CI, one intuitive idea is to take the quantiles of this distribution, of level $\alpha/2$ and $1-\alpha/2$. This yields the \textbf{percentile bootstrap} CI $$CI_{perc} = [\theta^*_{\alpha/2}, \theta^*_{1-\alpha/2}]$$

\subsubsection{Basic (reverse percentile) bootstrap interval}

We place ourselves in the exact same setup as for the percentile bootstrap. We construct our quantiles the same way to obtain $\theta^*_{\alpha/2}$ and $\theta^*_{1-\alpha/2}$. The difference from the percentile method is that we try to estimate the distribution of the bias of the estimator rather than the distribution of the estimator itself. If the estimator is biased, then maybe the bootstrap distribution will be biased as well, producing a biased interval. This method is based on the rationale that the distribution of the bias $\Hat{\theta} - \theta$ can be estimated by the empirical distribution of $\theta^* - \Hat{\theta}$. We can use this assumption to write:
$$
\mathbb{P}(a\le \Hat{\theta} - \theta \le b) \approx \mathbb{P}(a\le \theta^* - \Hat{\theta} \le b) = 1-\alpha
$$
Taking the quantiles as before, we obtain: 
$$
1-\alpha = \mathbb{P}(\theta^*_{\alpha/2} - \Hat{\theta}\le \Hat{\theta} - \theta \le \theta^*_{1-\alpha/2} - \Hat{\theta}) = \mathbb{P}(2\Hat{\theta} - \theta^*_{1-\alpha/2}\le \theta \le 2\Hat{\theta} - \theta^*_{\alpha/2})
$$

We thus obtain \textbf{basic bootstrap} CI

$$CI_{basic} = [2\Hat{\theta} - \theta^*_{1-\alpha/2}, 2\Hat{\theta} - \theta^*_{\alpha/2}]$$

\subsubsection{Bias-corrected and accelerated (BCa) bootstrap interval}\label{BCa_appendix}

This bootstrap method was introduced by \citetsupp{efron1987bettersuppl} to account for biased and skewed bootstrap distributions. The hypothesis made here is that there exists a transformation function $g$ such that, denoting $\phi=g(\theta)$ and $\Hat{\phi} = g(\Hat{\theta})$, and $\tau$ the constant (not estimated) standard error of $\phi$, $\frac{\Hat{\phi}-\phi}{\tau} \sim \mathcal{N}(-z_0\sigma_\phi,\sigma_\phi^2)$, where $z_0$ is called \textbf{bias constant} and $\sigma_\phi = 1+a\phi$, where $a$ is the \textbf{acceleration constant}. We will now see how these constants are computed.

\paragraph{Computation of $\boldsymbol{z_0}$.}

For the bias constant, the idea is pretty simple. Let $\theta^*_1, ..., \theta^*_B$ be our bootstrap distribution. We will compute the proportion of the bootstrap distribution that is below the estimate of the statistic, namely:
$$
\Hat{p} = \frac{\sum_b \mathbbm{1}\{\theta^*_b\le\Hat{\theta}\}}{B}
$$

And then, denoting $\Phi$ the cumulative distribution function of the standard Gaussian distribution, $z_0 = \Phi^{-1}(\Hat{p})$.

\paragraph{Computation of $\boldsymbol{a}$.}
\label{BCa_accel}

To compute the acceleration coefficient, we will use the jackknife method. For each value of $i=1,...,n$, we will compute the statistic of the original sample where the $i$-th observation is left out. This will yield a collection of $n$ leave-one-out statistics $(\Hat{\theta}_{(i)})_{i=1}^n$. We will define the mean $\bar{\theta}_{(.)} = \frac{1}{n}\sum_i \Hat{\theta}_{(i)}$ of these leave-one-out statistics and use the empirical skewness of this mean to compute the acceleration $a$:
\begin{equation}
    a = \frac{\sum_i (\bar{\theta}_{(.)} - \Hat{\theta}_{(i)})^3}{6[\sum_i(\bar{\theta}_{(.)} - \Hat{\theta}_{(i)})^2]^{3/2}}
    \label{eqn:bca_a}
\end{equation}

\paragraph{Derivation of the BCa interval from $\boldsymbol{z_0}$ and $\boldsymbol{a}$.}

Instead of using the regular $\alpha/2$ and $1-\alpha/2$ quantiles of the bootstrap distribution such as with percentile bootstrap, we will use other quantiles of level $\alpha_1$ and $\alpha_2$, defined as:
\begin{align*}
    \alpha_1 &= \Phi\left(z_0 + \frac{z_0 + z_{\alpha/2}}{1-a(z_0 + z_{\alpha/2})}\right)\\
    \alpha_2 &= \Phi\left(z_0 + \frac{z_0 + z_{1-\alpha/2}}{1-a(z_0 + z_{1-\alpha/2})}\right)
\end{align*}

Finally, the \textbf{BCa} CI is defined as:
$$CI_{BCa} = [\theta^*_{\alpha_1}, \theta^*_{\alpha_2}]$$

\subsection{Non-parametric confidence intervals from concentration inequalities}
\label{sec:concentration}

Concentration inequalities describe how much a random variable concentrates around a typical value (e.g. its mean or median)~\citepsupp{boucheron2013concentrationsuppl,abhishek2019introductionsuppl}. They are widely known to ML researchers. They can be used to derive CIs, among many other applications. They make only weak distributional assumptions (typically boundedness or sub-normality\footnote{Boundedness implies sub-normality from Hoeffding's lemma. Thus, all inequalities that apply to sub-normal distributions apply to bounded distributions.}) and provide a finite-sample guarantee that the coverage of the CI is at least $1-\alpha$. There are multiple inequalities and corresponding CIs. We only recall two of them here, which are used in the paper, and which both apply to bounded random variables\footnote{In fact, they only assume that the random variable is bounded almost surely but this does not make a difference in practice.}. They are both widely known and have a different spirit.  The first is Hoeffding’s inequality~\citepsupp{hoeffding1963probabilitysuppl}, which is arguably the most common. One of its key characteristics is that it does not depend on the variance of the data. The second one is the so-called Empirical Bernstein inequality \citepsupp{maurer2009empiricalsuppl}, which makes use of the empirical variance. Let us briefly describe them here.

\subsubsection{Hoeffding's inequality confidence interval}

Hoeffding's inequality is as follows. Let $a$ and $b$ two reals such that $\mathbb{P}(a\le X \le b) = 1$~\footnote{Meaning $X$ is bounded almost surely.}, and $\bar{X}_n = \frac{1}{n}\sum_i X_i$. 
Then:
$$
    \forall t>0, \mathbb{P}(|\bar{X}_n - \mathbb{E}[X]|\ge t) \le 2\exp\left(\frac{-2nt^2}{(b-a)^2}\right)
$$

The left-hand side of this expression can also be written $\mathbb{P}(\bar{X}_n-t\le \mathbb{E}[X]\le \bar{X}_n+t)$, which is very close to the definition of a CI. One can then let $\alpha$ equal the right-hand side to solve for $t$ and derive an $1-\alpha$-level CI: 
\begin{align*}
    &\alpha = 2\exp\left(\frac{-2nt^2}{(b-a)^2}\right)\\
    \Leftrightarrow \;&\ln(\alpha/2) = \frac{-2nt^2}{(b-a)^2}\\
    \Leftrightarrow \;&(b-a)^2\ln(2/\alpha) = 2nt^2\\
    \Leftrightarrow \;&t = \sqrt{\frac{(b-a)\ln(2/\alpha)}{2n}}
\end{align*}

In the case of a metric bounded [0,1], and for a $\alpha=95\%$ CI, the CI is: 
$$
\text{CI}_{\text{H}} = [\bar{X}_n \pm \frac{\sqrt{\ln(2/0.05)}}{\sqrt{2n}}] \approx [\bar{X}_n \pm \frac{1.358}{\sqrt{n}}]
$$

% \begin{figure}[ht]
%     \centering
%     \includegraphics[width=0.5\linewidth]{supp_figs/hoeffding_t.pdf}
%     \caption{\textbf{Concentration inequalities produce too wide intervals.} Comparison between the width of the CI produced using Hoeffding's inequality with a CI produced by the parametric $t$ method. Data is distributed according to a uniform distribution, and sample sizes range from 10 to 500.}
%     \label{fig:hoeffding}
% \end{figure}

\subsubsection{Empirical Bernstein confidence interval}

As one can see, Hoeffding's CI does not depend on the variance of $X$. This is a drawback as, even when variance is very low, the intervals are still very wide. Bennett's inequality takes into account $\mathbb{V}[X]$. However, in practice, one only has access to the empirical variance, not the true variance $\mathbb{V}[X]$. The Empirical Bernstein inequality provides a bound that depends on the empirical variance, instead of the true variance \citepsupp{maurer2009empiricalsuppl}. With $X$ in $[0,1]$\footnote{One can also write it more generally with $[a,b]$}, $X_1, \ldots, X_n \in [0,1]$, $\bar{X}_n = \frac{1}{n}\sum_i X_i$ and $V_n(X)
=\frac{1}{n-1}\sum_{i=1}^n (X_i-\bar{X_n})^2$, one has with probability at least $1-\alpha$ %over the i.i.d.\ vector $\mathbf{X} = (X_1,\dots,X_n)$
:  

$$
\mathbb{E}[X] - \bar{X}_n
\le
\sqrt{\frac{2 V_n (X)\,\ln(2/\alpha)}{n}}
+
\frac{7 \ln(2/\alpha)}{3 (n-1)} .
$$

We substitute $V_n (X)$ with $\hat{\sigma}^2$.
One can subsequently derive the two-sided CI at $1-\alpha$ with a similar derivation than for Hoeffding's:
$$\text{CI}_{\text{EB}}=\left[\Bar{X}_n \pm \hat{\sigma}\sqrt{\frac{2\,\ln(4/\alpha)}{n}}+\frac{7\ln(4/\alpha)}{3(n-1)}\right]$$

\newpage
\section{Segmentation: experimental data and  characterization}

\subsection{Metric hyperparameters}
\label{sec:hyperparameter}
NSD and Boundary IoU metrics require a hyperparameter. We used the same values as proposed in the original Medical Segmentation Decathlon paper \citepsupp{antonelli2022medicalsuppl}. These values are task-dependent, to adapt to the characteristics of the organ. Quoting the Decathlon paper \citepsupp{antonelli2022medicalsuppl} “NSD was defined at task level and was the same for all the targets of each task. The value represented what they would consider an acceptable error for the segmentation they were performing. The following values have been chosen for the individual tasks (in mm): Brain-5; Heart-4; Hippocampus-1; Liver-7; Lung-2; Prostate-4; Pancreas-5; Colon-4; Hepatic vessel-3; Spleen-3.". The same values were used for  Boundary IoU.

\subsection{Diversity of distributions}
\label{subsec:diversity_segm}
To analyze the variety of the metric distributions at our disposal, we looked at the mean, standard-deviation as well as higher-order moments, namely skewness and kurtosis (respectively order 3 and 4). These descriptive statistics provide insights about the shape of a distribution, as skewness indicates how asymmetrical the distribution is and kurtosis reflects the “tailedness" of a distribution. Denoting $X$ a 1D real random variable with existing finite mean $\mu$, variance $\sigma^2$, skewness $\gamma$ and kurtosis $\kappa$, one has: 
\begin{align*}
    \gamma &= \mathbb{E}\left[\frac{(X-\mu)^3}{\sigma^3}\right]\\
    \kappa &= \mathbb{E}\left[\frac{(X-\mu)^4}{\sigma^4}\right]
\end{align*}

Results are presented in Supplementary Figure~\ref{fig:skewness_kurt_segm}.

\label{sec:diversity_segm}
\begin{figure}[H]
    \centering
    \includegraphics[width=0.8\linewidth]{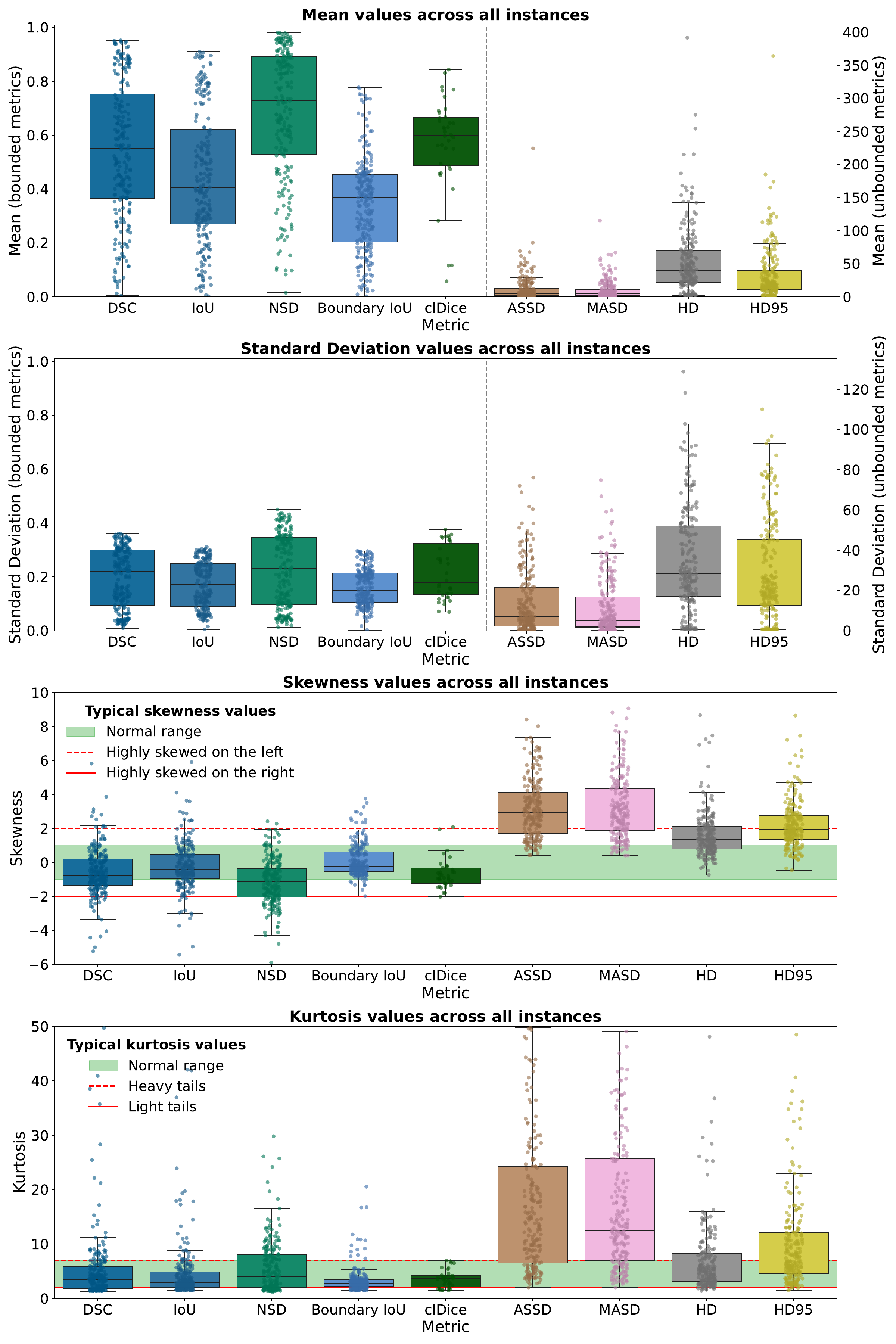}
    
    \caption{\textbf{Mean, standard-deviation (STD), skewness and kurtosis values for different metrics.} One point represents the mean (resp. STD, skewness, kurtosis) of given empirical metric distribution for one benchmark instance. The green interval shows the values for a Gaussian distribution while red lines display values which typically correspond to highly skewed distributions (resp. distribution with very heavy/flat tails).}
     \vspace{0.2ex}
   {\scriptsize \sl \linespread{0.2}\selectfont  Across the paper, boxplots display the median and interquartile (IQR) range (Q1-Q3) across benchmark instances, with whiskers extending to the most extreme values within 1.5 times the IQR range. Points represent individual benchmark instances.
      \par}
    \label{fig:skewness_kurt_segm}
\end{figure}

\newpage
\subsection{Kolmogorov-Smirnov test results}

\begin{table}[ht]
\centering
\resizebox{\textwidth}{!}{%
\begin{tabular}{lcccccccccccc}
\toprule
\textbf{Metric} & \textbf{Normal} & \textbf{Student} & \textbf{Skewnorm} & \textbf{Log-Normal} & \textbf{Logistic} & \textbf{Exponential} & \textbf{Chi2} & \textbf{Beta} & \textbf{Gamma} & \textbf{Pareto} & \textbf{Weibull} \\

\midrule
Boundary IoU & 36\% & 36\% & 36\% & 51\% & 36\% & 95\% & 58\% & 58\% & 40\% & 97\% & 39\% \\
ASSD          & 78\% & 58\% & 66\% & 1\%  & 66\% & 69\% & 58\% & --   & 51\% & 30\% & 37\% \\
clDice        & 53\% & 43\% & 47\% & 70\% & 47\% & 94\% & 74\% & 58\% & 66\% & 98\% & 43\% \\
DSC           & 54\% & 48\% & 44\% & 65\% & 44\% & 98\% & 83\% & 60\% & 76\% & 99\% & 42\% \\
% HD            & 77\% & 66\% & 70\% & 23\% & 70\% & 67\% & 60\% & --   & 37\% & 63\% & 34\% \\
% HD95     & 87\% & 80\% & 83\% & 28\% & 83\% & 82\% & 72\% & --   & 62\% & 50\% & 57\% \\
IoU           & 41\% & 37\% & 37\% & 54\% & 37\% & 96\% & 68\% & 57\% & 56\% & 97\% & 41\% \\
MASD          & 74\% & 52\% & 57\% & 1\%  & 57\% & 63\% & 51\% & --   & 48\% & 34\% & 36\% \\
NSD           & 72\% & 60\% & 60\% & 81\% & 60\% & 97\% & 89\% & 56\% & 83\% & 99\% & 48\% \\
\bottomrule
\end{tabular}
}
\caption{\textbf{Segmentation metrics do not follow any simple parametric distribution.} This table shows the proportion of segmentation benchmark instances that have been rejected by the Kolmogorov-Smirnov test (significance level 5\%, Bonferroni corrected). Each instance has been tested for each segmentation metric (in rows, Boundary IoU, ASSD, clDice, DSC, IoU, MASD, NSD) against some simple parametric distributions in columns (Normal, Student, Skewnorm, Log-Normal, Logistic, Exponential, Chi2, Beta, Gamma, Pareto, Weibull). }
\label{tab:distribution-results}
\end{table}

\newpage
\section{Kernel density estimation algorithm}
\label{sec:kde}
To estimate a continuous density from a finite sample, we performed kernel density estimation (KDE). As a reminder, we first recall the classical KDE framework. We then present our approach, which is closely inspired from~\citetsupp{wilcox2012introductionsup}.

\subsection{Usual framework}
\label{sec:kde-usual}
The usual KDE algorithm is the following: let $X_1, ..., X_n$ be i.i.d. random variables (or their realizations) on $\mathbb{R}^d$. Let $K : \mathbb{R}^d \longrightarrow \mathbb{R}$ be a probability density function (PDF), called kernel.  Let $h_1,...,h_n \in \mathbb{R_+^*}$ be the bandwidths associated with each data point. The KDE estimate is:

\begin{align*}
    \forall x \in \mathbb{R}^d, \hat{f}(x) = \frac{1}{n}\sum_{i=1}^n \frac{1}{h_i^d}K\left(\frac{x-X_i}{h_i}\right)
\end{align*}

\subsubsection{Usual kernels}
The most standard kernel choice is the Gaussian kernel, defined as $K(x) = \frac{1}{(2\pi)^{d/2}}e^{-\frac{\lVert x\rVert^2}{2}}$. However, it is unbounded which is a problem for our metrics. A standard bounded and smooth kernel choice is the Epanechnikov kernel \citepsupp{epanechnikovsup} (also called parabolic kernel), defined as $K(x) = \frac{(d+2)\Gamma(\frac{d}{2}+1)}{2\pi^{d/2}}(1-\lVert x \rVert^2) \mathbbm{1}\{\lVert x \rVert^2\le1\}$.

\subsubsection{Usual bandwidth choices}

Another standard choice is to use the same bandwidth for all data points. The two most common methods for uniform bandwidth computation are Scott's and Silverman's: 

\begin{align*}
    h_{\text{Scott}} &= n^{-1/(d+4)} \, \Sigma^{1/2} \\
    h_{\text{Silverman}} &= \left( \frac{4}{d+2} \right)^{1/(d+4)} n^{-1/(d+4)} \, \Sigma^{1/2}
\end{align*}
where $\Sigma$ is the variance-covariance matrix
of the distribution of $X_1$.

\subsubsection{Usual boundary correction method}

A problem can arise when performing KDE on bounded distributions. In particular, as data points get closer to the domain bounds, the likelihood of mass spilling outside the interval increases. One would then resort to a boundary correction procedure, to ensure the domain of the fitted KDE distribution remains inside the bounds. A classical option is truncation/renormalization: one ignores all the mass that spills outside, and renormalizes the mass inside the interval for each kernel fitted to an individual data point so that it stays equal to $1/n$. However, this implies that individual kernels can lose their symmetry property, and thus the KDE does not preserve the expectation (see proof in Supplementary Section \ref{sec:trunc_renorm}).

\subsection{Chosen framework}

We carefully chose the kernel and the bandwidth estimation algorithm to adapt to the specificities of our data. Our method closely follows \citetsupp{wilcox2012introductionsup}.

We used the Epanechnikov kernel, as it is bounded but smooth around the data point. Regarding bandwidth, choosing a uniform bandwidth was problematic with our data, as the mass concentration was highly variable across distributions. Therefore, we were inspired once again by the adaptive bandwidth algorithm described in \citetsupp{wilcox2012introductionsup}. 

Overall, we made two modifications to the approach presented in \citetsupp{wilcox2012introductionsup}: i) adapt it to multivariate data (for classification); ii) introduce an alternative boundary correction method which preserves the symmetry of individual fitted kernels. This was necessary as some of our metric distributions are inherently bounded, and some data points are located on the boundary. Since we want the distribution to be as close to the original data as possible, keeping the symmetry of individual kernels is necessary. Therefore, the usual boundary correction methods presented above cannot be used for our procedure.
The original method is presented in Algorithm \ref{alg:kde_wilcox_orig}. Our approach is presented in Algorithm \ref{alg:kde_ours}.

\begin{algorithm}[hp]
\caption{Original adaptive Kernel Density Estimation \protect\citepsupp{wilcox2012introductionsup}}
\label{alg:kde_wilcox_orig}
\begin{algorithmic}[1]

    \STATE \textbf{Input:} Data points $X_1, X_2, \dots, X_n \in \mathbb{R}$
    \STATE \textbf{Output:} Adaptive kernel density estimate $\hat{f}(t)$

    \STATE Compute the initial (pilot) estimate via: 
    $$
    \Tilde{f}(X_i)
    $$

    \STATE Define
    $$
    \log g = \frac{1}{n} \sum_{i=1}^n \log \Tilde{f}(X_i)
    $$

    \STATE Compute the local bandwidth modifiers:
    $$
    \lambda_i = \left( \frac{\Tilde{f}(X_i)}{g} \right)^{-a}, \quad 0 \le a \le 1
    $$
    where typically $a = 0.5$

    \STATE Define the span: 
    $$
    h = 1.06 \frac{A}{n^{1/5}}, \quad 
    A = \min(s, \text{IQR}/1.34)
    $$
    where $s$ is the standard deviation and IQR is the interquartile range

    \STATE Define the kernel function:
    $$
    K(t) =
    \begin{cases} 
    \dfrac{3}{4}\left(1 - \dfrac{1}{5}t^2\right)/\sqrt{5}, & |t| < \sqrt{5}, \\[6pt]
    0, & \text{otherwise.}
    \end{cases}
    $$

    \STATE Compute the adaptive kernel estimate:
    $$
    \Hat{f}(t) = \frac{1}{n} \sum_{i=1}^n \frac{1}{h \lambda_i} 
    K \left( h^{-1} \lambda_i^{-1} (t - X_i) \right)
    $$

\end{algorithmic}
\end{algorithm}

\begin{algorithm}[hp]
\caption{Our adaptive Kernel Density Estimation}
\label{alg:kde_ours}
\begin{algorithmic}[1]

    \STATE \textbf{Input:} Data points $X_1, X_2, \dots, X_n \in \mathbb{R}^d$, Data boundaries $a,b \in \mathbb{R}^d$
    \STATE \textbf{Output:} Adaptive kernel density estimate $\hat{f}(x)$

    \STATE Define the initial bandwidth for all data points: 
    $$
    \forall i, \Tilde{h}_i = 1.06\Sigma^{1/2}n^{-1/5} \in \mathcal{M}_d(\mathbb{R})
    $$
    where $\Sigma$ is the empirical variance-covariance matrix of $X_1,...,X_n$

    \STATE Compute the initial (pilot) estimate: 
    $$
    \Tilde{f}(X_i) = \frac{1}{n}\sum_{j=1}^n \frac{1}{\Tilde{h}_j}K\left(\frac{X_i-X_j}{\Tilde{h}_j}\right)
    $$

    \STATE Define
    $$
    \log g = \frac{1}{n} \sum_{i=1}^n \log \Tilde{f}(X_i)
    $$

    \STATE Compute the local bandwidth modifiers:
    $$
    \lambda_i = \left( \frac{\Tilde{f}(X_i)}{g} \right)^{-1/2}
    $$

    \STATE Trim the modifiers to prevent putting mass outside the data domain:
    $$
    \alpha_i = \min(\min(\lVert X_i-a\rVert, \lVert X_i -b \rVert), \lambda_i)
    $$

    \STATE Compute adapted bandwidths:
    $$
    \forall i, h_i = \alpha_i\Tilde{h}_i
    $$

    \STATE Define the kernel function:
    $$
    K(x) = \frac{(d+2)\Gamma(\frac{d}{2}+1)}{2\pi^{d/2}}(1-\lVert x \rVert^2) \mathbbm{1}\{\lVert x \rVert^2\le1\}
    $$

    \STATE Compute the adaptive kernel estimate:
    $$
    \Hat{f}(x) = \frac{1}{n} \sum_{i=1}^n \frac{1}{h \alpha_i} 
    K \left( h^{-1} \alpha_i^{-1} (x - X_i) \right)
    $$

\end{algorithmic}
\end{algorithm}

\newpage
\subsection{Faithfulness of KDE representation of samples for segmentation metrics distribution}
\label{sec:kde_segm_faith}
\begin{figure}[H]
    \centering
    \includegraphics[width=0.9\linewidth]{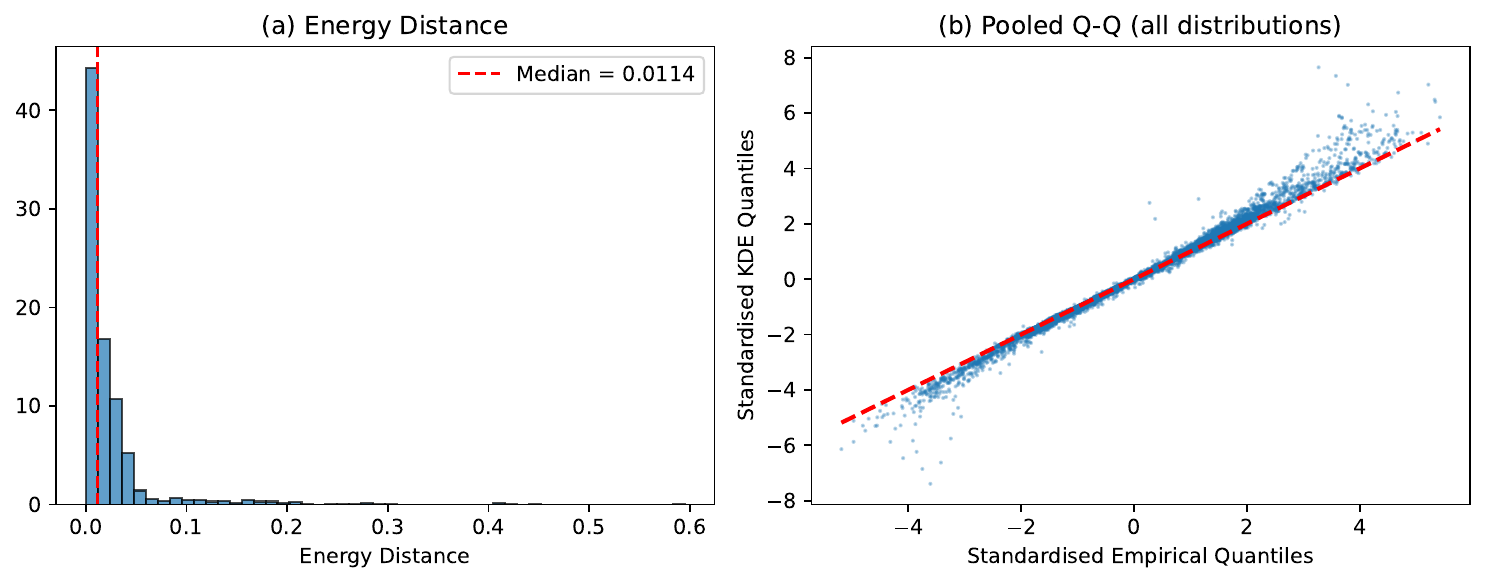}
    \caption{\textbf{Faithfulness of KDE representation of samples for segmentation metrics distribution.} Panel (a): Energy distance between the KDE distribution and the samples. Panel (b): QQ-plot between KDE and empirical cumulative distribution functions. 1 point = 1 instance.}
    \label{fig:kde_faithfulness}
\end{figure}

To assess the faithfulness of the representation of the samples by the proposed KDE procedure, we used energy distance to quantify the discrepancy between estimated and sample distributions. The energy distance between two distributions having cumulative distribution functions $U$ and $V$ is defined as $\Big(2\int(U-V)^2\Big)^{1/2}$. Further, we compared the quantiles of estimated and original cumulative distribution functions.  Results are shown  on Supplementary Figure \ref{fig:kde_faithfulness}. One can see that the median energy distance is 0.011 (Supplementary Figure \ref{fig:kde_faithfulness} (a)), indicating close agreement between samples and KDE distributions. The pooled standardized quantile-quantile plot (Supplementary Figure \ref{fig:kde_faithfulness} (b)) shows excellent alignment within $\pm 2$ standard deviations; with increased scatter beyond $\pm 3$ standard deviations which is attributable to sampling variability in the tails of the original data rather than systematic KDE bias. Overall, these results indicate that the KDE distributions are good approximations of the original sample distributions.

\subsection{Preservation of properties relevant to AUC and AP by the KDE}
\label{sec:kde_class_assessment}
\begin{figure}[H]
    \centering
    \includegraphics[width=0.9\linewidth]{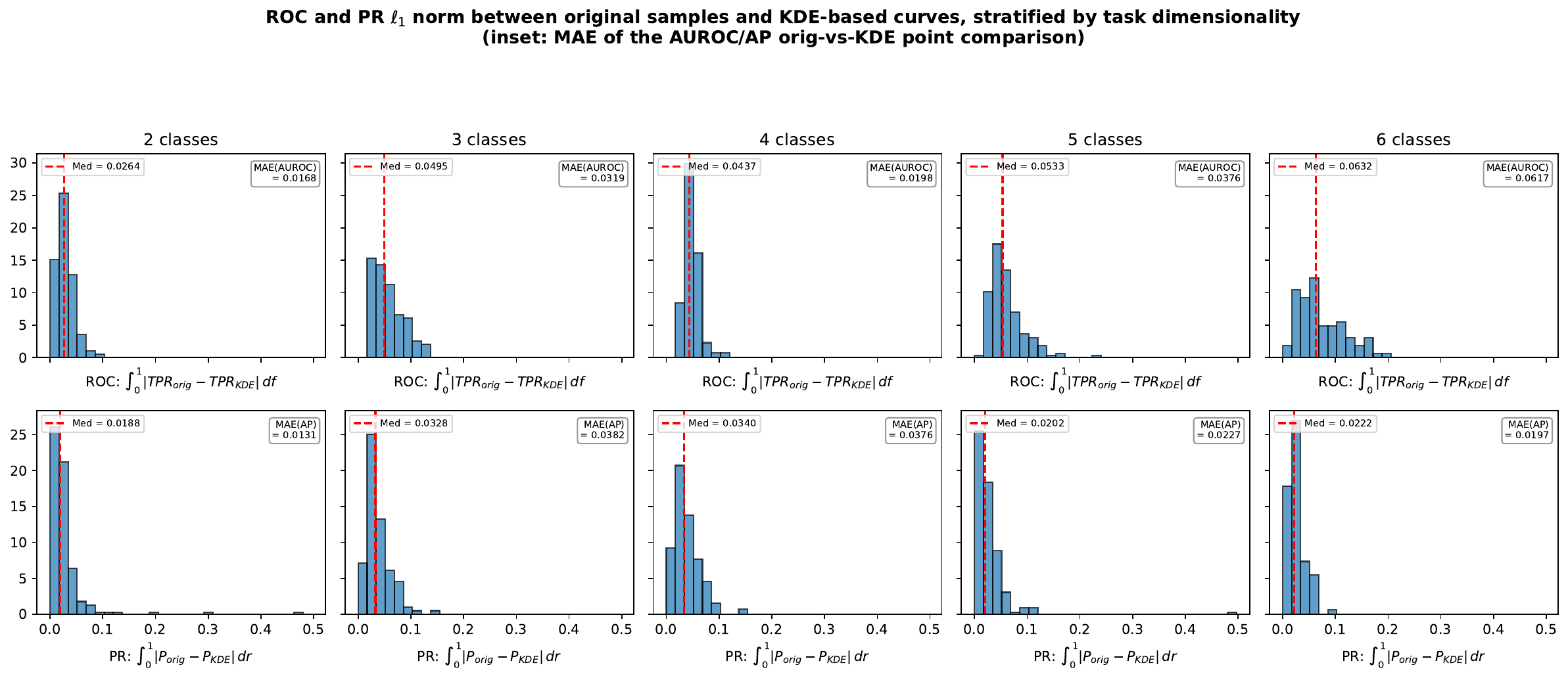}
    \caption{\textbf{KDE preservation of per-class ROC and PR curves across  studied dimensionality regimes.} First row: $l_1$ norm between per-class ROC curves computed from KDE and original samples, stratified by dimensionality. 1 observation = 1 class. Second row: $l_1$ norm between per-class PR curves computed from KDE and original samples, stratified by dimensionality.}
    \label{fig:kde_curve_preservation}
\end{figure}

To analyze whether properties relevant to AUC and AP are preserved by the KDE sampling, we computed the per-class receiver operating characteristic (ROC) and precision-recall (PR) curves for each instance, computed from the original samples and from the KDE. We computed the $l_1$ norm between the original and KDE curves and plotted the histogram across all curves corresponding to each dimensionality regime. The rationale behind investigating this quantity is that PR and ROC curves directly depend on the ranking of positive samples. Results are shown in Figure \ref{fig:kde_curve_preservation}. We also computed the mean absolute error (MAE) for AUC and AP. The MAE values are indicated inside each plot. As one can see, there is no increase of errors as dimension increases, and differences remain small across all dimensions. Therefore, this analysis hints towards the KDE distributions preserving the properties of original samples.

\section{Number of samples required for coverage estimation}
\label{sec:n_samples}
We aim to estimate a true coverage value, i.e. the probability that the CI contains the true summary statistic $p \in [0,1]$.  Here, we  determine the number of samples $n$ required to estimate $p$ with a given margin of error $\varepsilon$ at a 95\% confidence level. We set an acceptable error at $\varepsilon = 0.01$. We determine $n$  using a CI\footnote{Importantly, note that this is the CI of the coverage of the CI}.  $\hat{p}$ is a binomial proportion and a CI of the coverage can be computed as\footnote{Such interval is not ideal when $\hat{p}$ approaches 1 or when $n$ is small but this is not the case here.}:
$$
\hat{p} \pm z \cdot \sqrt{\frac{\hat{p} (1 - \hat{p} )}{n}}
$$

To ensure that the margin of error is less than 0.01, we solve:

$$
z \cdot \sqrt{\frac{\hat{p} (1 - \hat{p} )}{n}} < 0.01
$$

Using $z = 1.96$ (for 95\% confidence), we derive the minimum required sample size:

$$
n > \frac{z^2 \cdot \hat{p} (1 - \hat{p} )}{\varepsilon^2} = \frac{(1.96)^2 \cdot \hat{p} (1 - \hat{p} )}{0.0001} = \frac{3.8416 \cdot \hat{p} (1 - \hat{p} )}{0.0001} = 38416 \cdot \hat{p} (1 - \hat{p} )
$$

Supplementary Table~\ref{tab:sample_size_coverage} shows $n$ for different values of $\hat{p} $. We consider the worst-case scenario which is $\hat{p} =0.50$ which corresponds to $n=9604$ simulations. We rounded it to $n=10,000$. This is clearly an upper bound as one can reasonably argue that being precise at $\hat{p} =0.50$ is less important than for higher empirical coverage values. In practice, users who want to compute coverage could settle for lower values (e.g. $3,500$). 

\begin{table}[H]
\begin{center}
\begin{tabular}{@{}ccc@{}}
\toprule
\textbf{Coverage $\hat{p}$} & \textbf{Variance $\hat{p}(1-\hat{p})$} & \textbf{Required Sample Size \( n \)} \\
\midrule
0.50 & 0.25   & 9604 \\
0.80 & 0.16   & 6147 \\
0.90 & 0.09   & 3457 \\
0.95 & 0.0475 & 1825 \\
0.99 & 0.0099 & 380  \\
\bottomrule
\end{tabular}
\caption{\textbf{Sample size for coverage computation.} Determination of the number of samplings from the true distribution needed to achieve a coverage computation error less than $\varepsilon = 0.01$.}
\label{tab:sample_size_coverage}
\end{center}
\end{table}

\newpage
\section{Classification: experimental data and  characterization}

\subsection{Dataset characteristics}

\begin{table}[hbpt]
\centering
\scriptsize
\begin{tabularx}{\textwidth}{l c c *{6}{>{\centering\arraybackslash}X}}

\toprule
\textbf{Task} & \textbf{n} & \textbf{\#Classes} & \textbf{Class 1} & \textbf{Class 2} & \textbf{Class 3} & \textbf{Class 4} & \textbf{Class 5} & \textbf{Class 6} \\
\midrule
CheXpert Enlarged Cardiomediastinum & 7624 & 2 &
\begin{tabular}[c]{@{}c@{}}2224 \\ (29\%)\end{tabular} &
\begin{tabular}[c]{@{}c@{}}5400 \\ (71\%)\end{tabular} & -- & -- & -- & -- \\
APTOS19 Blindness Detection & 733 & 5 &
\begin{tabular}[c]{@{}c@{}}361 \\ (49\%)\end{tabular} &
\begin{tabular}[c]{@{}c@{}}74 \\ (10\%)\end{tabular} &
\begin{tabular}[c]{@{}c@{}}200 \\ (27\%)\end{tabular} &
\begin{tabular}[c]{@{}c@{}}39 \\ (5\%)\end{tabular} &
\begin{tabular}[c]{@{}c@{}}59 \\ (8\%)\end{tabular} & -- \\
Kaggle Cataract Dataset & 121 & 4 &
\begin{tabular}[c]{@{}c@{}}60 \\ (50\%)\end{tabular} &
\begin{tabular}[c]{@{}c@{}}20 \\ (17\%)\end{tabular} &
\begin{tabular}[c]{@{}c@{}}21 \\ (17\%)\end{tabular} &
\begin{tabular}[c]{@{}c@{}}20 \\ (17\%)\end{tabular} & -- & -- \\
ISIC20 & 6541 & 2 &
\begin{tabular}[c]{@{}c@{}}6424 \\ (98\%)\end{tabular} &
\begin{tabular}[c]{@{}c@{}}117 \\ (2\%)\end{tabular} & -- & -- & -- & -- \\
COVID-CT-Dataset & 150 & 2 &
\begin{tabular}[c]{@{}c@{}}80 \\ (53\%)\end{tabular} &
\begin{tabular}[c]{@{}c@{}}70 \\ (47\%)\end{tabular} & -- & -- & -- & -- \\
Dataset of Breast Ultrasound Images & 156 & 3 &
\begin{tabular}[c]{@{}c@{}}87 \\ (56\%)\end{tabular} &
\begin{tabular}[c]{@{}c@{}}42 \\ (27\%)\end{tabular} &
\begin{tabular}[c]{@{}c@{}}27 \\ (17\%)\end{tabular} & -- & -- & -- \\
CheXpert Pleural Effusion & 24318 & 2 &
\begin{tabular}[c]{@{}c@{}}7080 \\ (29\%)\end{tabular} &
\begin{tabular}[c]{@{}c@{}}17238 \\ (71\%)\end{tabular} & -- & -- & -- & -- \\
DeepDRiD Quality & 241 & 2 &
\begin{tabular}[c]{@{}c@{}}125 \\ (52\%)\end{tabular} &
\begin{tabular}[c]{@{}c@{}}116 \\ (48\%)\end{tabular} & -- & -- & -- & -- \\
MURA Forearm & 366 & 2 &
\begin{tabular}[c]{@{}c@{}}233 \\ (64\%)\end{tabular} &
\begin{tabular}[c]{@{}c@{}}133 \\ (36\%)\end{tabular} & -- & -- & -- & -- \\
AIDA-E Barrett’s Esophagus Diagnosis & 53 & 3 &
\begin{tabular}[c]{@{}c@{}}6 \\ (11\%)\end{tabular} &
\begin{tabular}[c]{@{}c@{}}35 \\ (66\%)\end{tabular} &
\begin{tabular}[c]{@{}c@{}}12 \\ (23\%)\end{tabular} & -- & -- & -- \\
Derm7pt & 127 & 5 &
\begin{tabular}[c]{@{}c@{}}6 \\ (5\%)\end{tabular} &
\begin{tabular}[c]{@{}c@{}}31 \\ (24\%)\end{tabular} &
\begin{tabular}[c]{@{}c@{}}12 \\ (9\%)\end{tabular} &
\begin{tabular}[c]{@{}c@{}}72 \\ (57\%)\end{tabular} &
\begin{tabular}[c]{@{}c@{}}6 \\ (5\%)\end{tabular} & -- \\
HyperKvasir Anatomical Landmarks & 823 & 6 &
\begin{tabular}[c]{@{}c@{}}202 \\ (25\%)\end{tabular} &
\begin{tabular}[c]{@{}c@{}}2 \\ (0\%)\end{tabular} &
\begin{tabular}[c]{@{}c@{}}79 \\ (10\%)\end{tabular} &
\begin{tabular}[c]{@{}c@{}}200 \\ (24\%)\end{tabular} &
\begin{tabular}[c]{@{}c@{}}153 \\ (19\%)\end{tabular} &
\begin{tabular}[c]{@{}c@{}}187 \\ (23\%)\end{tabular} \\
\bottomrule
\end{tabularx}
\caption{{\bf \sffamily Dataset statistics for different classification tasks.} Please refer to \protect\citepsupp{godau2024beyondsupp} for a description of tasks. $n$ is the total sample size (across all classes) of the test sets used to compute CIs.}
\label{tab:classif}
\end{table}

\subsection{Diversity of distributions}

Definitions given in Section~\ref{subsec:diversity_segm} are for 1D random variables, but can be extended to the multivariate case \citepsupp{mori1994multivariatesup}. Let us denote $X = (X_1,...,X_d)$, $\mu$ is the mean vector of $X$ and $\Sigma$ is the variance-covariance matrix of $X$. Let us consider the standardized vector $Y = (X-\mu)\Sigma^{-1/2}$. We define the multivariate skewness $\gamma$ and the multivariate kurtosis $\kappa$ as follows: 
$$
\gamma = \mathbb{E}\left[\sum_{i=1}^d \sum_{j=1}^d \sum_{k=1}^d Y_i Y_j Y_k\right], \quad
\kappa = \mathbb{E}\left[\lVert Y\rVert_2^4\right]
$$

\begin{figure}[ht]
    \centering
    \includegraphics[width=0.9\linewidth]{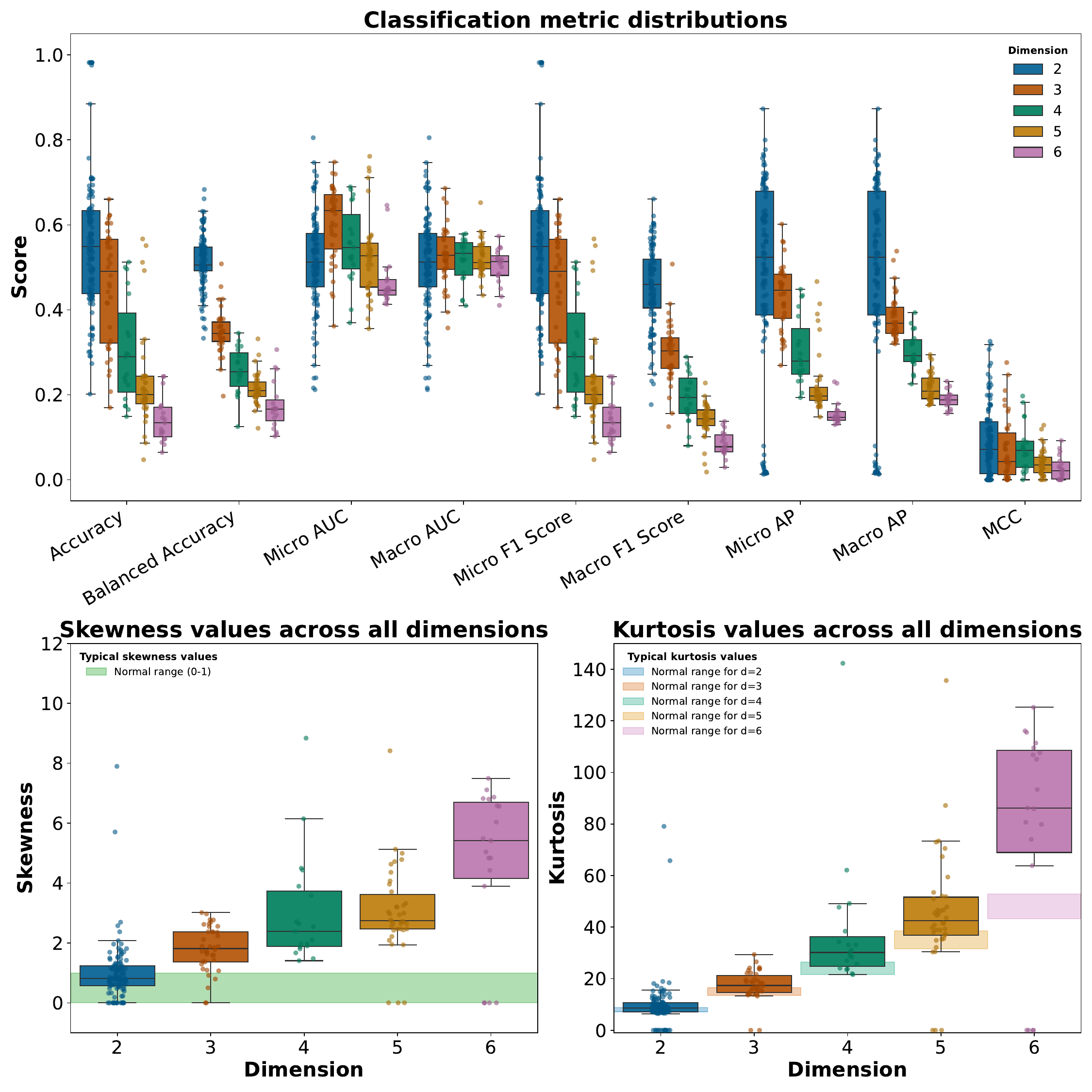}
    
    \caption{\textbf{Classification performance metrics and multivariate skewness and kurtosis values across number of classes.} One point represents the metric values,  the skewness and the kurtosis  of a given benchmark instance. For skewness and kurtosis, the shaded regions represent the values for a multivariate Gaussian distribution (for kurtosis, they depend on the dimension)}
    \label{fig:skewness_kurt_classif}
\end{figure}

Results are presented on Supplementary Figure~\ref{fig:skewness_kurt_classif}. 

\vspace{10cm}

\newpage

\subsection{Kolmogorov-Smirnov test results}
\begin{table}[ht]
\centering
%\resizebox{\textwidth}{!}{%
\begin{tabular}{lcccccc}
\toprule
\textbf{Distribution} & \textbf{Normal} & \textbf{Student} & \textbf{Skewnorm} &  \textbf{Logistic}  &\textbf{Pareto}  \\

\midrule
Proportion of reject& 73\% & 59\% & 62\% & 58\% & 100\% \\
\bottomrule
\end{tabular}
%}

\caption{\textbf{The logits don't follow any simple parametric distribution.} This table shows the proportion of classification instances that have been rejected by the Kolmogorov Smirnov test (significance level of 5\%, Bonferroni corrected) against some simple parametric distributions (in columns, Normal, Student, Skewnorm, Logistic and Pareto). %We restricted the distributions to those that cover the full range $[-\infty, \infty]$. 
A given instance logit distribution is considered rejected if at least one of its marginal class distributions is rejected. }
\label{tab:distribution-results-classif-bonferonni}
\end{table}

\section{Coverage convergence pace (CCP)}
\label{sec:ccp}

\subsection{CCP definition and estimation}
\label{sec:ccp_def}
Let $Y$ be the empirical coverage value of a given instance at a given sample size $n$. Empirical coverage for two-sided intervals converges to theoretical coverage at rate $1/n+O(n^{-2})$ \citepsupp{hall1992bootstrapsup}. Thus coverage gets closer to $0.95$ as the sample size increases, at an approximate rate of $1/n$. Accordingly, we consider the following model:
\begin{equation}
    Y = 0.95 +\frac{\beta}{n}
    \label{eq: ccp}
\end{equation}
We define the coverage convergence pace (CCP) as the parameter $\beta$. Its empirical estimate, denoted by $\hat{\beta}$ is obtained by fitting the above model to observed coverage values. The model is fitted using a simple ordinary least squares uni-variate linear regression.

\begin{figure}[H]
    \centering
    \includegraphics[width=0.75\linewidth]{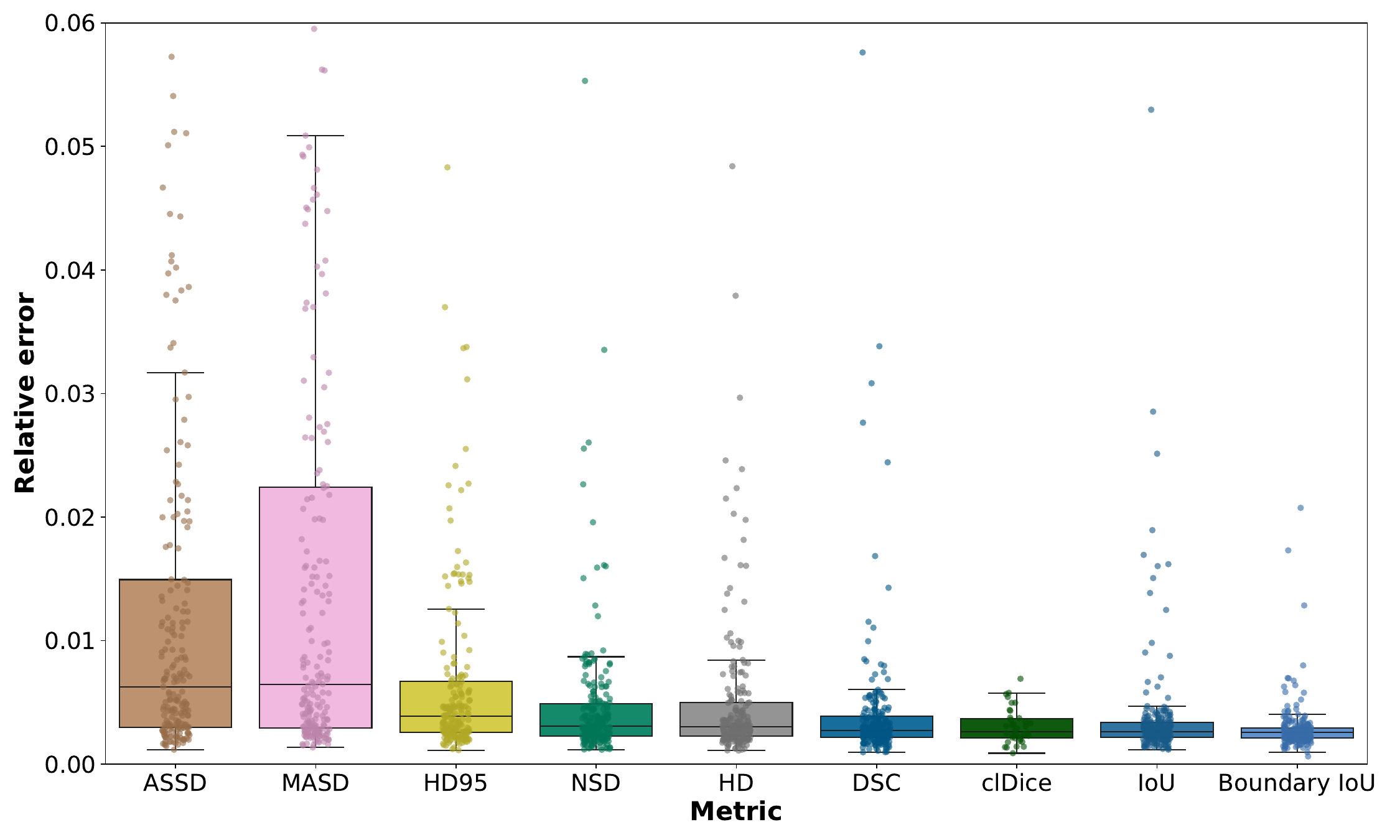}
    \caption{\textbf{Relative errors for fitted CCP values.} One point corresponds to the fitted CCP of one instance and the colors represent the metrics (DSC, NSD, IoU, Boundary IoU, HD, 95HD, ASSD, MASD). }
    \label{fig:relative_errors_ccp}
\end{figure}

The relative error for the model is the following: $$\sqrt{\frac{\sum_{i}\left(Y_i-\hat{Y}_i\right)^2}{\sum_iY_i^2}}$$ where $\hat{Y}=0.95 +\frac{\hat{\beta}}{n}$. 
Figure~\ref{fig:relative_errors_ccp} displays the relative errors obtained when fitting the model to each segmentation metric for percentile CI of the mean. One can see that the relative errors are small.

\section{Statistical testing of the claims}
\label{sec:tests}

% Here, we briefly describe the statistical tests used to assess differences in CCP.

We performed statistical testing for each of the main claims of the results section. We used permutation tests and performed False Discovery Rate correction for multiple testing across the full statistical testing landscape of the paper (378 tests).

In this section, we first describe the statistical testing procedure (Supplementary Section~\ref{subsec:stat_test_proc}) and then the results (Supplementary Section~\ref{subsec:stat_test_res}). 

\subsection{Statistical testing procedure}
\label{subsec:stat_test_proc}
\subsubsection{Paired one-sided permutation test}

Let $m_1$ and $m_2$ be the two parameters to compare (for example, $m_1=$ percentile method vs $m_2=$ BCa method and we want to compare their coverage). 
Let $X_1, \ldots, X_{228} $ be the values (i.e. coverage, width or CCP values) obtained for each instance for $m_1$ and $Y_1, \ldots, Y_{228} $ be the value (i.e. coverage, width or CCP values) obtained for each instance for  $m_2$. 

Let $D_i=X_i-Y_i$ difference between the observations from $m_1$ and $m_2$ for instance $i$.  

The null hypothesis is 
$H_0:
\mathbb{E}\left(D\right)=0
$.
A natural test statistic is the mean difference:
$$
T_{\text{obs}} = \frac{1}{n} \sum_{i=1}^n D_i
$$

Under the null hypothesis \(H_0\), the sign of each paired difference \(D_i\) is arbitrary: if there is no difference between the values, then flipping a difference from \(D_i\) to \(-D_i\) should not change the statistical behavior of the data. Thus, for each pair, one randomly chooses to keep or flip the sign.
Formally, for each permutation (or “sign flip"):
$$
D_i^{*} = s_i D_i,
\quad \text{where } s_i = 
\begin{cases}
+1 & \text{with probability } 1/2, \\
-1 & \text{with probability } 1/2.
\end{cases}
$$

For each permutation, one obtains the permuted test statistic
$$
T^{*} = \frac{1}{n} \sum_{i=1}^n D_i^{*}.
$$

Repeating the random sign-flipping many times yields the permutation
distribution of the test statistic under \(H_0\).  
For the one-sided alternative
$$
H_1:\mathbb{E}(D)>0,
$$
the \(p\)-value is
$$
p=
\frac{
\#\left\{T^{*}\ge T_{\mathrm{obs}}\right\}
}{
100\,000
}
$$
with $100\,000$ permutations.

\subsubsection{Unpaired permutation test}
\label{sec:diff_segm_classif}

Let two independent samples:
$$
X_1, X_2, \ldots, X_{n_1} \qquad \text{and} \qquad
Y_1, Y_2, \ldots, Y_{n_2},
$$
representing, the observations of $m_1$ across $n_1$ instances and the observations of $m_2$ across $n_2$ instances.

The null hypothesis is $
H_0: \mathbb{E}\left(X\right) = \mathbb{E}\left(Y\right)
$. A natural test statistic is:
$$
T_{\text{obs}} = \bar{X} - \bar{Y}
= \frac{1}{n_1}\sum_{i=1}^{n_1} X_i
  - \frac{1}{n_2}\sum_{j=1}^{n_2} Y_j .
$$

Under the null hypothesis $H_0$, all observations come from the same distribution.  
Therefore, the group labels (“X" vs.\ “Y") are arbitrary.  
We thus simulate the behavior of the test statistic under $H_0$ by randomly reassigning the labels, leading to the permuted test statistic
    $$
    T^{*} = \bar{X}^{*} - \bar{Y}^{*}.
    $$

Repeating this process many times yields the permutation distribution of the statistic under the null hypothesis.

For the one-sided alternative
$$H_1:\mathbb{E}(D)>0,
$$
the \(p\)-value is
$$
p=
\frac{
\#\left\{T^{*}\ge T_{\mathrm{obs}}\right\}
}{
100\,000
}
$$
with $100\,000$ permutations.

\subsubsection{Correction for multiple testing}
We corrected for multiple testing using the False Discovery Rate \citep{benjamini1995controlling} across the full testing landscape of the paper (3090 tests in total).

\newpage

\subsection{Statistical testing results}
\label{subsec:stat_test_res}
\subsubsection{Basic almost always underperforms}

We are testing whether the basic bootstrap method produces CIs with worse coverage than other bootstrap methods. 

\begin{figure}[H]
    \centering
    \includegraphics[width=\linewidth]{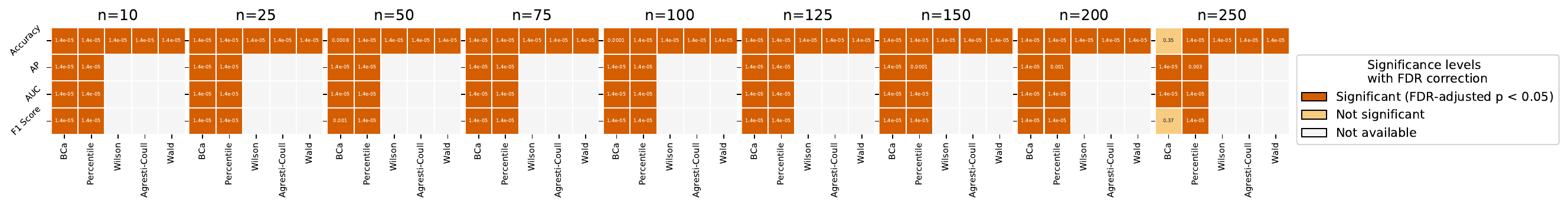}
    \caption{Significance matrix for the claim that basic bootstrap has worst coverage than other bootstrap methods for micro classification metrics. P-values are computed using paired permutation tests and corrected using FDR.}
    \label{fig:test_basic_micro}
   
\end{figure}

\begin{figure}[H]
    \centering
    \includegraphics[width=\linewidth]{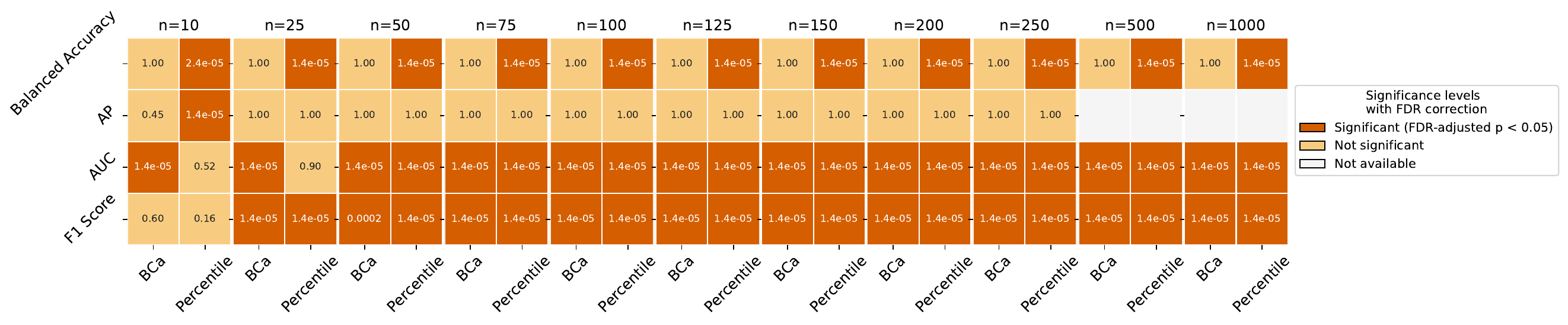}
    \caption{Significance matrix for the claim that basic bootstrap has worst coverage than other bootstrap methods for macro classification metrics. P-values are computed using paired permutation tests and corrected using FDR.}
    \label{fig:test_basic_macro}
   
\end{figure}

For each sample size, we tested where the coverage of a CI produced by the basic bootstrap method across all instances is lower that procedure by the BCa or the percentile method, respectively. We tested for each classification micro and macro metrics and all segmentation metrics for the CIs of all summary statistics. We used paired permutation test because the instances are the same. 
P-values are reported in Figures  \ref{fig:test_basic_micro}, \ref{fig:test_basic_macro} and \ref{fig:test_basic_segm}.  We did 9 sample sizes x 9 segmentation metrics x (2 bootstrap methods  x 4 summary statistics + 4 CI methods) +  9 sample sizes x (7 classification metrics x 2 bootstrap methods + 1 classification metric x 6 CI methods) = 1152 tests. 2+32+77=111/1152=9\% of tests were not significant, supporting our claim that BCA is the worst in the vast majority of cases.

\begin{figure}[H]
    \centering
    \includegraphics[width=4cm, height=20cm]{supp_figs/tests/cov_basic_segm.pdf}
    \caption{Significance matrix for the claim that basic bootstrap has worst coverage than other bootstrap methods for segmentation metrics. P-values are computed using paired permutation tests and corrected using FDR.}
    \label{fig:test_basic_segm}
   
\end{figure}
\newpage 
\subsubsection{BCa is better than percentile for the mean and for some classification metrics}
We are testing whether the coverage of the BCa bootstrap method is higher than that of the percentile method.
For each sample size, we tested if the coverage of CIs produced using the BCa method is higher than the percentile method for CIs of the mean across all segmentation metrics and as well as for micro AP and AUC. We used a paired permutation tests.

\begin{figure}[H]
    \centering
    \includegraphics[width=\linewidth]{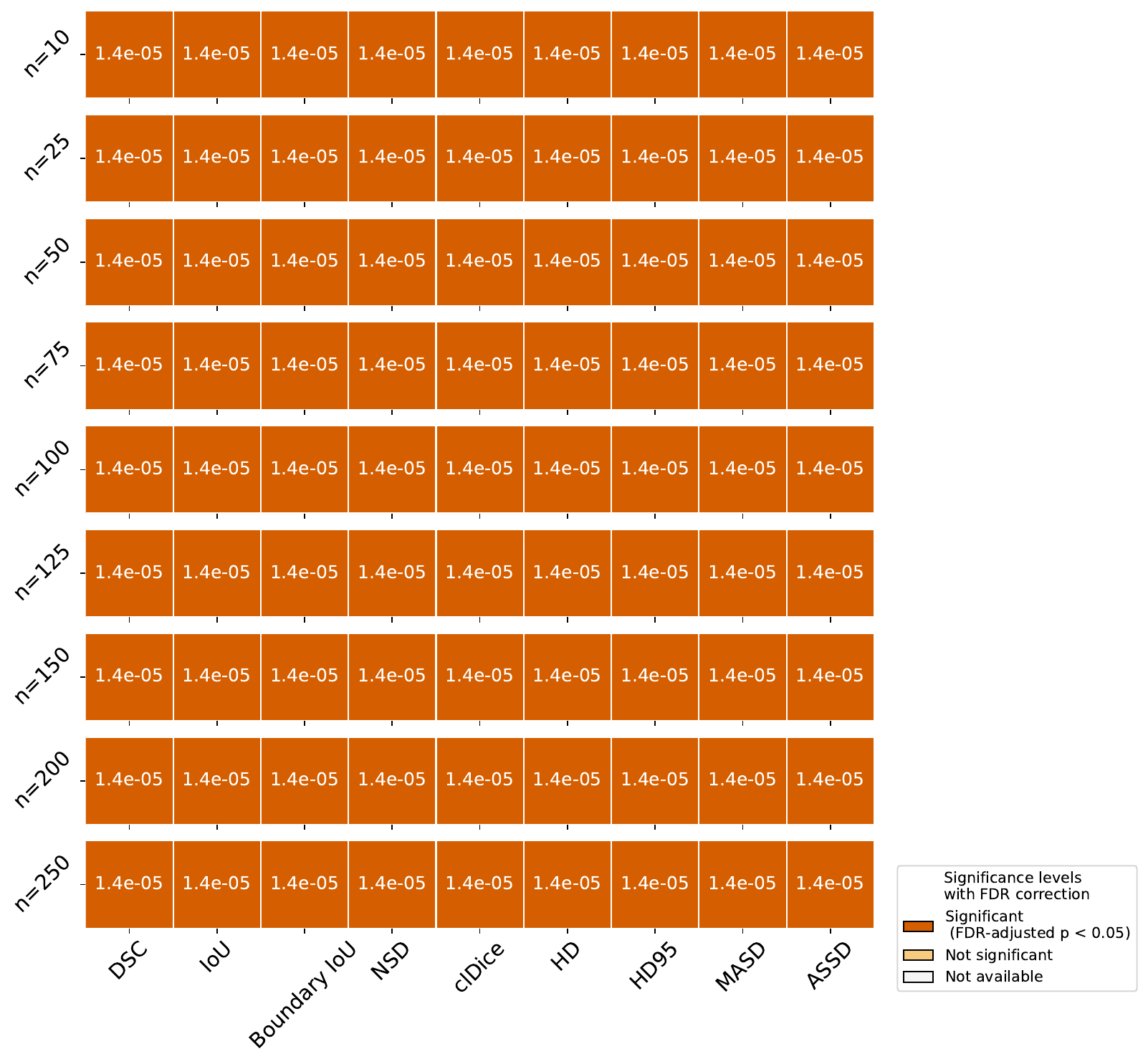}
    \caption{Significance matrix for the claim that the coverage of the BCa bootstrap is higher than percentile's for segmentation metrics. P-values were obtained using paired permutation tests and corrected using FDR. }
    \label{fig:test_bca_segm}
\end{figure}

\begin{figure}[H]
    \centering
    \includegraphics[width=0.7\linewidth]{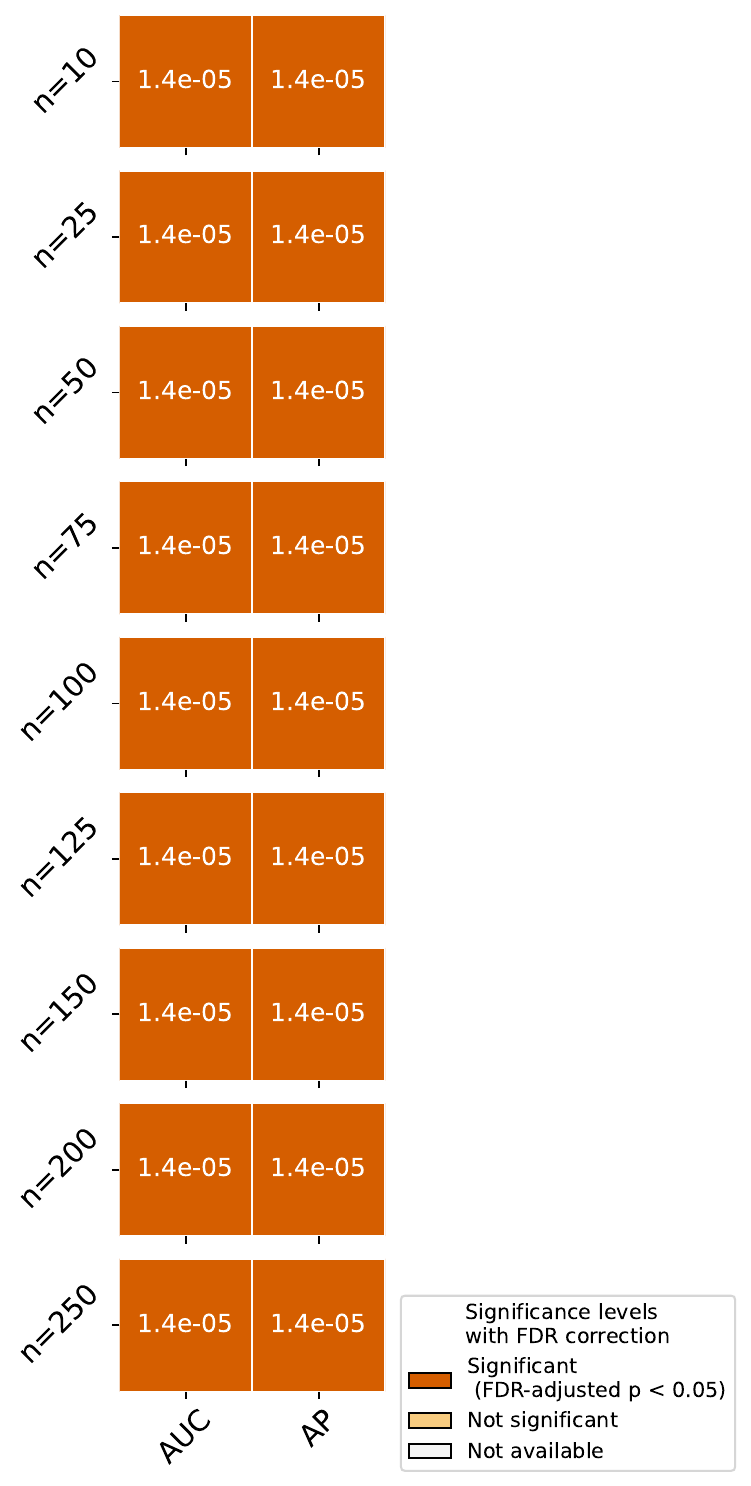}
    \caption{Significance matrix for the claim that the coverage of the BCa bootstrap is higher than percentile's for micro AUC and micro AP classification metrics. P-values were obtained using paired permutation tests and corrected using FDR.}
    \label{fig:test_bca_classif}
\end{figure}

P-values are reported in Figures \ref{fig:test_bca_segm} and \ref{fig:test_bca_classif}. We did 9 sample sizes x (9 segmentation metrics + 2 classification metrics)=99 tests. All tests were significant.

\newpage

\newpage

\subsubsection{Parametric t has a better coverage than bootstrap methods on small samples}

We tested if the coverage of parametric t is higher than that of BCa, percentile and basic bootstrap methods for CIs of the mean across all segmentation metrics. All tests were done for all sample sizes, even though our claims were on small samples, and we used paired permutation tests across all segmentation samples. 

\begin{figure}[H]
    \centering
    \includegraphics[width=0.9\linewidth]{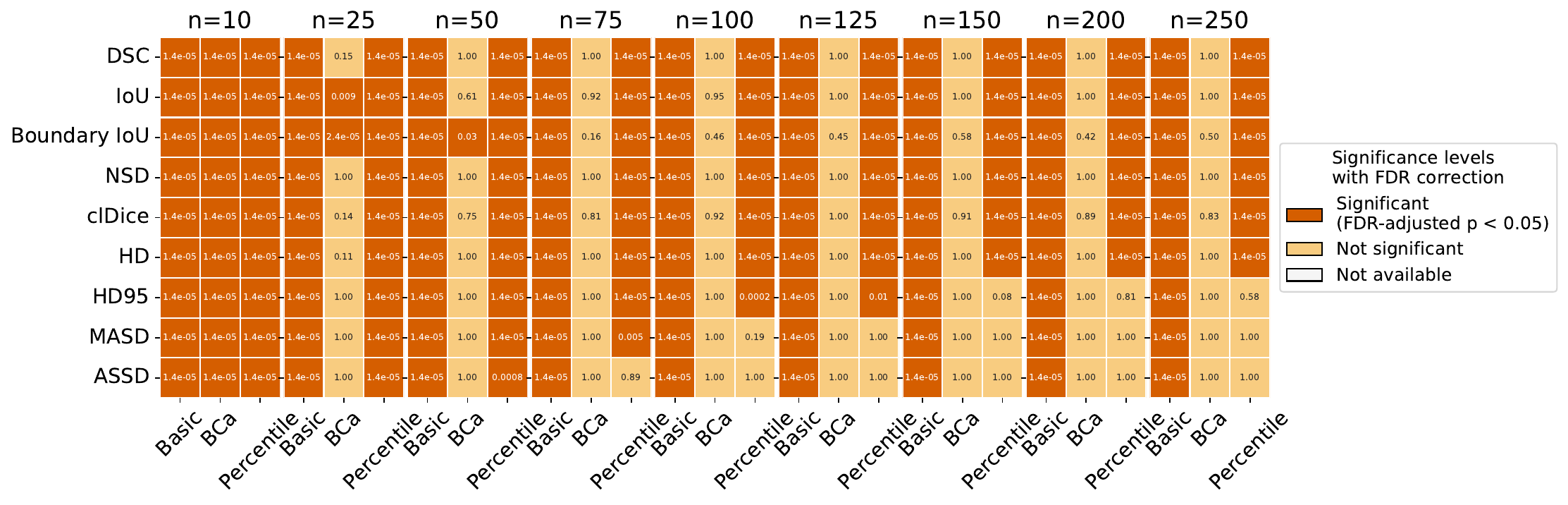}
    \caption{Significance matrix for the claim that parametric t method has better coverage than other bootstrap methods for segmentation metrics. P-values were obtained using paired permutation tests and corrected using FDR.}
    \label{fig:test_param_boot_segm}
\end{figure}

P-values are reported in Figure  \ref{fig:test_param_boot_segm}. For the coverage of segmentation, we did 9 sample sizes x 3 bootstrap methods x 9 metrics = 243 tests. Tests were significant in very small samples (all cases for n=10, only some compared to BCa for n=25, n=50) which is consistent with our claim. For larger values of n, the difference is never significant compared to BCa and the difference is very small, again supporting our claim.

\subsubsection{Parametric t produces larger intervals than bootstrap methods}

We tested if the CIs of the mean produced by parametric t were larger than those of bootstrap methods. We used paired permutation tests across all  segmentation samples. 

\begin{figure}[H]
    \centering
    \includegraphics[width=0.9\linewidth]{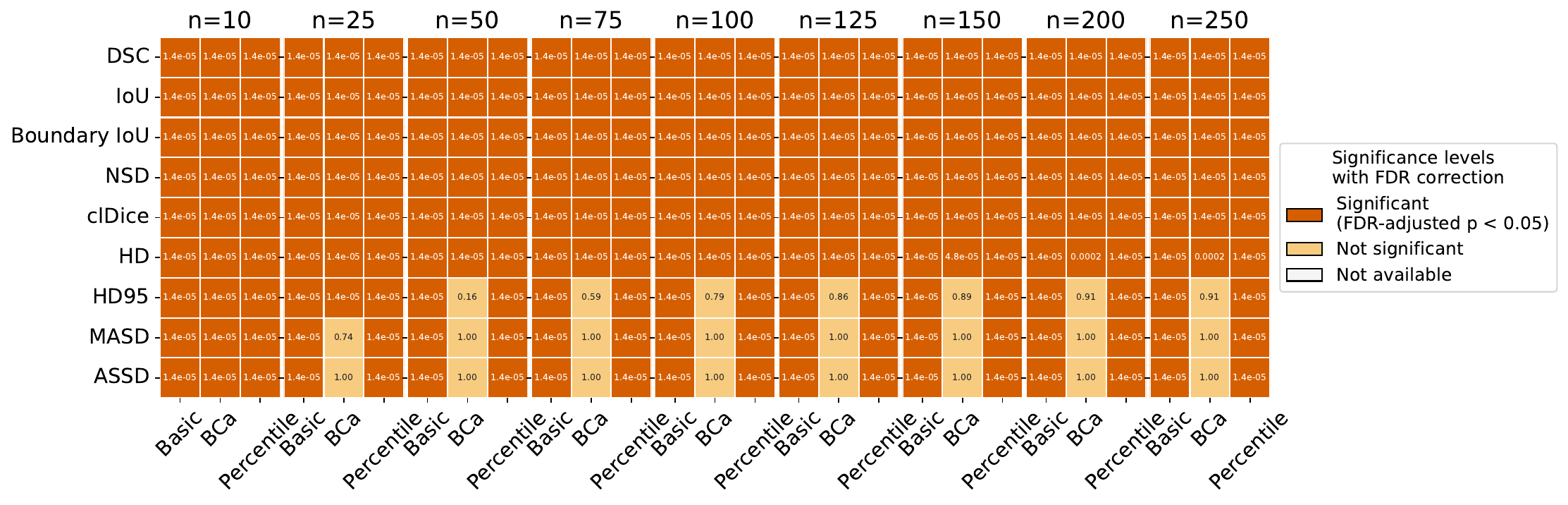}
    \caption{Significance matrix for the claim that parametric t method produces wider CIs than other bootstrap methods for segmentation metrics. P-values were obtained using paired permutation tests and corrected using FDR. }
    \label{fig:test_width_param_boot_segm}
\end{figure}

P-values are reported in Figure \ref{fig:test_width_param_boot_segm}. For the width of segmentation, we did  9 sample sizes x 3 bootstrap methods x 9 metrics = 243 tests. 23/243=9\% of tests are not significant, all when testing parametric t vs BCa and for ASSD, MASD or HD95 metrics. Again, this is compatible with our claim.

\newpage

\subsubsection{Wilson outperforms bootstrap methods on small samples}

We tested if the Wilson method produces CIs with a better coverage than BCA, percentile and basic bootstrap methods for CIs of the accuracy. All tests were done for all sample sizes, even though our claims were on small samples, and we used paired permutation tests across all classification and samples. 

\begin{figure}[H]
    \centering
    \includegraphics[width=0.4\linewidth]{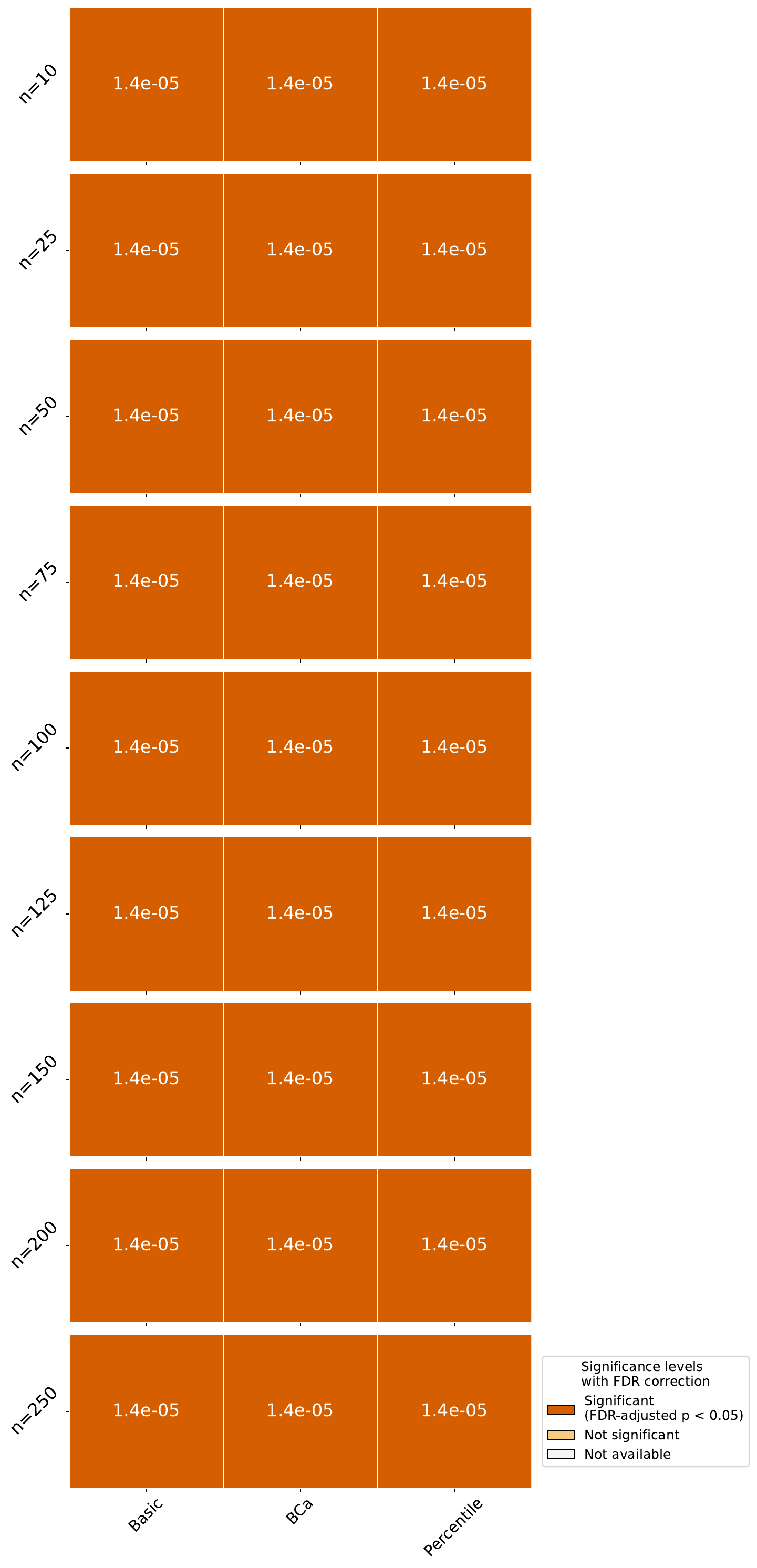}
    \caption{Significance matrix for the claim that Wilson method has better coverage than other bootstrap methods for classification metrics. P-values were obtained using paired permutation tests and corrected using FDR.}
    \label{fig:test_param_boot_classif}
\end{figure}

P-values are reported in Figure \ref{fig:test_param_boot_classif}. We did 9 sample sizes x 3 bootstrap methods = 27 tests and all tests were significant. Nevertheless, the magnitude of the difference becomes small when $n \geq 50$.

\newpage

\subsubsection{Coverages are lower for classification macro than for segmentation}

We are testing whether segmentation metrics produces CIs with lower coverage than any macro-aggregated classification metric.

We then tested for all sample size, if the coverage the CI of the mean for a segmentation metric using percentile method has a higher coverage than the CI of a macro-aggregated classification metric. As segmentation and classification contain different instances, we used unpaired permutation tests.

\begin{figure}[H]
    \centering
    \includegraphics[width=0.9\linewidth]{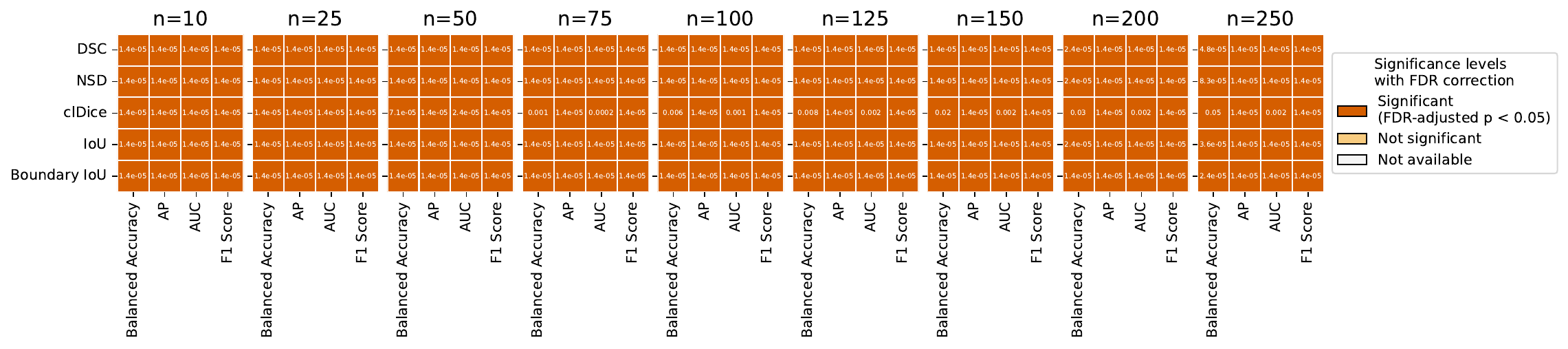}
    \caption{Significance matrix for the claim that macro-aggregated classification metrics have worse coverage than segmentation metrics for CI of the mean. P-values were obtained using unpaired permutation tests and corrected using FDR.}
    \label{fig:test_segm_classif}
\end{figure}

P-values are reported in Figure \ref{fig:test_segm_classif}. We did 9 sample sizes x 4 macro classification metrics x 5 bounded segmentation metrics = 180 tests. All tests were significant.

\subsubsection{CIs are wider for classification than for segmentation}
We are testing whether CIs of classification metrics are wider than for segmentation metrics. 

We tested for all sample sizes, if the CI of any classification micro and macro metric is wider than the CI for the mean using the percentile method for any segmentation metric. 
We used unpaired permutation tests. 

\begin{figure}[H]
    \centering
    \includegraphics[width=0.8\linewidth]{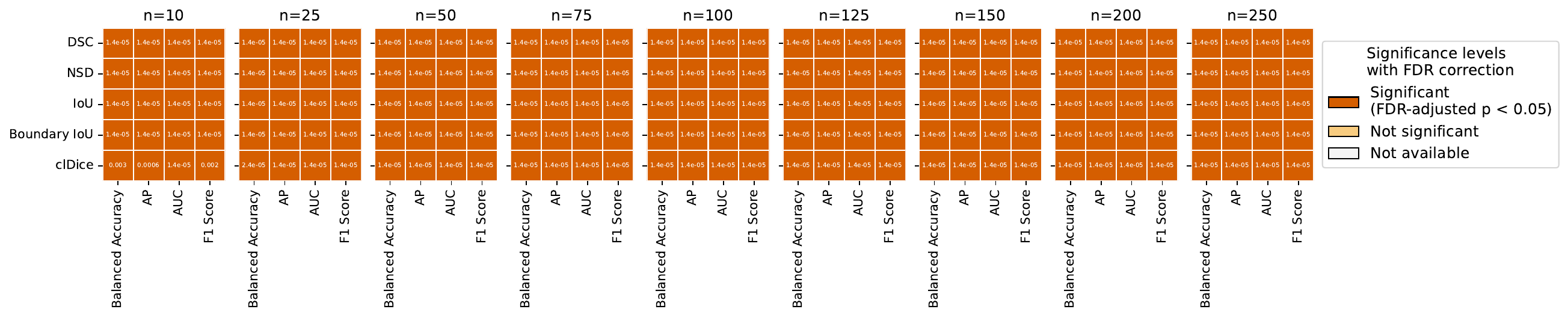}
    \includegraphics[width=0.8\linewidth]{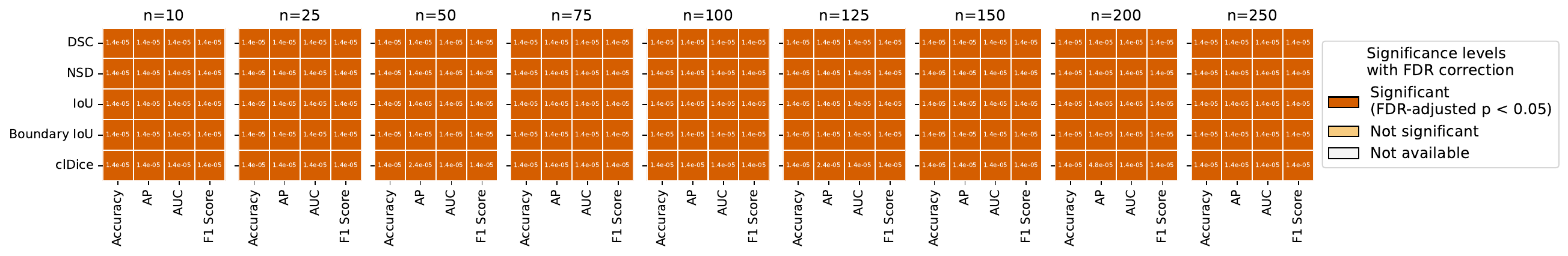}
    \caption{Significance matrix for width difference between macro (top) and  micro (bottom) -aggregated classification metrics and segmentation metrics for CI of the mean. P-values were obtained using unpaired permutation tests and corrected using FDR.}
    \label{fig:test_width_segm_classif}
\end{figure}

P-values are reported in Figure \ref{fig:test_width_segm_classif}. We did 9 sample sizes x 8 classification metrics x 5 bounded segmentation metrics = 360 tests. All tests were significant.

\subsubsection{Different segmentation metrics exhibit different coverage convergence pace}

We are testing whether the CCP are different depending on the segmentation metric. 
We fitted the CCP as described in Supplementary Section \ref{sec:ccp_def}, and tested if the value of segmentation metrics are different using a two sided paired permutation tests. We tested for CI of the mean and std and all CI methods. We performed 2x9x8 segmentation metric couples x ( 5 CI methods  for mean + 3 CI methods for std) = 576 tests. 62/576=10\% of tests were not significant. This is consistent with our claim that did not concern all metrics. Non-significant tests concerned metrics for which CCP is very close to each other, as expected.
P-values are shown in Figure \ref{fig:test_segm}.
%\newpage

\begin{figure}[H]
    \centering
    \includegraphics[width=0.4\linewidth]{supp_figs/tests/cov_segm.pdf}
    \caption{Significance matrix for testing the difference in CCP for segmentation metrics for CI of the mean and std using different CI methods. P-values were obtained using paired permutation tests and corrected using FDR.}
    \label{fig:test_segm}
\end{figure}

\newpage
\subsubsection{Coverage is lower for classification macro than micro}

We are testing whether micro-aggregated classification metrics produces CIs with better coverage than macro-aggregated metrics. 

For all sample sizes, we tested if the coverage of CIs produced using percentile method have a higher coverage for micro-aggregated metrics than their macro-aggregated version. We used paired permutation tests across all classification instances.

\begin{figure}[H]
    \centering
    \includegraphics[width=0.7\linewidth]{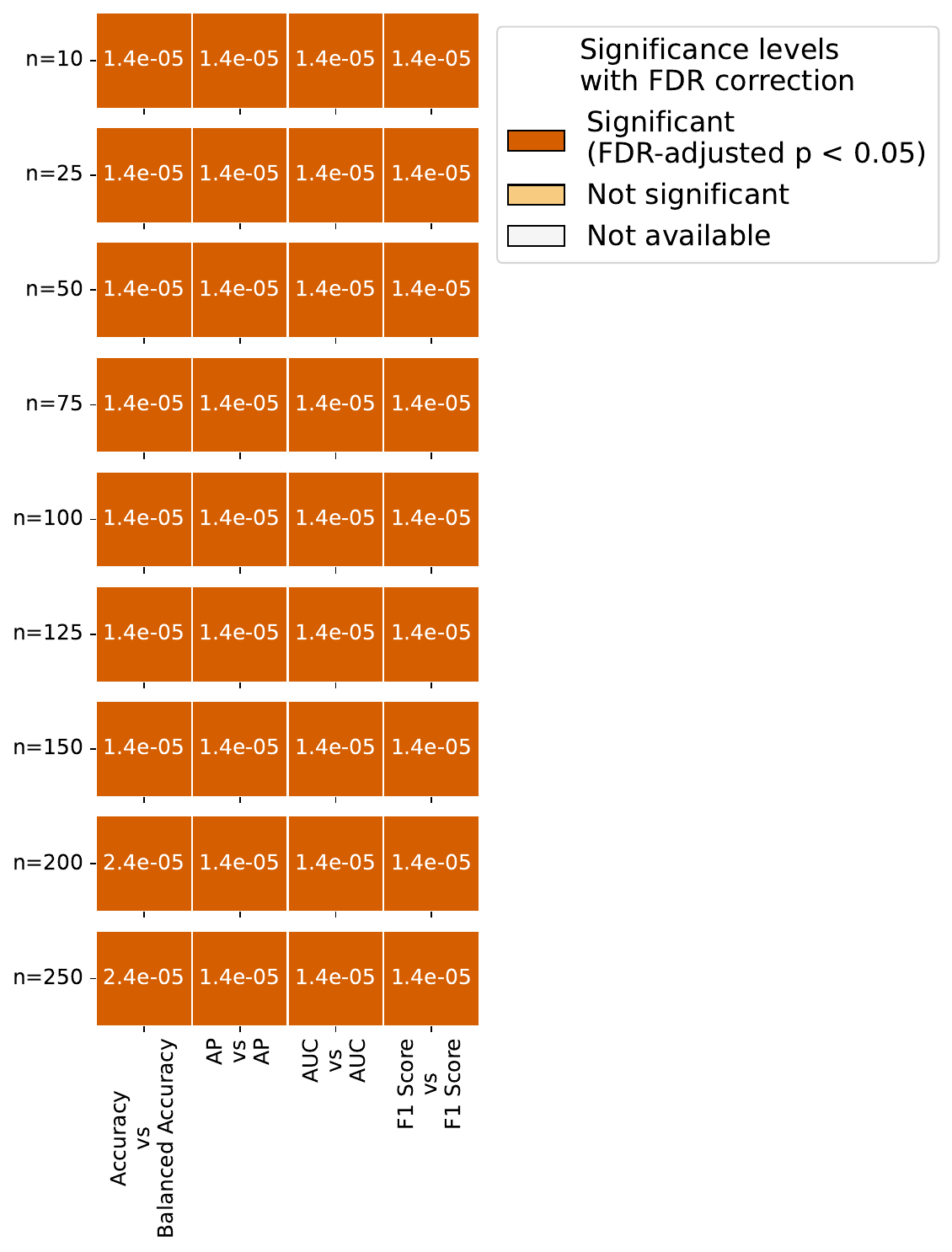}
    \caption{Significance matrix for coverage difference between micro-aggregated metrics and macro-aggregated metrics.}
    \label{fig:test_micro_macro}
\end{figure}

P-values are reported in Figure \ref{fig:test_micro_macro}. We did 9 sample sizes x 4 classification metric couples= 36 tests. All tests were significant.

\newpage

\subsubsection{Coverages of central tendency vs spread measures}

We are testing whether the coverage of CIs of central tendency summary statistics have a better coverage than CIs of spread measures. 
For all samples sizes, we tested if the coverage of CI of the mean is higher than CI of the standard deviation using percentile method, across all segmentation metrics. We did the same test for CIs of the median vs IQR length. We used paired permutation tests.

\begin{figure}[H]
    \centering
    \includegraphics[width=\linewidth]{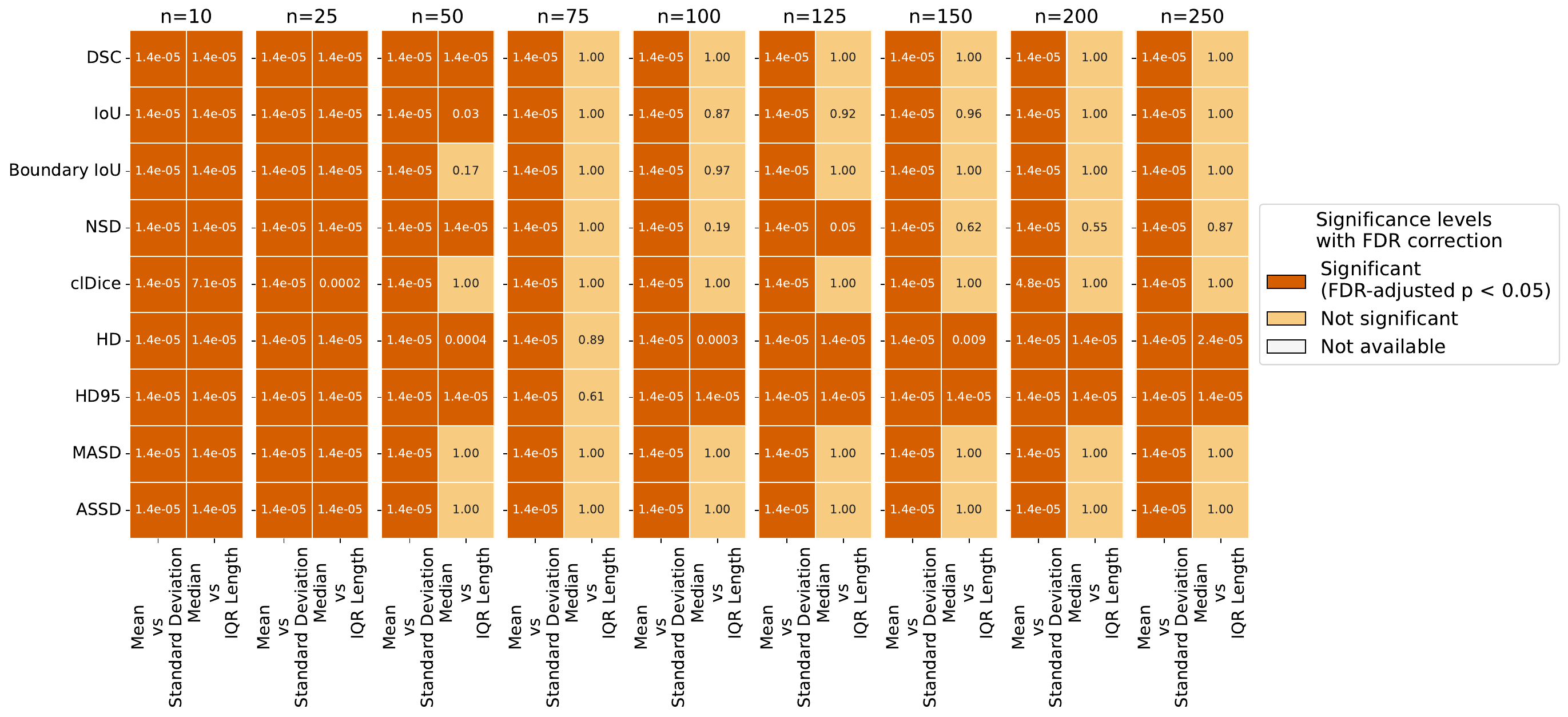}
    \caption{Significance matrix for the claim that the coverage of central tendency summary statistics is higher than spread measure for all segmentation metrics. P-values were obtained using paired permutation tests and corrected using FDR.}
    \label{fig:test_spread_central}
\end{figure}

P-values are reported in Figure \ref{fig:test_spread_central}. We did 9 sample sizes x 2 summary stat couples x 9 segmentation metrics = 162 tests. All tests were significant for mean vs standard-deviation which supports our claim. Only 34/81=42\% of them were significant for median vs IQR. For median vs IQR, they were significant across all metrics only for very small samples ($n=10$ and $n=25$). 

\newpage
\section{Coverage and width: full results}
\label{sec:full_results}
\subsection{Segmentation}
\begin{figure}[H]
    \centering
    \includegraphics[width=0.9\linewidth]{supp_figs/coverages_segm.pdf}
    \caption{Coverages for all metrics and all summary statistics for segmentation instances. To ease zooming, this figure is also available as a separate PDF files at \url{\githubrepofig/blob/main/supplementary/all_cov_segm.pdf}.}
          \vspace{0.2ex}
   {\scriptsize \sl \linespread{0.2}\selectfont If there are instances below $50\%$ or above $99.5\%$ coverage, their number is indicated above a triangle. This applies to other coverage boxplots.
   \par}

    \label{fig:cov_segm_all}
\end{figure}

\begin{figure}[H]
    \centering
    \includegraphics[width=0.9\linewidth]{supp_figs/width_segm.pdf}
    \caption{CI widths for all metrics and all summary statistics for segmentation instances.  To ease zooming, this figure is also available as a separate PDF files at \url{\githubrepofig/blob/main/supplementary/all_width_segm.pdf}.}
    \label{fig:width_segm_all}
\end{figure}

\newpage
\subsection{Classification}
\begin{figure}[H]
    \centering
    \includegraphics[width=0.9\linewidth]{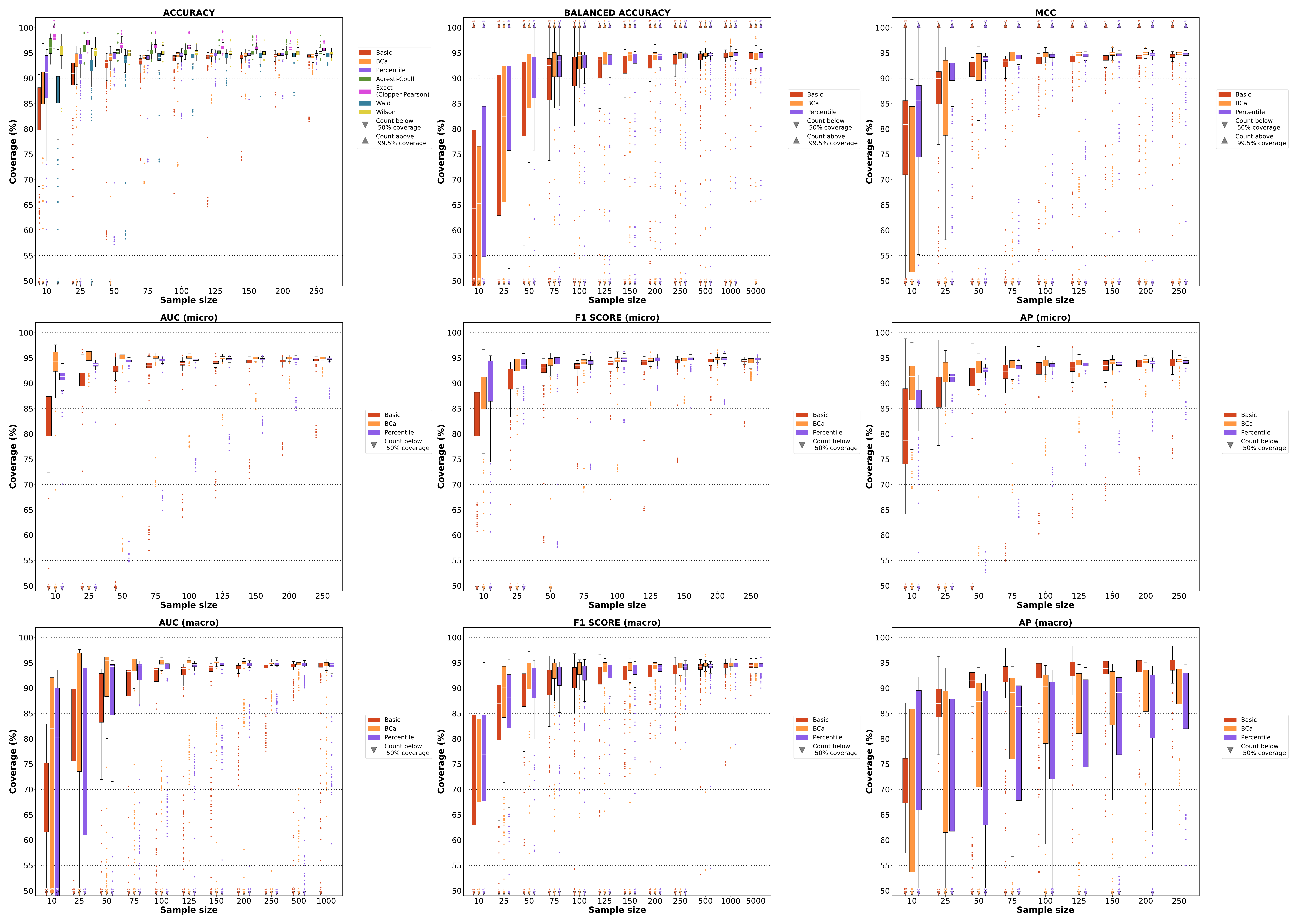}
    \caption{Coverages for all metrics and all aggregation methods for classification instances. To ease zooming, this figure is also available as a separate PDF files at \url{\githubrepofig/blob/main/supplementary/all_cov_classif.pdf}.}
          \vspace{0.2ex}
   {\scriptsize \sl \linespread{0.2}\selectfont If there are instances below $50\%$ or above $99.5\%$ coverage, their number is indicated above a triangle. This applies to other coverage boxplots.
   \par}

    \label{fig:cov_classif_all}
\end{figure}
\newpage
\begin{figure}[H]
    \centering
    \includegraphics[width=0.9\linewidth]{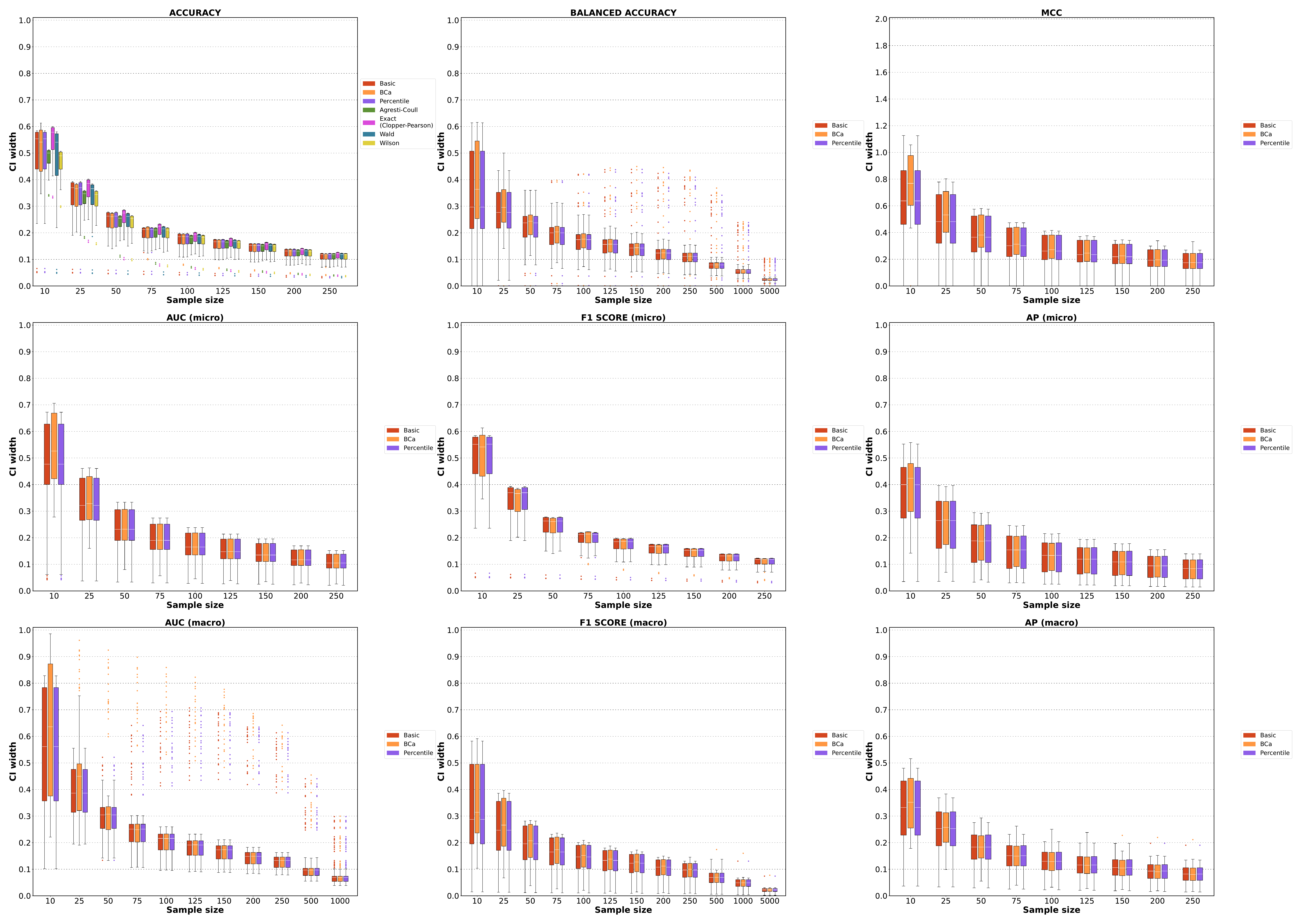}
    \caption{CI widths for all metrics and aggregation methods for classification instances.  To ease zooming, this figure is also available as a separate PDF files at \url{\githubrepofig/blob/main/supplementary/all_width_classif.pdf}.}
    \label{fig:width_classif_all}
\end{figure}

\section{Behavior of different CI methods}
\subsection{BCa fails for CIs of the median}
\begin{figure}[H]
    \centering
    \includegraphics[width=0.95\linewidth]{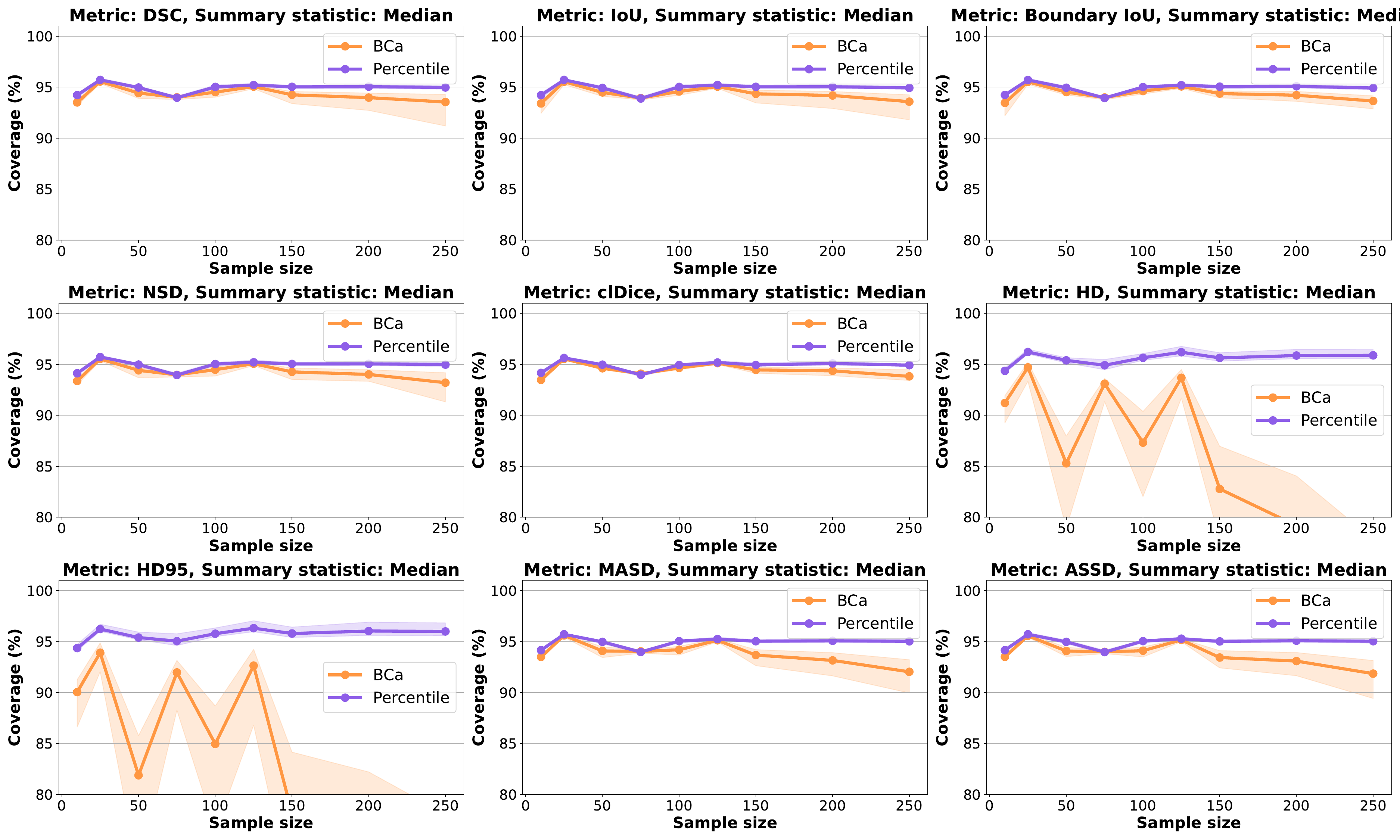}
    \caption{\textbf{BCa bootstrap fails for CIs of the median.} BCa’s coverage (orange) decreases with increasing sample size and can even be highly unstable while that of percentile (purple) is adequate. One plot by segmentation metric (IoU, Boundary IoU, ASSD, clDice, HD, DSC, NSD, MASD, HD95).}
    \label{fig:bca_all}
     \vspace{0.2ex}
   {\scriptsize \sl \linespread{0.2}\selectfont  Solid lines show the median across benchmark instances, with $Q1-Q3$ shown as a shaded region, this applies to other figures unless stated otherwise.
   \par}
\end{figure}

\subsection{Concentration inequalities: extended results}
\label{subsec:concentration_extended}
\begin{figure}[H]
    \centering
     \includegraphics[width=0.9\linewidth]{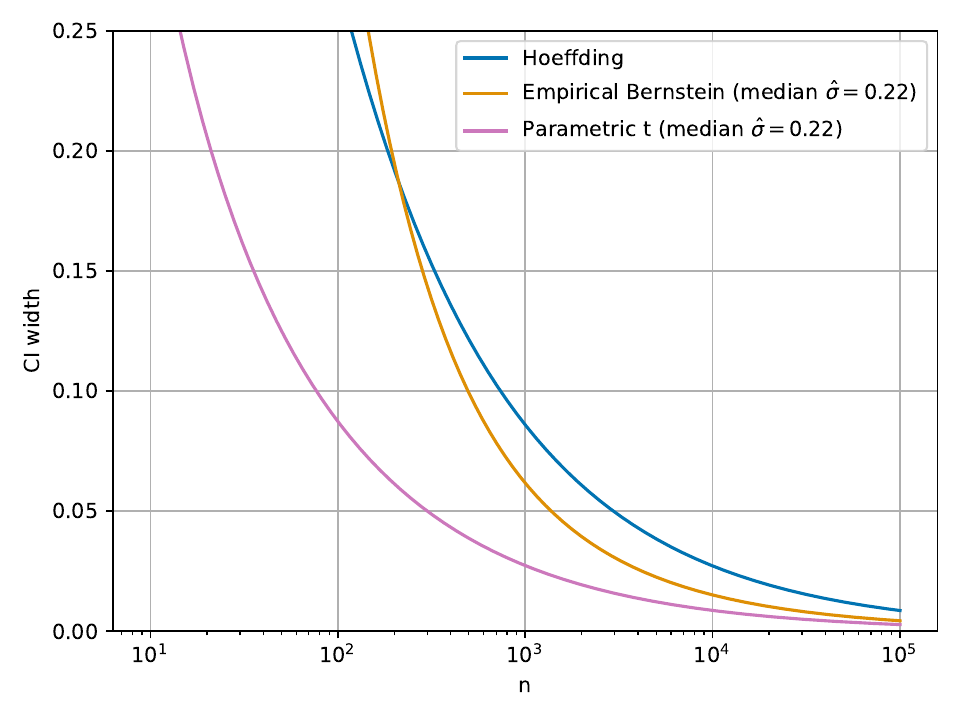}
    \caption{\textbf{Concentration inequalities vs parametric $t$ for $n$ up to $100,000$.} Empirical Bernstein beats Hoeffding for $n>200$ but it is way wider than parametric $t$. It is still more than twice wider than parametric t at $n=2,000$. Empirical Bernstein and parametric $t$ require a standard-deviation (STD) value and are shown for a typical STD, $\hat{\sigma}=0.22$, which is the median of the STD of the DSC over benchmark instances.}
    \label{fig:concentration_10_5}
\end{figure}

\begin{figure}[H]
    \centering
    \includegraphics[width=0.9\linewidth]{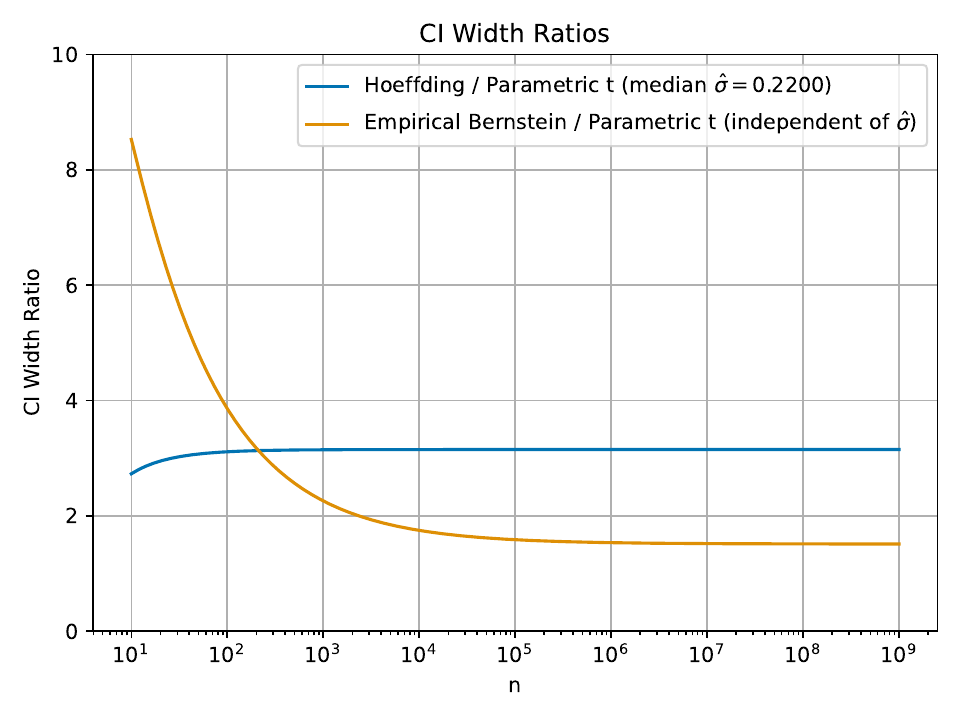}
    \caption{\textbf{Width ratio between CIs from concentration inequalities and parametric $t$ for $n$ up to $10^9$.} The ratio between Empirical Bernstein and parametric $t$ is independent of $\hat{\sigma}$ and converges to $1.51$ when $n\rightarrow \infty$. The ratio between Hoeffding and parametric $t$ depends on $\hat{\sigma}$. For the typical $\hat{\sigma}=0.22$ (median of the standard-deviation of the DSC over benchmark instances), the ratio converges to $1.51$ when $n\rightarrow \infty$. Note that the figure was produced without the “large $n$" approximations done in Supplementary Table~\ref{tab:concentration_summary}.}
    \label{fig:concentration_10_9}
\end{figure}

Supplementary Figure~\ref{fig:concentration_10_5} shows the behavior of CIs from concentration inequalities compared to parametric $t$ for larger sample sizes (up to $n=100,000$). One can observe that Empirical Bernstein is more than twice wider than parametric $t$ at $n=2,000$, meaning that one would need $4$ times more observations to get a given width. Supplementary Figure~\ref{fig:concentration_10_9} shows comparison of widths for even larger samples (up $n=10^9$). For Empirical Bernstein, the ratio converges to $1.51$ when $n\rightarrow \infty$, thus one still needs $2.28$ more observations to achieve a given width. Supplementary Table~\ref{tab:concentration_summary} provides a summary of the CIs from concentration inequalities, their width and their comparison to parametric $t$.

\newpage
\begin{table}[H]
\centering
\scriptsize
\renewcommand{\arraystretch}{2}
\resizebox{\textwidth}{!}{
\begin{tabular}{lccc}
\toprule
 & Hoeffding & Empirical Bernstein & Parametric $t$ \\
\midrule
CI
& $\bar X_n \pm \sqrt{\frac{\log(2/\alpha)}{2n}}$
& $\bar X_n \pm \hat\sigma\sqrt{\frac{2\log(4/\alpha)}{n}} + \frac{7\log(4/\alpha)}{3(n-1)}$
& $\bar X_n \pm t_{1-\alpha/2,n-1}\dfrac{\hat\sigma}{\sqrt n}$ \\
Width
& $2\sqrt{\frac{\log(2/\alpha)}{2n}}$
& $2\!\left(\hat\sigma\sqrt{\frac{2\log(4/\alpha)}{n}} + \frac{7\log(4/\alpha)}{3(n-1)}\right)$
& $2\,t_{1-\alpha/2,n-1}\dfrac{\hat\sigma}{\sqrt n}$ \\
Width ($\alpha=0.05$)
& $\dfrac{2.7162}{\sqrt n}$
& $\hat\sigma\dfrac{5.9208}{\sqrt n} + \dfrac{20.4495}{n-1}$
& $\hat\sigma\dfrac{3.9199}{\sqrt n}$ \\
Large-$n$ width
& $\dfrac{2.7162}{\sqrt n}$
& $\hat\sigma\dfrac{5.9208}{\sqrt n}$
& $\hat\sigma\dfrac{3.9199}{\sqrt n}$ \\
$\text{Width}/W_t$
& $\dfrac{0.6930}{\hat\sigma}$ & $1.5104$ & $1$ \\
$\text{Width}/W_t$ (median $\hat{\sigma}$)
& $3.15$ & $1.5104$ & $1$ \\
\bottomrule
\end{tabular}}
\caption{\textbf{Summary of Hoeffding, Empirical Bernstein and parametric $t$ confidence intervals.}. Numerical values correspond to $95\%$ CIs ($\alpha=0.05$). Large-$n$ widths make the following approximations: 1)~in Empirical Bernstein, the term in $1/n$ is neglected as it is vastly dominated by the term in $1/\sqrt{n}$; 2)~one uses the Gaussian approximation $t_{0.975,n-1}\approx z_{0.975}$. The last two lines show the ratio with respect to the width of parametric $t$ (denoted as $W_t$) for large $n$. The last line shows the value for a typical standard-deviation (STD), $\hat{\sigma}=0.22$, which is the median of the STD of the DSC over benchmark instances.}
\label{tab:concentration_summary}
\end{table}

\section{Segmentation vs classification}

\begin{figure}[H]
    \centering
    \includegraphics[width=0.63\linewidth]{supp_figs/macro_vs_segm_stats.pdf}
    \caption{\textbf{CI coverage and width for classification macro vs bounded segmentation metrics  across all summary statistics.} CIs are obtained using the percentile method. }
    \label{fig:segm_classif}
\end{figure}

\begin{figure}[H]
    \centering
    \includegraphics[width=0.63\linewidth]{supp_figs/micro_vs_segm_stats.pdf}
    \caption{\textbf{CI coverage and width for classification micro vs bounded segmentation metrics  across all summary statistics.} CIs are obtained using the percentile method. }
    \label{fig:segm_classif_micro}
\end{figure}

\newpage
\section{Impact of metrics}

\newpage
\section{Micro vs macro}
\label{sec:micro_macro}

\newpage
\section{Spread measures vs central tendency}

\newpage
\section{Fail cases}
\subsection{Fail cases segmentation}

\begin{figure}[H]
    \centering
   \includegraphics[width=\linewidth]{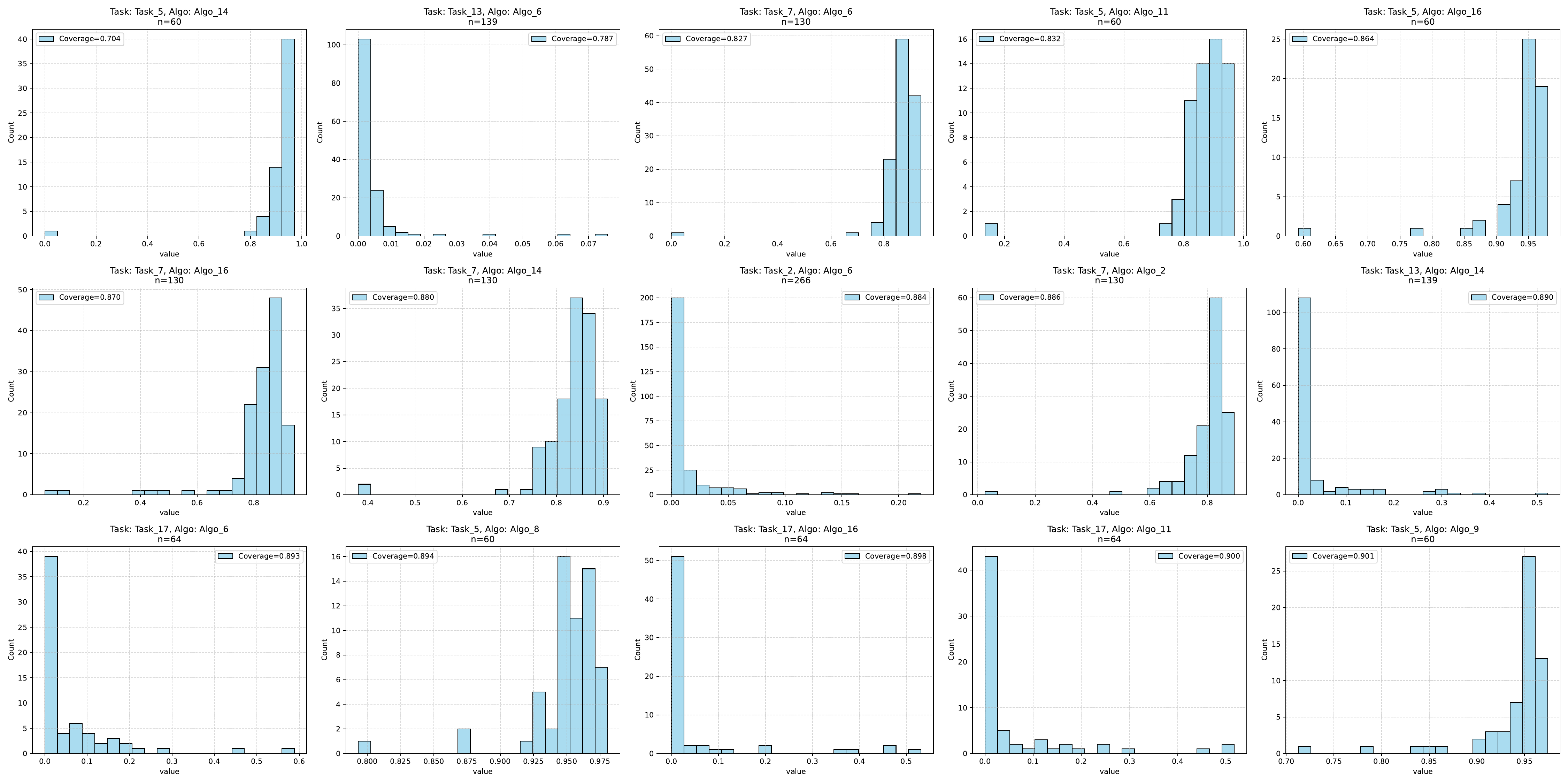}
    \caption{\textbf{Segmentation instances with poor coverage for CIs of the mean DSC.} The CI method is  percentile and $n=50$. These instances present outliers in the empirical distributions of the performance metric.}
    \label{fig:fail_segm}
\end{figure}

\newpage
\subsection{Fail cases classification}
\begin{table}[H]
\centering
\resizebox{\textwidth}{!}{
\begin{tabular}{llrrrrl}
\toprule
Task & Algorithm & Coverage & Accuracy & Total N & Number of classes & Class repartition \\
\midrule
Task 1 & Algo 10 & 0.89 & 0.10 & 733 & 5 & [0.49, 0.27, 0.1, 0.08, 0.05] \\
Task 1 & Algo 11 & 0.91 & 0.16 & 733 & 5 & [0.49, 0.27, 0.1, 0.08, 0.05] \\
Task 1 & Algo 12 & 0.91 & 0.18 & 733 & 5 & [0.49, 0.27, 0.1, 0.08, 0.05] \\
Task 1 & Algo 15 & 0.92 & 0.19 & 733 & 5 & [0.49, 0.27, 0.1, 0.08, 0.05] \\
Task 10 & Algo 3 & 0.89 & 0.10 & 823 & 6 & [0.25, 0.24, 0.23, 0.19, 0.1, 0.0] \\
Task 10 & Algo 7 & 0.90 & 0.11 & 823 & 6 & [0.25, 0.24, 0.23, 0.19, 0.1, 0.0] \\
Task 10 & Algo 12 & 0.92 & 0.18 & 823 & 6 & [0.25, 0.24, 0.23, 0.19, 0.1, 0.0] \\
Task 11 & Algo 2 & 0.92 & 0.88 & 6541 & 2 & [0.98, 0.02] \\
Task 11 & Algo 3 & 0.69 & 0.98 & 6541 & 2 & [0.98, 0.02] \\
Task 11 & Algo 5 & 0.58 & 0.98 & 6541 & 2 & [0.98, 0.02] \\
Task 11 & Algo 6 & 0.59 & 0.98 & 6541 & 2 & [0.98, 0.02] \\
Task 11 & Algo 9 & 0.57 & 0.98 & 6541 & 2 & [0.98, 0.02] \\
Task 11 & Algo 10 & 0.59 & 0.98 & 6541 & 2 & [0.98, 0.02] \\
Task 11 & Algo 11 & 0.58 & 0.98 & 6541 & 2 & [0.98, 0.02] \\
Task 11 & Algo 12 & 0.58 & 0.98 & 6541 & 2 & [0.98, 0.02] \\
\bottomrule
\end{tabular}
}
\caption{\textbf{Characterization of “fail cases" for CIs of the accuracy.} All “fail cases" were from imbalanced datasets. Task, algorithms, coverage value, accuracy value, total number of observations in the test set, number of classes and repartition of individuals in the different classes. Instances were considered as “fail cases" if the CI coverage was $<Q1 - 1.5*IQR$. CIs were computed using percentile bootstrap.}
\label{tab:fail_accuracy}
\end{table}

\begin{table}[H]
\centering
\resizebox{\textwidth}{!}{
\begin{tabular}{llrrrrl}
\toprule
Task & Algorithm & Coverage & Balanced Accuracy & Total n & Number of classes & Class repartition \\
\midrule
Task 2 & Algo 13 & 0.62 & 0.33 & 53 & 3 & [0.66, 0.23, 0.11] \\
Task 4 & Algo 3 & 0.50 & 0.49 & 7624 & 2 & [0.71, 0.29] \\
Task 5 & Algo 16 & 0.39 & 0.50 & 24318 & 2 & [0.71, 0.29] \\
Task 9 & Algo 6 & 0.33 & 0.26 & 121 & 4 & [0.5, 0.17, 0.17, 0.17] \\
Task 9 & Algo 15 & 0.72 & 0.25 & 121 & 4 & [0.5, 0.17, 0.17, 0.17] \\
Task 10 & Algo 5 & 0.44 & 0.16 & 823 & 6 & [0.25, 0.24, 0.23, 0.19, 0.1, 0.0] \\
Task 10 & Algo 6 & 0.62 & 0.25 & 823 & 6 & [0.25, 0.24, 0.23, 0.19, 0.1, 0.0] \\
Task 10 & Algo 7 & 0.30 & 0.23 & 823 & 6 & [0.25, 0.24, 0.23, 0.19, 0.1, 0.0] \\
Task 10 & Algo 8 & 0.17 & 0.29 & 823 & 6 & [0.25, 0.24, 0.23, 0.19, 0.1, 0.0] \\
Task 10 & Algo 10 & 0.56 & 0.25 & 823 & 6 & [0.25, 0.24, 0.23, 0.19, 0.1, 0.0] \\
Task 11 & Algo 1 & 0.22 & 0.44 & 6541 & 2 & [0.98, 0.02] \\
Task 11 & Algo 2 & 0.12 & 0.52 & 6541 & 2 & [0.98, 0.02] \\
Task 11 & Algo 3 & 0.08 & 0.54 & 6541 & 2 & [0.98, 0.02] \\
Task 11 & Algo 4 & 0.20 & 0.47 & 6541 & 2 & [0.98, 0.02] \\
Task 11 & Algo 7 & 0.48 & 0.66 & 6541 & 2 & [0.98, 0.02] \\
Task 11 & Algo 8 & 0.28 & 0.39 & 6541 & 2 & [0.98, 0.02] \\
Task 11 & Algo 13 & 0.23 & 0.41 & 6541 & 2 & [0.98, 0.02] \\
Task 11 & Algo 14 & 0.27 & 0.45 & 6541 & 2 & [0.98, 0.02] \\
Task 11 & Algo 15 & 0.34 & 0.61 & 6541 & 2 & [0.98, 0.02] \\
Task 11 & Algo 16 & 0.50 & 0.50 & 6541 & 2 & [0.98, 0.02] \\
Task 11 & Algo 17 & 0.39 & 0.41 & 6541 & 2 & [0.98, 0.02] \\
Task 11 & Algo 18 & 0.26 & 0.51 & 6541 & 2 & [0.98, 0.02] \\
Task 11 & Algo 19 & 0.26 & 0.37 & 6541 & 2 & [0.98, 0.02] \\
\bottomrule
\end{tabular}}
\caption{\textbf{Characterization of “fail cases" for CIs of the balanced accuracy.} $78\% (18/23)$  of the “fail cases" were from imbalanced datasets. Task, algorithms, coverage value, balanced accuracy value, total number of observations in the test set, number of classes and repartition of individuals in the different classes. Instances were considered as “fail cases" if the CI coverage was $<Q1 - 1.5*IQR$. CIs were computed using percentile bootstrap.}
\label{tab:fail_ba}
\end{table}

\section{CIs produced by the basic bootstrap are shifted}

\newpage
\section{Failure of the BCa confidence interval for the median}
\label{sec:fail_bca}
As presented in Section \ref{BCa_appendix}, the BCa interval relies on the computation of the acceleration coefficient $a$:
\begin{equation}
    a = \frac{\sum_i (\bar{\theta}_{(.)} - \Hat{\theta}_{(i)})^3}{6[\sum_i(\bar{\theta}_{(.)} - \Hat{\theta}_{(i)})^2]^{3/2}}
    \label{eqn:bca_a_bis}
\end{equation}

where $(\Hat{\theta}_{(i)})_{i=1}^n$ are the jackknife estimates of the statistic  and $\bar{\theta}_{(.)} = \frac{1}{n}\sum_i \Hat{\theta}_{(i)}$ is their mean. Here the statistic is the median and thus $\Hat{\theta}_{(i)}$ is the median of the leave-one-out sample without the $i$-th observation). Please note the jackknife estimates are computed from the original sample and not from the bootstrap samples.

Let us now look at the possible values of $\Hat{\theta}_{(i)}$. We will assume without loss of generality that the random sample $(X_i)_{i=1}^n$ is sorted. There are two cases, depending on whether $n$ is even  or odd. If $n$ is even, ($n = 2k$), $\Hat{\theta}_{(i)}$ has only two possible values:
$$
\Hat{\theta}_{(i)} =
\begin{cases}
    X_{k+1}, & i\le k\\
    X_k, & i\ge k
\end{cases}
$$

%We know that, in such a case, the sum of squared differences to the mean is 
$$\frac{n(X_k-X_{k+1})^2}{4}$$

%It is easy to see that if $X_k\approx X_{k+1}$, then the denominator will be almost null, leading to a numerical error. We will now consider the second case.\\

If $n$ is odd ($n = 2k+1$),  $\Hat{\theta}_{(i)}$ has only three possible values:
$$
\Hat{\theta}_{(i)} =
\begin{cases}
    \frac{X_k + X_{k+1}}{2}, & i\le k\\
    \frac{X_{k-1} + X_{k+1}}{2}, & i=k+1\\
    \frac{X_{k-1} + X_{k}}{2}, & i\ge k+2
\end{cases}
$$
% \begin{cases}
%     b := \frac{X_k + X_{k+1}}{2}, & i\le k\\
%     c := \frac{X_{k-1} + X_{k+1}}{2}, & i=k+1\\
%     d := \frac{X_{k-1} + X_{k}}{2}, & i\ge k+2
% \end{cases}

Thus, $\Hat{\theta}_{(i)}$ can at most take three different values.  It is thus a very poor estimation of how fast the standard error of the statistic changes as its true value changes. This poor behavior for the median had only been noted by \citetsupp{abdi2010jackknife}. Further, if $X_{k}=X_{k+1}$ ($X_{k-1}=X_{k}=X_{k+1}$ if $n$ is even), $\Hat{\theta}_{(i)}$ collapses to a single value, the numerator and denominator of Equation~\ref{eqn:bca_a_bis} are zero and $a$ is undefined. If $X_{k}\approx X_{k+1}$ (resp. $X_{k-1}\approx X_{k}\approx X_{k+1}$), one gets numerical instability.

\newpage
\section{Bootstrap computation times}
\label{sec:suppl_bootstrap_time}
Here, we provide the reader with typical computation times for the bootstrap as a function of sample size $n$. The computational complexity of the percentile bootstrap is $O(B\times T(n))$ where $T(n)$ is the computational complexity of the summary statistic (or sample-level metric) for a sample of size $n$ and $B$ is the number of bootstrap resamplings ($B=9,999$ in our study). For some common summary statistics (e.g. mean, accuracy), $T(n)=O(n)$. For some others (e.g. median, AUC), $T(n)=O(n\log(n))$ when no specific optimization is performed (like enforcing bootstrap samples to be already sorted, e.g. by performing sorted bootstrap). Higher computational complexities are rare. Importantly, note that the computation time of {\sl individual-level} metric (e.g. DSC, NSD, HD) plays no role here since they are only computed once, before the bootstrap procedure. Typical values are provided in Supplementary Table~\ref{tab:bootstrap_times_max}. There are worst-case scenarii as these were computed without any optimization on a simple MacBook Air M1 from 2021 with a single core. Even for a sample of one million, computation times are low and almost negligible compared to deep learning training.

\begin{table}[H]
\centering
%\resizebox{\textwidth}{!}{
\begin{tabular}{llll}
\toprule
Sample size $n$ & Metric & Metric complexity & Time\\
\midrule
1,000 & Accuracy   & $O(n)$ & 4.73 s\\
1,000 & AUC        & $O(n\log(n))$ & 22.72 s\\
10,000 & Accuracy   & $O(n)$ & 8.93 s\\
10,000 & AUC        & $O(n\log(n))$ & 1.02 min\\
100,000 & Accuracy   & $O(n)$ & 48.07 s\\
100,000 & AUC        & $O(n\log(n))$ & 10.25 min\\
1,000,000 & Accuracy   & $O(n)$ & 6.62 min\\
1,000,000 & AUC & $O(n\log(n))$ & 2.38 h \\
\bottomrule
\end{tabular}
%}
\caption{\textbf{Typical bootstrap computation times without optimization, depending on the sample-level metric.} This table provides typical percentile bootstrap computation times  with $B=9,999$ boostrap resamplings. No code optimization was performed and standard metric implementations were used. Computations were done with a standard MacBook Air M1 using one core. For these reasons, the values differ from that of Table~\ref{tab:combined_times_vertical_groups} but the order of magnitudes are consistent.}
\label{tab:bootstrap_times_max}
\end{table}

\newpage
\section{KDE sensitivity analyses}

\subsection{Ablation definitions}
\label{sec:abl}
As we presented in Section \ref{sec:kde}, a KDE procedure is composed of 3 elements: a kernel, a bandwidth selection procedure and a boundary correction method. Our proposed KDE approach uses the Epanechnikov kernel, and adaptive bandwidth selection method, and a modifier trimming to account for the boundedness of the considered distributions.
To study the robustness of our claims to the KDE settings, we chose to change the kernel and the bandwidth selection procedure. We considered the Gaussian kernel and Scott's bandwidth, as these are two of the most usual parameters used to perform KDE. 

As the Gaussian kernel is unbounded, using it to perform KDE of unbounded distributions requires using an appropriate boundary correction method. However, the proposed modifier trimming method only works with bounded kernels. Thus, we resorted to using the truncation/renormalization method to correct for boundaries. This explains why the ablations 2 and 3 (see Table \ref{tab:kde_configs}) vary both the kernel and the boundary correction method. 

\newpage
\subsection{Ablation results: Robustness of claims}

\begin{table}[H]
\centering
\setlength{\tabcolsep}{4pt}
\renewcommand{\arraystretch}{1.15}
\begin{tabular}{@{} L{4.6cm} *{4}{C{2.15cm}} @{}}
\toprule
\diagbox[width=\dimexpr 4.4cm+2\tabcolsep\relax, height=3\line]{\textbf{Tests}}{\textbf{KDE}}
  & \makecell{Epan. \\ adaptive \\ trimmed}
  & \makecell{Epan.\\Scott \\ trimmed}
  & \makecell{Gaussian\\adaptive}
  & \makecell{Gaussian\\Scott} \\
\midrule
1.\ Basic                                    & 90.5\%  & 90.6\%  & 87.8\%  & 88.1\%  \\
2.\ BCa                                      & 100.0\% & 100.0\% & 99.0\%  & 100.0\% \\
3.\ Coverage: parametric $t$ vs.\ bootstrap  & 65.8\%  & 65.0\%  & 64.2\%  & 65.4\%  \\
4.\ Width: parametric $t$ vs.\ bootstrap     & 90.5\%  & 90.5\%  & 90.5\%  & 90.5\%  \\
5.\ Wilson vs.\ bootstrap                    & 100.0\% & 100.0\% & 100.0\% & 100.0\% \\
6.\ Coverage: segmentation vs.\ classification  & 100.0\% & 99.4\%  & 100.0\% & 100.0\% \\
7.\ Width: segmentation vs.\ classification     & 100.0\% & 100.0\% & 100.0\% & 100.0\% \\
8.\ Segmentation metrics                     & 89.2\%  & 87.5\%  & 87.2\%  & 87.2\%  \\
9.\ Micro vs.\ macro averaging               & 100.0\% & 100.0\% & 100.0\% & 100.0\% \\
10.\ Spread vs.\ central tendency            & 71.0\%  & 69.1\%  & 70.4\%  & 69.8\%  \\
\bottomrule
\end{tabular}
\caption{\textbf{KDE sensitivity analysis: robustness of claims.}
\% of significant tests (p-value $<0.05$) for each result in the paper.
Details of the results can be found in Section~\ref{sec:results} and the
corresponding tests in \ref{sec:tests}. One can see that these proportions are very stable across different KDE configurations. \\ \footnotesize\itshape The claims are as follows. 1) “Basic”: in the vast majority of cases, coverage of basic bootstrap is lower than the coverage of any other bootstrap method. 2) “BCa”: coverage of CI of the mean using BCa bootstrap method is higher than coverage of percentile method across all segmentation metrics as well as for micro AUC and micro AP. 3) “Cov param t vs bootstrap”: For small sample sizes, coverage of the CI of the mean using parametric t method is higher than all bootstrap methods across all segmentation metrics. 4) “Width param t vs bootstrap”: CIs of the mean using parametric t method are larger than the ones of all  bootstrap methods in the vast majority of cases. 5) “Wilson vs bootstrap”: Coverage of the Wilson CI for the accuracy is higher than all bootstrap methods. 6) “Cov segm vs classif”: Coverage of the CI for any macro classification metric is less than the coverage of any bounded segmentation metric for CI of the mean. 7) “Width segm vs classif”: CIs of any micro and macro classification metric are wider than those of any bounded segmentation metric for CI of the mean (using percentile method). 8) “Segmentation metrics”: the coverage convergence pace of CIs of the mean and standard deviation is different depending on the segmentation metric. 9) “Micro vs macro”: the coverage of the micro aggregated version of any classification metric is higher than that of the macro version (using percentile method). 10) “Spread vs central”: the coverages of CIs of central measures are higher than those of spread measures (using percentile method), across all segmentation metrics, this holding across sample sizes for mean vs STD but only for very small samples for median vs IQR.}
\label{tab:kde_sensitivity_claims}
\end{table}

\subsection{Ablation results: coverage and width}

Here, we provide results for ablation studies of the KDE procedure, varying both the kernel and the bandwidth method used. 
We report raw coverage and width values using the same type of figure as Figures \ref{fig:cov_segm_all} to \ref{fig:width_classif_all}. 

\subsubsection{Segmentation}

\begin{figure}[H]
    \centering
    \includegraphics[width=0.9\linewidth]{supp_figs/epanechnikov_scott_trimmed/coverages_segm.pdf}
    \caption{Coverages for all metrics and all summary statistics for segmentation instances for Ablation 1. To ease zooming, this figure is also available as a separate PDF files at \url{\githubrepofig/blob/main/supplementary/epanechnikov_scott_trimmed/coverages_segm.pdf}.}
    \label{fig:cov_segm_all_ep_scott_trimmed}
\end{figure}

\begin{figure}[H]
    \centering
    \includegraphics[width=0.9\linewidth]{supp_figs/epanechnikov_scott_trimmed/width_segm.pdf}
    \caption{CI widths for all metrics and all summary statistics for segmentation instances for Ablation 1. To ease zooming, this figure is also available as a separate PDF files at \url{\githubrepofig/blob/main/supplementary/epanechnikov_scott_trimmed/width_segm.pdf}.}
    \label{fig:width_segm_all_ep_scott_trimmed}
\end{figure}

\begin{figure}[H]
    \centering
    \includegraphics[width=0.9\linewidth]{supp_figs/gaussian_adaptive/coverages_segm.pdf}
    \caption{Coverages for all metrics and all summary statistics for segmentation instances for Ablation 2. To ease zooming, this figure is also available as a separate PDF files at \url{\githubrepofig/blob/main/supplementary/gaussian_adaptive/coverages_segm.pdf}.}
    \label{fig:cov_segm_all_gauss_adaptive}
\end{figure}

\begin{figure}[H]
    \centering
    \includegraphics[width=0.9\linewidth]{supp_figs/gaussian_adaptive/width_segm.pdf}
    \caption{CI widths for all metrics and all summary statistics for segmentation instances for Ablation 2. To ease zooming, this figure is also available as a separate PDF files at \url{\githubrepofig/blob/main/supplementary/gaussian_adaptive/width_segm.pdf}.}
    \label{fig:width_segm_all_gauss_adaptive}
\end{figure}

\begin{figure}[H]
    \centering
    \includegraphics[width=0.9\linewidth]{supp_figs/gaussian_scott/coverages_segm.pdf}
    \caption{Coverages for all metrics and all summary statistics for segmentation instances for Ablation 3. To ease zooming, this figure is also available as a separate PDF files at \url{\githubrepofig/blob/main/supplementary/gaussian_scott/coverages_segm.pdf}.}
    \label{fig:cov_segm_all_gauss_scott}
\end{figure}

\begin{figure}[H]
    \centering
    \includegraphics[width=0.9\linewidth]{supp_figs/gaussian_scott/width_segm.pdf}
    \caption{CI widths for all metrics and all summary statistics for segmentation instances for Ablation 3. To ease zooming, this figure is also available as a separate PDF files at \url{\githubrepofig/blob/main/supplementary/gaussian_scott/width_segm.pdf}.}
    \label{fig:width_segm_all_gauss_scott}
\end{figure}

\subsubsection{Classification}

\newpage
\subsection{Proof that truncation/renormalization method produces a biased expectation}
\label{sec:trunc_renorm}
\subsubsection{Setup and notations}

Let $x_1, \dots, x_n$ be observations from a distribution supported on
$[a,b]$, where $a$ and/or $b$ may be infinite. Let $\bar{x} = \frac{1}{n}\sum_{i=1}^n x_i$ the empirical mean. Consider a KDE with
symmetric kernel $K$ and per-point bandwidths $h_i > 0$:
\begin{equation}\label{eq:kde}
  \hat{f}(x)
  = \frac{1}{n}\sum_{i=1}^n \frac{1}{h_i}\,
    K\!\left(\frac{x - x_i}{h_i}\right).
\end{equation}

One realization $\hat{X} \sim \hat{f}$ is obtained as follows:
\begin{enumerate}
  \item Pick $I \sim \mathrm{Uniform}\{1,\dots,n\}$.
  \item Draw $U \sim K$.
  \item Set $\hat{X} = x_I + h_I\, U$.
\end{enumerate}

\subsubsection{Proposed adaptive KDE}

The proposed KDE handles the bounded case through trimming the local bandwidth modifiers (Algorithm~\ref{alg:kde_ours}). As a result, the kernel stays symmetric.

\begin{proposition}\label{prop:proposed}
With the proposed adaptive KDE of Algorithm~\ref{alg:kde_ours}:
\begin{equation}
  \E_{\mathrm{U}}[\hat{X}] = \bar{x}, \quad \text{(unbiased)} 
\end{equation}
\end{proposition}

\begin{proof}
$$\E_{\mathrm{U}}[\hat{X}]
  = \frac{1}{n}\sum_{i=1}^n \bigl(x_i + h_i\,\E[U]\bigr).
$$

Since trimming only rescales $h_i$
and never truncates the kernel, the kernel stays symmetric and $\E[U] = 0$. Therefore
$$
  \E_{\mathrm{U}}[\hat{X}]
  = \bar{x}.
$$
\end{proof}

\subsubsection{Truncation/renormalization approach}

For each point $x_i$, work in the standardized coordinate
$u = (x - x_i)/h_i$ of its kernel, in which the interval endpoints
$a$ and $b$ map to
$$
  \alpha_i = \frac{a - x_i}{h_i},
  \qquad
  \beta_i = \frac{b - x_i}{h_i}.
$$
The kernel mass of point $i$ falling outside $[a,b]$ splits into a
left part (below $a$) and a right part (above $b$),
$$
  w_i^a = \int_{-\infty}^{\alpha_i} K(u)\,\dd u,
  \qquad
  w_i^b = \int_{\beta_i}^{\infty} K(u)\,\dd u,
$$
so that the mass remaining inside $[a,b]$ (the per-point
renormalization constant) is
$$
  C_i = 1 - w_i^a - w_i^b.
$$

The renormalized KDE is:
$$
  \hat{f}_{\mathrm{renorm}}(x)
  = \frac{1}{n}\sum_{i=1}^n \frac{1}{C_i\, h_i}\,
    K\!\left(\frac{x - x_i}{h_i}\right),
  \qquad x \in [a,b].
$$
Drawing $\hat{X}\sim\hat{f}_{\mathrm{renorm}}$ amounts to drawing $R_I$ from
$K$ truncated to $[\alpha_I, \beta_I]$ with density $K(u)/C_I$, then
setting $\hat{X} = x_I + h_I R_I$.

Define the truncated kernel expectation:
$$
  \mu^{(i)}
  = \frac{1}{C_i}\int_{\alpha_i}^{\beta_i} u\, K(u)\,\dd u,
$$
and the per-component quantities:
$$
  \delta_i = h_i\,\mu^{(i)}
$$

\begin{proposition}\label{prop:mean-R}
\begin{equation}
  \E_{\mathrm{R}}[\hat{X}]
  = \bar{x} + \bar{\delta},
\end{equation}
where $\bar{\delta} = \frac{1}{n}\sum_{i=1}^n \delta_i$ and:
\begin{equation}\label{eq:delta-i}
  \delta_i
  = \frac{%
      (x_i - a)\,w_i^a
    - (b - x_i)\,w_i^b
    + \ell_i^a - \ell_i^b
  }{C_i},
\end{equation}
with per-point expected overshoots:
$$
  \ell_i^a
  = \int_0^\infty y\,\frac{1}{h_i}\,
    K\!\left(\frac{a - y - x_i}{h_i}\right)\dd y,
  \qquad
  \ell_i^b
  = \int_0^\infty y\,\frac{1}{h_i}\,
    K\!\left(\frac{b + y - x_i}{h_i}\right)\dd y.
$$
\end{proposition}

\begin{proof}
Since $\E[R_I \mid I=i] = \mu^{(i)}$:
$$
  \E_{\mathrm{R}}[\hat{X}]
  = \frac{1}{n}\sum_{i=1}^n
    \bigl(x_i + h_i\,\mu^{(i)}\bigr)
  = \bar{x} + \bar{\delta}.
$$
To derive~\eqref{eq:delta-i}, use the symmetry of $K$ to write:
$$
  C_i\,\mu^{(i)}
  = \int_{\alpha_i}^{\beta_i} u\,K(u)\,\dd u
  = -\int_{-\infty}^{\alpha_i} u\,K(u)\,\dd u
   - \int_{\beta_i}^{\infty} u\,K(u)\,\dd u.
$$
For the left tail, substitute $u = \alpha_i - v$, $v \geq 0$:
$$
  \int_{-\infty}^{\alpha_i} u\,K(u)\,\dd u
  = \alpha_i \int_{-\infty}^{\alpha_i} K(u)\,\dd u
    - \int_0^\infty v\,K(\alpha_i - v)\,\dd v
  = \alpha_i\, w_i^a - \frac{\ell_i^a}{h_i}.
$$
For the right tail, substitute $u = \beta_i + v$, $v \geq 0$:
$$
  \int_{\beta_i}^{\infty} u\,K(u)\,\dd u
  = \beta_i\, w_i^b + \frac{\ell_i^b}{h_i}.
$$
Therefore:
$$
  C_i\,\mu^{(i)}
  = -\alpha_i\,w_i^a + \frac{\ell_i^a}{h_i}
    -\beta_i\,w_i^b - \frac{\ell_i^b}{h_i}.
$$
Multiplying by $h_i$ and using $h_i\alpha_i = a - x_i$,
$h_i\beta_i = b - x_i$:
$$
  C_i\,\delta_i
  = (x_i - a)\,w_i^a + \ell_i^a
    - (b - x_i)\,w_i^b - \ell_i^b.
$$
When at least one point lies close enough to a boundary that its
kernel leaks mass ($w_i^a > 0$ or $w_i^b > 0$), the terms in
$\delta_i$ do not cancel in general, so $\bar\delta \neq 0$ and the
expectation is biased. The bias vanishes only in the degenerate
configuration where every point is interior
($w_i^a = w_i^b = 0$ for all $i$) or where left and right
contributions cancel exactly.
\end{proof}

\section{Decision tree creation procedure}
\label{sec:decision_tree}

We constructed preliminary trees as follows. We discarded CIs of the standard deviation and interquartile range, as only few people report CI for these quantities. We considered the order of hierarchy: Task type (classification/segmentation) $\rightarrow$ Aggregation $\rightarrow$ Metric $\rightarrow$ Sample size $\rightarrow$ CI method. Each method is characterized by its typical behavior (median coverage across instances) and a worst-case behavior (lower whisker of the boxplot). Each leaf was tagged with either: i) a green checkmark if the typical value is $92\%\le x$ and the worst-case is $\ge90\%$, ii) an orange warning sign if the typical value is $92\%\le x$ but the worst-case is $<90\%$ and iii) a red cross if the typical value is $<92\%$.

A fusion was performed on leaves, following these rules: 

\begin{itemize}
    \item If some CI methods are marked with a checkmark, list all methods fulfilling this criterion that have a coverage $\le96\%$. If there is only one method, it is kept even if its coverage is $\ge96\%$.
    \item Else, if some CI methods are marked with a warning, choose the one with the highest typical behavior, or the highest worst-case behavior in case of tie
    \item If all CI methods are marked with a red cross, no method is suitable
\end{itemize}

We then performed an iterative aggregation process in order to reduce both the number of leaves and the number of methods per leaf. This process relied on the conclusions reported in the results section and on a systematic inspection of all box plots. It was guided by two objectives: keeping the number of distinct CI methods needed within a given study as low as possible, and ensuring that the recommendations remain safe, in the sense that no method was ever upgraded from orange to green.

%\bibliographystylesupp{abbrvnat}
\newpage
\bibliographystylesupp{apalike}
\bibliographysupp{references}

\end{document}